\documentclass{article}

\usepackage[final,main]{neurips_2026}
\usepackage{multirow}
\usepackage{amsmath}
\usepackage{graphicx}

\usepackage[utf8]{inputenc} %
\usepackage[T1]{fontenc}    %
\usepackage{hyperref}       %
\usepackage{url}            %
\usepackage{booktabs}       %
\usepackage{placeins}       %
\usepackage{float}          %
\usepackage{needspace}      %
\usepackage{amsfonts}       %
\usepackage{nicefrac}       %
\usepackage{microtype}      %
\usepackage[table]{xcolor}  %
\usepackage{amsmath, amssymb, amsthm}
\usepackage{bm}
\usepackage{subcaption}
\usepackage{algorithm}
\usepackage{diagbox}
\usepackage{algpseudocode}
\usepackage{xfrac}
\usepackage{mathrsfs}
\usepackage[most]{tcolorbox}
\usepackage{tikz}
\usepackage{enumitem}
\usetikzlibrary{calc}
\usepackage{epigraph}

\usepackage{tabularx}
\usepackage{array}

\newcolumntype{Y}{>{\raggedright\arraybackslash}X}

\csname theoremstyle\endcsname{plain}
\newtheorem{proposition}{Proposition}
\newtheorem{corollary}{Corollary}
\newtheorem{lemma}{Lemma}
\newtheorem{theorem}{Theorem}
\newtheorem*{nonproposition}{Proposition}
\csname theoremstyle\endcsname{remark}
\newtheorem{remark}{Remark}

\usepackage[normalem]{ulem}
\usepackage{etoolbox}
\AtEndEnvironment{quote}{\vspace{-0.2em}}

\newcommand{\best}[1]{{\color{red}{\textbf{#1}}}}
\newcommand{\secbest}[1]{{\color{blue}{\textbf{#1}}}}

\definecolor{accentblue}{HTML}{1F4E79}
\definecolor{accentteal}{HTML}{1F6F78}
\definecolor{softblue}{HTML}{F4F8FC}
\definecolor{softteal}{HTML}{F2FAF8}
\definecolor{framegray}{HTML}{D7DEE8}
\definecolor{textgray}{HTML}{5B6573}

\csname tcbset\endcsname{
methodboxstyle/.style={
enhanced,
breakable,
colback=white,
colframe=framegray,
boxrule=0.35pt,
arc=1.2mm,
left=1.6mm,
right=1.6mm,
top=1.0mm,
bottom=1.0mm,
before skip=8pt,
after skip=8pt,
borderline west={1.1pt}{0pt}{accentblue}
}
}

\newtcolorbox{methodbox}[1][]{methodboxstyle,#1}
\algrenewcommand\algorithmiccomment[1]{\hfill{\color{textgray}\footnotesize $\triangleright$ #1}}

\title{Echoes of Error: Residual-Directional Local Rollout Consistency for Timeseries Forecasting}

\author{%
  Zheng Wang\thanks{Corresponding authors: Zheng Wang and Kaixuan Zhang.}\quad
  Kaixuan Zhang\footnotemark[2]\quad
  Xiaonan Lu\quad
  Wanfang Chen\\
  Bosch Center for Artificial Intelligence (BCAI)\\
  Bosch (China) Investment Co., Ltd.\\
  \texttt{\{david.wang3,kaixuan.zhang,xiaonan.lu,wanfang.chen\}@cn.bosch.com}
}
\begin{document}

\setcounter{footnote}{1}
\maketitle
\setcounter{footnote}{0}

\begin{abstract}

In time-series forecasting, rollout errors are not merely one-step fitting discrepancies: once fed back into the evolving state, they can perturb future predictions. We formalize this effect through \emph{local rollout consistency}, which compares the next prediction produced under teacher forcing with that produced under free running.
In principle, this consistency can be enforced directly by tracking the residual-induced perturbation of the next state. In practice, however, the residual-to-state alignment operator is often difficult to design when the input state and output block have different dimensions, heterogeneous components, or architecture-dependent layouts. We therefore use a first-order local surrogate that penalizes the Jacobian-vector product of the predictor along the actual feedback perturbation, suppressing precisely the directions through which self-generated errors remain visible to future predictions.
We further give the same loss a geometric interpretation near a predictive manifold: by controlling residual-induced directions that probe the normal bundle, the objective yields theoretical conditions under which free-running trajectories remain inside a stability tube around the manifold. Experiments on time series forecasting tasks support this view, showing that the proposed regularization improves predictive accuracy. More broadly, our results suggest a training principle for time-series forecasters and sequential predictors: a model should not only make small one-step errors, but also learn not to let its own errors echo into the future.
Code is available at \url{https://github.com/Wangzzzzzzzz/Echoes-Of-Error}.
\end{abstract}

\begin{flushright}
\small\itshape
A little fire is quickly trodden out, which, being suffered, rivers cannot quench.\\[0.4em]
\normalfont---William Shakespeare, \textit{Henry VI, Part 3}
\end{flushright}
\vspace{0.3em}

\section{Introduction}
\label{sec:introduction}

Time-series forecasting is a fundamental problem across high-stakes domains, including energy, climate, finance, transportation, and scientific discovery \citep{kim2025comprehensive, kong2025deep}. The long-horizon setting is especially challenging: from a finite history, a model must extrapolate far into the future while remaining robust to noise, nonstationarity, and long-range temporal dependencies \cite{zhou2021informer}.

Much of the recent progress has come from \emph{architectural} design. Existing models encode inductive biases for robust long-horizon extrapolation through explicit decomposition \citep{zhang2022lightts,chen2023tsmixer,wang2024timemixer,wu2021autoformer}, frequency-aware modules \citep{zhou2022fedformer,xu2024fits,yi2024filternet,eldele2024tslanet}, patching \cite{nie2023patchtst}, smoothing \cite{woo2022etsformer}, sparse or lightweight mixing\cite{lin2024sparsetsf, liu2024itransformer,zeng2023transformers}, and period-based representations \cite{dai2024pdf, wu2023timesnet}. These designs have produced strong and efficient forecasters. However, because robustness is often intertwined with the chosen parameterization, it remains difficult to isolate \emph{why} a model mitigates error propagation, which component is essential, or whether the same effect can transfer to a different model class. 

We instead take a data-driven perspective, designing loss functions that enforce forecasting robustness. In standard forecasting pipelines, a long time series is transformed into training pairs via sliding windows: the input history, the observed future target, and the model forecast are all segments of the same underlying series. Standard mean-squared-error training asks the forecast to match the immediate target, but it does not enforce this fundamental property of time series data. If the model has learned the dynamics correctly, \textbf{these blocks should be consistent with one another under the same physical or statistical law}. Rather than hard-coding this law into a specific architecture, we enforce the resulting consistency during training: when a forecast is treated as a local continuation of the current trajectory, the model's subsequent prediction should match the prediction obtained when conditioning on the observed future states.

This idea leads to a local consistency objective. Conceptually, for each training window, we compare two candidate next windows: a teacher-forced \citep{williams1989learning} window obtained by shifting in the observed future and a free-running \citep{williams1989learning} window obtained by shifting in the model's own forecast. The ideal consistency functional penalizes the discrepancy between the subsequent predictions induced by these two windows, thereby measuring whether the forecasted block remains compatible with the local trajectory. However, this exact construction is mainly a conceptual target: building a full next window can be ambiguous when history and forecast lengths differ, and fixed insertion creates \emph{positional bias} by probing residual sensitivity at only fixed temporal locations.\footnote{Under a fixed sliding-window update, copied history coordinates are identical under teacher forcing and free running. Their difference is nonzero only where the observed or predicted block is inserted, so consistency discrepancy probes residual sensitivity only at that location. Randomized insertion averages this probe across admissible locations; see Section~\ref{subsec:linear-nonlinear-reduction}.}

To obtain the practical objective, we replace this exact comparison with a first-order residual-sensitivity surrogate. We randomly insert the model residual into the history window and penalize the resulting Jacobian--vector response (JVP) of the forecaster. Random insertion removes the fixed-location bias, while the JVP penalty directly suppresses directions through which the model's own errors would perturb later predictions. The method requires no architectural change and importantly, keeps inference \emph{one-shot}, as in standard long-term forecasting benchmarks.

We also provide a geometric interpretation of when this JVP objective can improve free-running robustness. Under a predictive-manifold view \citep{bengio2013representation,fefferman2016testing}, valid history and future blocks lie near a low-dimensional trajectory manifold, while noise, model bias, and shocks may push observations away from it. The model's residual directions then act as local probes of off-manifold drift. When these probes cover the relevant normal directions and the learned dynamics are accurate on the manifold, suppressing the corresponding Jacobian--vector responses reduces transverse sensitivity and helps forecasts stay inside a predictive tube. Our contributions are threefold:

\begin{figure*}[t]
  \centering
    \begin{minipage}[c]{\linewidth}
    \centering  \includegraphics[width=\linewidth]{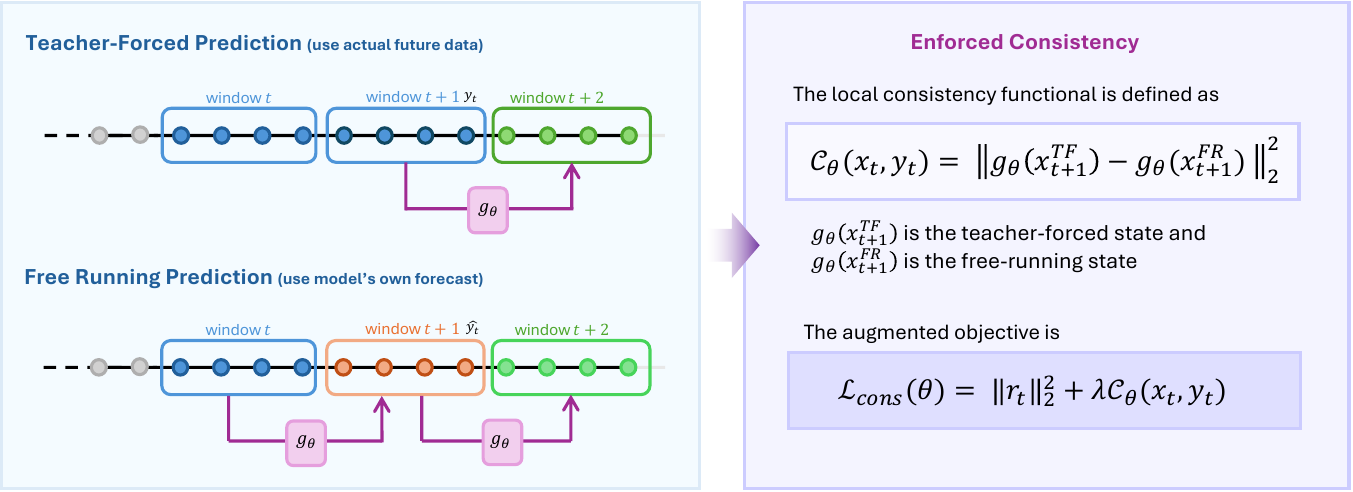}
  \end{minipage}
  \caption{Local rollout consistency. Given a history window $x_t$, teacher forcing constructs the next state using the observed target $y_t$, while free running constructs it using the model forecast $\hat{y}_t$. The ideal consistency loss penalizes the discrepancy between the next predictions from these two states; the practical objective uses a first-order residual-direction surrogate.}
    \label{fig:illustration-main-idea}
\end{figure*}

\begin{itemize}[leftmargin=15pt]
  \item \textbf{Local Rollout-Consistency Perspective:} We formulate long-horizon forecasting as a consistency problem induced by overlapping sliding windows, where a valid forecast should produce subsequent predictions compatible with those obtained from the observed continuation.
  \item \textbf{Residual-Direction First-Order Objective:} We derive a practical JVP-based penalty along randomly inserted residual directions, showing that it approximates the leading-order consistency functional while avoiding full next-window construction and fixed-location bias.
  \item \textbf{Geometric Interpretation and Empirical Validation:} We provide a predictive-manifold interpretation showing when residual-direction control can suppress off-manifold sensitivity, and empirically show that the objective makes a simple residual MLP highly competitive while further improving several strong forecasting backbones.
\end{itemize}

\section{A Local Rollout Consistency Objective}
\label{sec:rollout-consistency}

We now turn the idea from the introduction into a concrete training objective. Beyond matching the immediate target, a forecast should remain compatible with the local trajectory when it is fed back into the model. Since time-series training windows overlap, the next prediction obtained after feeding back the model forecast should be close to the next prediction obtained after feeding back the true future block.

To formalize this requirement, we first introduce an idealized teacher-forced/free-running discrepancy between two nearby rollouts: one updates the window using the actual observed future block, and the other updates it using the model's forecast as a proxy. We then show that this discrepancy admits a local approximation as a residual-sensitivity penalty, leading to a practical objective that leaves both the forecasting architecture and the standard one-shot inference protocol unchanged.

\subsection{Teacher Forcing, Free Running, and an Exact Consistency Objective}
\label{subsec:tf-fr-consistency}

\begin{figure}[h]
\centering
\begin{minipage}{0.49\linewidth}
    \centering
    \includegraphics[width=\linewidth]{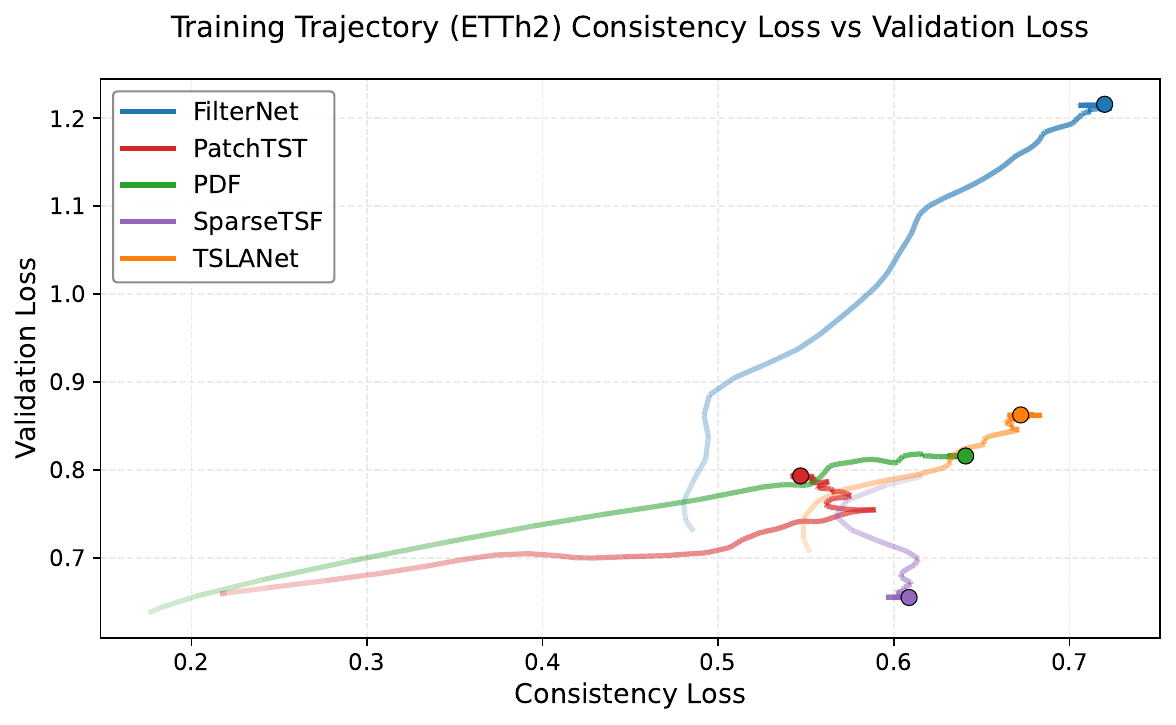}
\end{minipage}
\hfill
\begin{minipage}{0.49\linewidth}
    \centering
    \includegraphics[width=\linewidth]{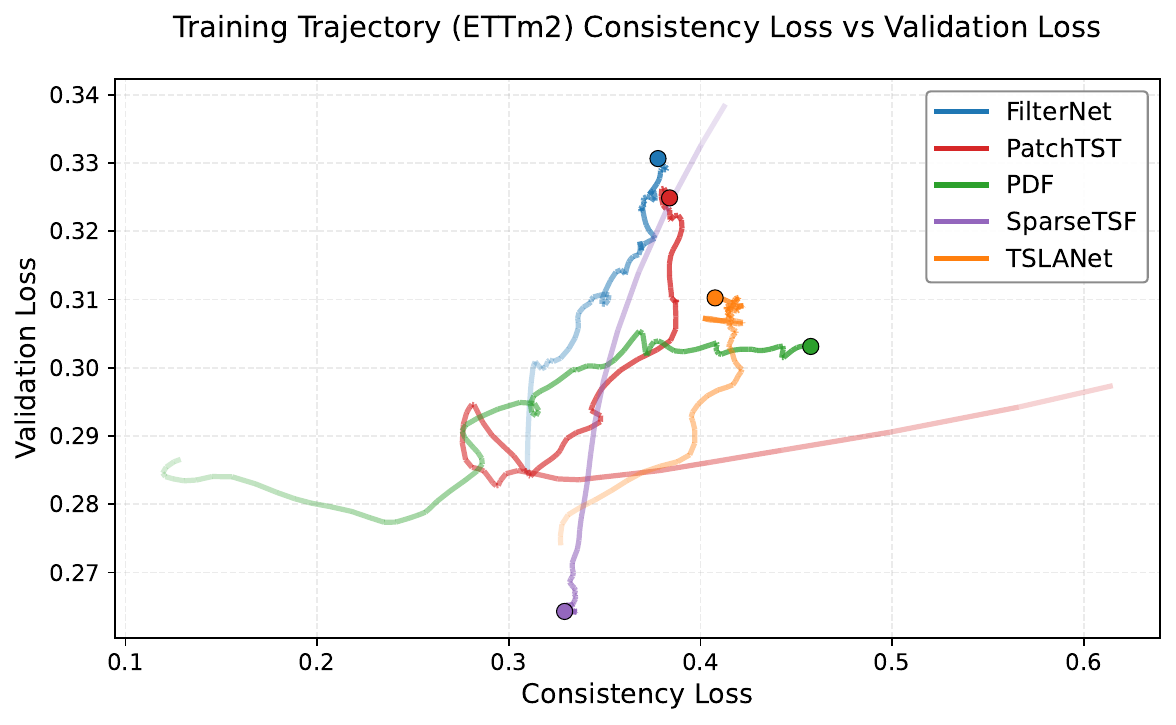}
\end{minipage}
\caption{Trajectory consistency versus validation error on ETTh2 and ETTm2. Point transparency encodes training time, with lighter points corresponding to earlier steps and darker points to later ones; circular markers denote where the model converges. Across training, smaller consistency gaps tend to coincide with smaller validation error, suggesting that teacher-forcing/free-running consistency captures a quantity relevant to generalization rather than a purely intuitive diagnostic.}
\label{fig:traj-consistency-tracks-val}
\end{figure}

We begin with the idealized setting in which a next history window can be constructed from either the observed future block or the predicted future block. This lets us define two comparable rollouts. Following the classical terminology from autoregressive learning \cite{williams1989learning,bengio2015scheduled, lamb2016professor}, \emph{teacher forcing} updates the next input with the ground-truth continuation, whereas \emph{free running} updates it with the model's own prediction. The forecaster itself remains a standard one-shot model: in both cases, the same map \(g_\theta\) is applied to a history window to produce a future block.

Specifically, let \(x_t \in \mathcal X\) denote a history state and let \(y_t \in \mathcal Y\) denote the corresponding future target at time step $t$. A one-shot block forecaster
\[
g_\theta : \mathcal X \to \mathcal Y
\]
produces
\[
\hat y_t := g_\theta(x_t),
\qquad
r_t := y_t - \hat y_t.
\]

Let
\(
S : \mathcal X \times \mathcal Y \to \mathcal X
\)
be the alignment operator that constructs the $t+1$-step history state from the $t$-step future.\footnote{This idealized operator is used only to define the teacher-forced/free-running discrepancy. Section~\ref{subsec:linear-nonlinear-reduction} explains why the practical objective replaces this global alignment map with a local residual-insertion surrogate.} The two candidate next states are
\begin{equation}
\label{eq:tf-fr-next-states}
x_{t+1}^{\mathrm{TF}} := S(x_t,y_t),
\qquad
x_{t+1}^{\mathrm{FR}} := S(x_t, \hat y_t).
\end{equation}

The free-running window $x_{t+1}^{\mathrm{FR}}$ should behave like a valid next window from the same time series, not like an off-trajectory input. We therefore compare the next prediction from this forecast-based window with the next prediction from the teacher-forced window, which motivates the exact local consistency objective $\mathcal C_\theta(x_t,y_t)$:
\begin{equation}
\label{eq:exact-consistency}
\mathcal C_\theta(x_t,y_t)
:=
\left\|
g_\theta(x_{t+1}^{\mathrm{TF}})
-
g_\theta(x_{t+1}^{\mathrm{FR}})
\right\|_2^2.
\end{equation}

Empirical observations in Figure~\ref{fig:traj-consistency-tracks-val} show that the consistency gap and validation error ``move together'': when one rises (e.g.\ due to overfitting), the other tends to rise as well, and when one falls, the other follows. In other words, the two quantities trace a roughly \emph{positively correlated} curve rather than diverging, suggesting that the consistency gap tracks generalization performance throughout training.
We therefore use the consistency-augmented objective as the conceptual target:
\begin{equation}
\label{eq:exact-consistency-loss}
\mathcal L_{\mathrm{cons}}(\theta)
=
\|r_t\|_2^2
+
\lambda\,\mathcal C_\theta(x_t,y_t).
\end{equation}

\subsection{From Global Alignment to Local Residual Sensitivity}
\label{subsec:linear-nonlinear-reduction}

\paragraph{Local Approximation of the Consistency Loss.}
Constructing a globally valid alignment operator is often ambiguous when the forecast block is shorter than the history window. Naively appending the forecast to the
end of the window fails to test consistency for certain fixed positions. Rather than resolving this global alignment problem, we derive a local approximation that depends only on the residual direction. The teacher-forced and free-running windows differ only because the second argument of \(S\) changes from \(\hat y_t\) to \(y_t=\hat y_t+r_t\). We denote the resulting next-state perturbation by
\begin{equation}
\label{eq:delta-next-state}
\delta_t
:=
x_{t+1}^{\mathrm{TF}}-x_{t+1}^{\mathrm{FR}}
=
S(x_t,y_t)-S(x_t,\hat y_t).
\end{equation}
With this notation, the exact consistency loss can be viewed as the model's response to a perturbation of its next input: it compares the prediction at \(x_{t+1}^{\mathrm{FR}}\) with the prediction at \(x_{t+1}^{\mathrm{FR}}+\delta_t=x_{t+1}^{\mathrm{TF}}\). Thus the consistency gap is controlled by the directional sensitivity of \(g_\theta\) along \(\delta_t\). The following proposition makes this connection precise: for linear models the expression is exact, while for nonlinear models it is captured to leading order by a Jacobian-vector product.

\begin{proposition}[Local consistency as directional sensitivity]
\label{thm:linear-nonlinear-consistency}
Let \(\delta_t=x_{t+1}^{\mathrm{TF}}-x_{t+1}^{\mathrm{FR}}\) be the next-state perturbation defined in Equation~\eqref{eq:delta-next-state}, and consider the exact consistency functional \(\mathcal C_\theta(x_t,y_t)\) in Equation~\eqref{eq:exact-consistency}. Then the following statements hold.

\begin{enumerate}
    \item[\textup{(i)}] \textbf{Linear case.}
    If \(g_\theta(x)=Gx\) is linear, then
    \begin{equation}
    \label{eq:linear-exact-consistency}
    \mathcal C_\theta(x_t,y_t)
    =
    \|G\delta_t\|_2^2.
    \end{equation}

    \item[\textup{(ii)}] \textbf{Nonlinear case.}
    Suppose \(g_\theta\) is \(C^2\) in a neighborhood of \(x_{t+1}^{\mathrm{FR}}\). Then, as \(\|\delta_t\|_2 \to 0\),
    \begin{equation}
    \label{eq:nonlinear-first-order-consistency}
    \mathcal C_\theta(x_t,y_t)
    =
    \left\|
    Jg_\theta(x_{t+1}^{\mathrm{FR}})\,\delta_t
    \right\|_2^2
    +
    O(\|\delta_t\|_2^3),
    \end{equation}
    where \(\delta_t=x_{t+1}^{\mathrm{TF}}-x_{t+1}^{\mathrm{FR}}\), and where \(Jg_\theta(x)\) denotes the Jacobian of \(g_\theta\) at \(x\). 
\end{enumerate}
\end{proposition}

Proposition~\ref{thm:linear-nonlinear-consistency} shows that consistency can be locally approximated by a sensitivity penalty: it penalizes directions \(\delta_t\) in the input space that strongly change the next prediction. The remaining question is how to represent \(\delta_t\) without constructing the full next window. The next remark answers this by linearizing the update map \(S\) with respect to the forecast block.

\begin{remark}[Residual insertion as a local surrogate]
\label{rmk:residual-domain-alignment}
Fix \(x_t\), define \(S_t(z):=S(x_t,z)\), and recall \(r_t:=y_t-\hat y_t\). Assume that \(S_t \in C^2\) in a neighborhood of \(\hat y_t\). Then
\begin{equation}
\label{eq:state-linearization}
\delta_t
=
S_t(y_t)-S_t(\hat y_t)
=
JS_t(\hat y_t)\,r_t
+
O(\|r_t\|_2^2).
\end{equation}

Moreover, suppose that changing the forecast block only changes the coordinates where that block is inserted, while all other coordinates are filled independently of the forecast value. Then the local derivative of \(S_t\) is simply a padding insertion operator \(P:\mathcal Y\to\mathcal X\), so
\[
JS_t(\hat y_t)=P.
\]
If the insertion rule is affine in the forecast block, then this approximation becomes exact: \(\delta_t=Pr_t\).
\end{remark}

Remark~\ref{rmk:residual-domain-alignment} justifies the practical surrogate used below. We do not need to build the entire padded window; we only need to insert the residual direction \(r_t\) into the chosen coordinates and fill all other coordinates with zeros. Randomizing the insertion position then probes the model's sensitivity across the window, which removes the fixed-location bias without requiring a globally valid random-padding construction.

\paragraph{Practical First-Order Consistency Objective.} Motivated by Proposition~\ref{thm:linear-nonlinear-consistency} and Remark~\ref{rmk:residual-domain-alignment}, we sample one insertion position $\omega_t\sim q$ per example as an unbiased Monte Carlo estimate \citep{metropolis1949monte} of the averaged consistency penalty over insertion positions. 

We therefore arrive at the following practical objective without positional bias:
\begin{methodbox}[title={Practical propagation-control objective}]
\begin{equation}
\label{eq:first-order-consistency-loss-readable}
\mathcal L_{\mathrm{fo}}(\theta)
=
\underbrace{\|r_t\|_2^2}_{\text{fit the current target}}
+
\lambda\,
\underbrace{\left\|
Jg_\theta(x_t)\,\mathrm{sg}[P_{\omega_t} r_t]
\right\|_2^2}_{\text{suppress response to the residual}},
\qquad \omega_t\sim q,
\end{equation}
\end{methodbox}
Here $q$ is the sampling distribution over insertion positions, and $P_{\omega_t}:\mathcal Y\to\mathcal X$ is the corresponding zero-padding insertion operator: it places the residual block $r_t$ at position $\omega_t$ and fills all other state coordinates with zeros. 
$x_t$ is the input history state and \(\mathrm{sg}[\cdot]\) denotes stop-gradient. 
The second term is implemented with a Jacobian-vector product, without materializing the full Jacobian. We use full-batch JVPs in the main-text experiments to evaluate the effectiveness of the proposed consistency objective. Appendix~\ref{app:exp-compute} analyzes the computational cost and shows that, in compute-constrained settings, batch subsampling can reduce training overhead while largely preserving forecasting accuracy.
Using the Jacobian at $x_t$ should be interpreted as a directional JVP approximation, not as an assumption that the full Jacobian is unchanged.\footnote{The approximation only requires the sampled directional responses to be close, namely $\|Jg_\theta(x_t)P_{\omega_t}r_t\|_2^2\approx\|Jg_\theta(x_{t+1}^{\mathrm{FR}})P_{\omega_t}r_t\|_2^2$. Equivalently, under a locally stationary residual distribution, $P_{\omega_t}r_t$ can be viewed as a resampled residual-direction probe at $x_t$. The approximation is exact for linear models, where $J(x)=G$ for all $x\in\mathcal X$.}
Appendix~\ref{app:exp-cl-jvp-alignment} empirically evaluates how well this practical JVP approximation aligns with the exact consistency objective.

\paragraph{Why Not Use a Global Alignment Operator?} The construction above deliberately avoids committing to a full next-window alignment operator. The exact consistency loss in Equation~\eqref{eq:exact-consistency} assumes that one can construct a complete next history window from either the true future $y_t$ or the forecast $\hat y_t$. This is unambiguous when the history and future blocks are already aligned. In many forecasting settings, however, the future block is shorter than the history window, so a global operator must specify both where to place the forecast and how to fill the remaining coordinates.

A deterministic choice, such as the standard sliding-window update that appends the forecast to the end of the window \citep{taieb2012review}, can introduce \emph{positional bias}: most coordinates are copied from the current history and are identical under teacher forcing and free running, so the discrepancy only probes the model's response to the residual at one fixed location. Removing this bias at the level of full windows would require randomly placing the forecast block while filling all other coordinates with valid values from the same time series, such as additional future observations, covariates, or time features. These quantities are not determined by the local pair $(x_t,y_t)$. The residual-insertion objective therefore keeps only the local direction that distinguishes teacher forcing from free running, and randomizes its insertion position without constructing a complete, globally valid next-step history window for each random placement. This interpretation is formalized in Appendix~\ref{app:positional-bias} and is empirically supported in Appendix~\ref{app:exp-pos-bias}, where fixed-end insertion tends to underperform randomized insertion.

\section{Geometric Consequences of Consistency Loss}
\label{sec:manifold-aware-consistency}

Following the intuition from the previous section, a forecast should not only match the next target value, but should also remain a plausible $t+1$-step history block when fed back into the model. We now interpret this requirement geometrically. Under the manifold assumption \citep{bengio2013representation,fefferman2016testing}, clean time-series segments generated by the underlying dynamics concentrate near a low-dimensional predictive manifold \(\mathcal M \subset \mathcal X\). In practice, however, observations are corrupted by measurement noise, sensor errors, model bias, and unexpected exogenous shocks \citep{tsay1988outliers,box1975intervention,kailath1970innovations}. These perturbations can push the observed time series segments away from the manifold, even when the one-step prediction error is small. 
A robust forecasting model should therefore not merely minimize pointwise error. It should also filter out unpredictable perturbations and produce forecasts that remain close to the predictive manifold.

\begin{figure}[ht]
\centering
    \includegraphics[width=\linewidth]{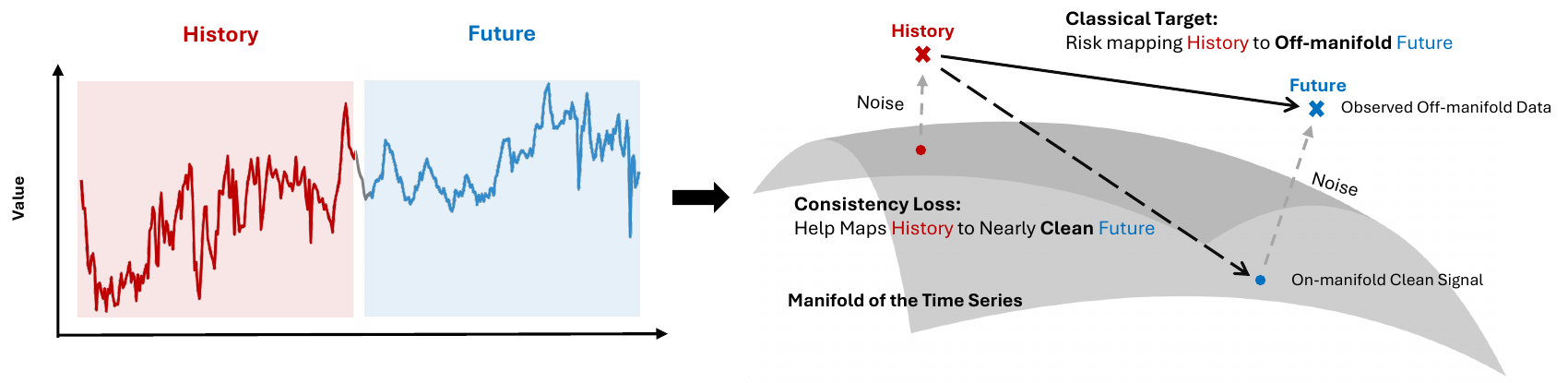}
\caption{Geometric intuition behind Theorem~\ref{thm:manifold-global-informal}. Perturbations can push observations away from the predictive manifold and a plain MSE minimizer may overfit such perturbations; consistency loss reduces the model's sensitivity to these off-manifold deviations, helping forecasted time series to remain close to the data-supported dynamics under suitable conditions.}
\label{fig:manifold-consistency}
\end{figure}

In this section, we provide a conditional geometric mechanism showing how the consistency loss in \eqref{eq:first-order-consistency-loss-readable} can act as a manifold-aware regularizer: under suitable residual excitation and approximate on-manifold correctness, it suppresses off-manifold response and helps free-running rollouts remain near the predictive manifold. Define the free-running rollout map by
\begin{equation}
\label{eq:full-residual-rollout-map}
\Phi_\theta(x):=S(x,g_\theta(x)).
\end{equation}
Let \(\mathcal M\subset\mathcal X\) be a compact predictive manifold, let \(\pi\) denote the nearest-point projection on a tubular neighborhood of \(\mathcal M\), and define the normal projector by
\begin{equation}
\label{eq:normal-projector-full-residual}
\Pi^\perp(x):=\Pi_{N_{\pi(x)}\mathcal M}.
\end{equation}
We have the following geometric interpretation for the rollout map trained with $\mathcal{L}_{\text{fo}}(\theta)$:

\begin{theorem}[Conditional Predictive-Tube Stability under Residual-Direction Normal Excitation]
\label{thm:manifold-global-informal}
Assume that \(\mathcal M\) is approximately invariant under the free-running rollout map \(\Phi_\theta\), that the residual-induced directions \(\delta_t\) sufficiently excite the normal directions of \(\mathcal M\) on visited states, that the penalty in \eqref{eq:first-order-consistency-loss-readable} keeps the expected directional response \(\mathbb E[\|Jg_\theta(x_{t+1}^{\mathrm{FR}})\delta_t\|_2^2\mid x_{t+1}^{\mathrm{FR}}]\) small, and that smoothness and coverage extend this local control across a predictive tube \(U_\rho(\mathcal M)\). Then the same training loss implies a uniform bound on the off-manifold response,
\begin{equation}
\label{eq:informal-normal-operator-control}
\sup_{x\in U_\rho(\mathcal M)}
\|Jg_\theta(x)\Pi^\perp(x)\|_F
\le
\gamma_N,
\end{equation}
which in turn yields a transverse recursion of the form
\begin{equation}
\label{eq:informal-local-recursion}
d(\Phi_\theta(x),\mathcal M)
\le
\varepsilon_{\mathcal M}
+
\kappa_N d(x,\mathcal M)
+
C_\rho d(x,\mathcal M)^2.
\end{equation}
If additionally
\[
\bar\kappa:=\kappa_N+C_\rho\rho<1,
\qquad
\varepsilon_{\mathcal M}\le (1-\bar\kappa)\rho,
\]
then every rollout initialized in \(U_\rho(\mathcal M)\) remains inside that tube and satisfies
\begin{equation}
\label{eq:informal-global-tube}
d(x_k,\mathcal M)
\le
\bar\kappa^k d(x_0,\mathcal M)
+
\frac{1-\bar\kappa^k}{1-\bar\kappa}\,\varepsilon_{\mathcal M},
\qquad k\ge 0.
\end{equation}
\end{theorem}

Intuitively, the residual directions $\delta_t$ provide reasonable samples of the deviations that make observed histories off-manifold. Consistency loss trains the model to be insensitive to these deviations. The theorem formalizes this intuition under a predictive-manifold model: when the residual directions cover the relevant off-manifold directions and the learned dynamics are accurate on the manifold, CL controls the model's off-manifold response. With small one-step drift and a mild contraction condition, this local control keeps the whole rollout inside a stable tube around the predictive manifold. Appendix~\ref{app:manifold-global-theorem} gives the full statement with assumptions and constants. Table~\ref{tab:rollout-vs-oneblock} further supports this stability interpretation through rollout experiments with the proposed method.

\section{Experiments}
\label{sec:exp}

\begin{table}[t]
\caption{Multivariate forecasting results for horizons $H \in \{96, 192, 336, 720\}$ with a lookback window of length $L = 720$. Each model directly predicts the entire $H$-step target in one shot, without recursively feeding predictions back. We apply consistency loss (CL) to a residual MLP block. Best results are highlighted in \best{red}, and second-best results are highlighted in \secbest{blue}.}
\label{table:main_res}
\centering
\scriptsize
\setlength{\tabcolsep}{4pt}
\scalebox{0.92}{
\begin{tabular}{cc|ccccccccc|c}
\csname toprule\endcsname
Dataset & $H$ & PatchTST & FilterNet & SparseTSF & TSLANet & ~~~PDF~~~ & ~~~AMD~~~ & ~~DUET~~ & TimeDistill & RootPurge & \textbf{MLP + CL} \\ \midrule
\multirow{4}{*}{\rotatebox{90}{ETTh1}} & 96  & 0.385 & 0.375 & 0.362 & 0.387 & 0.364 & 0.362 & 0.369 & 0.373 & \secbest{0.362} & \best{0.358} \\
 & 192 & 0.413 & 0.436 & 0.403 & 0.425 & 0.399 & 0.401 & 0.401 & 0.411 & \secbest{0.398} & \best{0.395} \\
 & 336 & 0.440 & 0.472 & 0.435 & 0.443 & 0.436 & \secbest{0.418} & \best{0.417} & 0.439 & 0.434 & 0.434 \\
 & 720 & 0.456 & 0.474 & \secbest{0.426} & 0.463 & 0.487 & 0.439 & 0.445 & 0.495 & \best{0.423} & 0.458 \\ \midrule
\multirow{4}{*}{\rotatebox{90}{ETTh2}} & 96  & 0.274 & 0.301 & 0.294 & 0.297 & 0.281 & 0.274 & 0.276 & 0.273 & \secbest{0.272} & \best{0.269} \\
 & 192 & 0.338 & 0.363 & 0.339 & 0.348 & 0.351 & 0.351 & 0.334 & 0.334 & \secbest{0.332} & \best{0.331} \\
 & 336 & 0.367 & 0.401 & 0.359 & 0.375 & 0.381 & 0.375 & 0.367 & 0.363 & \best{0.359} & \secbest{0.362} \\
 & 720 & 0.391 & 0.438 & \secbest{0.383} & 0.431 & 0.412 & 0.389 & 0.390 & 0.408 & \best{0.381} & 0.397 \\ \midrule
\multirow{4}{*}{\rotatebox{90}{ETTm1}} & 96  & 0.292 & 0.318 & 0.314 & 0.296 & 0.295 & 0.284 & \secbest{0.284} & 0.285 & 0.305 & \best{0.282} \\
 & 192 & 0.330 & 0.364 & 0.343 & 0.332 & 0.325 & \secbest{0.322} & 0.324 & 0.331 & 0.335 & \best{0.317} \\
 & 336 & 0.365 & 0.396 & 0.369 & 0.371 & 0.358 & 0.360 & \secbest{0.357} & 0.359 & 0.361 & \best{0.352} \\
 & 720 & 0.419 & 0.456 & 0.418 & 0.426 & 0.417 & 0.421 & \best{0.408} & 0.415 & 0.415 & \secbest{0.410} \\ \midrule
\multirow{4}{*}{\rotatebox{90}{ETTm2}} & 96  & 0.163 & 0.178 & 0.165 & 0.196 & \secbest{0.160} & 0.167 & 0.162 & 0.163 & 0.161 & \best{0.158} \\
 & 192 & 0.220 & 0.243 & 0.218 & 0.251 & 0.222 & 0.221 & 0.221 & 0.220 & \secbest{0.217} & \best{0.215} \\
 & 336 & 0.276 & 0.293 & 0.272 & 0.299 & 0.275 & 0.270 & 0.271 & 0.269 & \secbest{0.269} & \best{0.268} \\
 & 720 & 0.368 & 0.369 & 0.350 & 0.369 & 0.351 & 0.356 & 0.350 & \best{0.326} & 0.353 & \secbest{0.348} \\ \midrule
\multirow{4}{*}{\rotatebox{90}{Weather}} & 96  & 0.151 & 0.164 & 0.172 & 0.148 & 0.149 & 0.146 & 0.147 & 0.145 & \secbest{0.143} & \best{0.142} \\
 & 192 & 0.195 & 0.214 & 0.215 & 0.193 & 0.193 & 0.188 & 0.188 & 0.188 & \secbest{0.186} & \best{0.185} \\
 & 336 & 0.249 & 0.268 & 0.260 & 0.245 & 0.244 & 0.240 & 0.239 & 0.238 & \secbest{0.236} & \best{0.235} \\
 & 720 & 0.321 & 0.344 & 0.318 & 0.325 & 0.313 & 0.315 & 0.319 & 0.312 & \best{0.308} & \secbest{0.312} \\ \midrule
\multirow{4}{*}{\rotatebox{90}{ECL}} & 96  & 0.129 & 0.137 & 0.138 & 0.129 & 0.127 & 0.129 & \secbest{0.128} & 0.128 & 0.133 & \best{0.127} \\
 & 192 & 0.149 & 0.156 & 0.151 & 0.147 & 0.146 & 0.147 & \secbest{0.145} & 0.145 & 0.148 & \best{0.144} \\
 & 336 & 0.166 & 0.171 & 0.166 & 0.163 & 0.161 & 0.161 & 0.163 & \secbest{0.161} & 0.165 & \best{0.160} \\
 & 720 & 0.210 & 0.207 & 0.205 & 0.197 & 0.196 & \secbest{0.193} & \best{0.193} & 0.195 & 0.205 & 0.198 \\ \midrule
\multirow{4}{*}{\rotatebox{90}{Traffic}} & 96  & 0.366 & 0.393 & 0.389 & 0.362 & 0.364 & 0.366 & 0.360 & \secbest{0.358} & 0.385 & \best{0.358} \\
 & 192 & 0.388 & 0.408 & 0.398 & 0.377 & 0.381 & 0.381 & 0.383 & \secbest{0.375} & 0.397 & \best{0.375} \\
 & 336 & 0.398 & 0.424 & 0.411 & 0.391 & 0.402 & 0.397 & 0.395 & \secbest{0.389} & 0.411 & \best{0.388} \\
 & 720 & 0.457 & 0.462 & 0.448 & 0.431 & 0.460 & 0.429 & 0.435 & \secbest{0.429} & 0.449 & \best{0.428} \\ \midrule
\multicolumn{2}{c|}{Number of $1^{\text{st}}$ Places}  & 0 & 0 & 0 & 0 & 0 & 0 & 3 & 1 & 4 & 20 \\ \bottomrule
\end{tabular}
}
\end{table}
\vspace{-1pt}

\subsection{Experimental Setup}
\label{subsec:exp-setup}

\textbf{Datasets.}
We evaluate on standard long-term time-series forecasting benchmarks from traffic, energy, and weather domains. Specifically, we use the ETT benchmarks introduced by Informer \citep{zhou2021informer}, together with Traffic, Electricity (ECL), and Weather, following the evaluation protocol commonly adopted in long-horizon forecasting \citep{wu2021autoformer,nie2023patchtst,liu2024itransformer}. 

\textbf{Baselines.}
We compare against representative recent forecasting methods, including PatchTST \citep{nie2023patchtst}, DUET \citep{qiu2024duet}, AMD \citep{hu2025adaptive}, PDF \citep{dai2024pdf}, SparseTSF \citep{lin2024sparsetsf}, FilterNet \citep{yi2024filternet}, and TSLANet \citep{eldele2024tslanet}. We also include RootPurge \citep{wang2025characteristic}, a characteristic-root regularizer for linear forecasters, and TimeDistill \citep{ni2026timedistill}, a distillation method for MLP forecasters. Together, these baselines cover transformer, frequency-domain, lightweight linear/sparse, clustering-based, regularization, and distillation approaches.

\textbf{Implementation.} We evaluate the proposed consistency loss using a residual MLP block with three layers, 2048 hidden dimensions, and ReLU activations. The consistency loss is applied branch-wise: for the MLP branch, we search over $\lambda \in \{0, 0.5, 2, 8, 32, 128\}$, while for the linear shortcut, we fix $\lambda=0.5$. Following \citep{lin2024sparsetsf, wang2025characteristic}, we use a lookback window of length 720 and report MSE as the evaluation metric.
Additional implementation details are provided in Appendix~\ref{app:exp_details}.

\subsection{Main Results}
We list the comprehensive forecasting results in Table~\ref{table:main_res}, with the following key takeaways:

\textbf{Good Performance Can Be Achieved with a Simple Architecture.} MLP+CL achieves the best result on 20 out of 28 dataset--horizon settings, despite using only a plain residual MLP backbone. This result shows that forecasting accuracy can be improved substantially with a proper training strategy, without relying only on forecasting-specific architectural modules. In a controlled ablation on the same architecture, enabling CL on the nonlinear MLP branch lowers or retains forecasting error in all 28 settings. See Appendix~\ref{app:exp-mlp-zero} for details.

\textbf{Linear Models Win on Optimization, Not on Modeling Power.} RootPurge can be viewed as a special simplification of CL for linear forecasters, and its strong performance suggests that consistency-style regularization already helps in the linear regime. Our MLP+CL further improves over linear baselines on most settings, indicating that once optimization is properly regularized, nonlinear models can better capture the dynamics of real-world time series. Thus, the success of linear models observed in prior work \citep{zeng2023transformers} may stem more from easier optimization than from superior modeling power.

\subsection{Model Analyses}

We now examine four key aspects of the consistency loss in the main text: \emph{(i)} whether it generalizes across architectures, \emph{(ii)} whether it is robust to the choice of regularization weight $\lambda$, \emph{(iii)} why it helps, and \emph{(iv)} whether it improves free-running robustness. We further analyze the consistency penalty through padding ablations (Appendix~\ref{app:exp-padding}) and comparisons with isotropic Jacobian regularization (Appendix~\ref{app:exp-isotropic-cl}). These results support zero-padding for residual insertion and show that targeting the forecast-error direction yields better forecasting accuracy on average. We also study the possibility of reducing the computational cost of our method for resource-constrained settings in Appendix~\ref{app:exp-compute}.

\paragraph{Backbone Agnostic Property.} Because the consistency loss acts at the objective level, it is in principle independent of the forecasting architecture. To test this, we add it to four representative backbones spanning frequency-domain (FilterNet \citep{yi2024filternet}), convolutional (TSLANet \citep{eldele2024tslanet}), transformer (iTransformer \citep{liu2024itransformer}), and patch-based (PDF \citep{dai2024pdf}) designs. Table~\ref{tab:cl_results} reports the results. Adding CL improves all four backbones on most dataset--horizon settings. The objective is therefore not tied to our residual MLP, but provides a general regularizer that complements existing architectural choices. The fact that even already strong forecasters benefit further suggests that forecast-error propagation is a shared issue across model families, which CL helps mitigate. See Appendix~\ref{app:exp_details} for the experimental setups.

\begin{table*}[ht]
\centering
\caption{Comparison of baseline models with and without consistency loss (+CL) regularization on MSE and MAE metrics. Each model directly predicts the entire $H$-step target in one shot, without recursively feeding predictions back. CL improves forecasting performance in most settings across the four backbones and three datasets, supporting the backbone-agnostic property of the proposed objective. For each (Original, +CL) pair, the better value is shown in \textbf{bold}.}
\label{tab:cl_results}
\resizebox{\textwidth}{!}{%
\begin{tabular}{l l l |cccc|cccc|cccc}
\toprule
\multirow{2}{*}{\textbf{Model}} & \multirow{2}{*}{\textbf{Metric}} & \multirow{2}{*}{\textbf{Method}} & \multicolumn{4}{c}{ETTh2} & \multicolumn{4}{c}{ETTm2} & \multicolumn{4}{c}{ECL} \\
\cmidrule(lr){4-7}
\cmidrule(lr){8-11}
\cmidrule(lr){12-15}
 & &  & 96 & 192 & 336 & 720 & 96 & 192 & 336 & 720 & 96 & 192 & 336 & 720 \\
\midrule
\multirow{4}{*}{FilterNet} & \multirow{2}{*}{MSE} & Original & 0.312 & 0.363 & 0.402 & 0.438 & 0.178 & 0.243 & 0.293 & 0.370 & 0.137 & 0.155 & 0.170 & 0.206 \\
 &  & \textbf{+CL} & \textbf{0.278} & \textbf{0.341} & \textbf{0.382} & \textbf{0.428} & \textbf{0.164} & \textbf{0.224} & \textbf{0.273} & \textbf{0.352} & \textbf{0.137} & \textbf{0.153} & \textbf{0.169} & \textbf{0.206} \\
\cmidrule(lr){2-15}
 & \multirow{2}{*}{MAE} & Original & 0.365 & 0.398 & 0.431 & 0.468 & 0.265 & 0.309 & 0.345 & 0.396 & 0.236 & 0.253 & 0.269 & 0.302 \\
 &  & \textbf{+CL} & \textbf{0.340} & \textbf{0.383} & \textbf{0.420} & \textbf{0.464} & \textbf{0.255} & \textbf{0.298} & \textbf{0.332} & \textbf{0.383} & \textbf{0.236} & \textbf{0.252} & \textbf{0.269} & \textbf{0.301} \\
\midrule
\multirow{4}{*}{TSLANet} & \multirow{2}{*}{MSE} & Original & 0.297 & 0.348 & 0.376 & 0.431 & 0.197 & 0.251 & 0.302 & 0.369 & 0.129 & 0.147 & 0.163 & 0.196 \\
 &  & \textbf{+CL} & \textbf{0.270} & \textbf{0.332} & \textbf{0.364} & \textbf{0.411} & \textbf{0.190} & \textbf{0.239} & \textbf{0.282} & \textbf{0.358} & \textbf{0.128} & \textbf{0.144} & \textbf{0.160} & \textbf{0.194} \\
\cmidrule(lr){2-15}
 & \multirow{2}{*}{MAE} & Original & 0.360 & 0.396 & 0.420 & 0.462 & 0.283 & 0.319 & 0.355 & 0.399 & 0.224 & 0.241 & 0.259 & 0.288 \\
 &  & \textbf{+CL} & \textbf{0.335} & \textbf{0.380} & \textbf{0.410} & \textbf{0.456} & \textbf{0.277} & \textbf{0.309} & \textbf{0.339} & \textbf{0.393} & \textbf{0.224} & \textbf{0.240} & \textbf{0.256} & \textbf{0.287} \\
\midrule
\multirow{4}{*}{iTransformer} & \multirow{2}{*}{MSE} & Original & 0.308 & 0.387 & 0.419 & 0.452 & 0.171 & 0.234 & 0.286 & 0.367 & 0.136 & 0.157 & 0.169 & \textbf{0.190} \\
 &  & \textbf{+CL} & \textbf{0.284} & \textbf{0.351} & \textbf{0.391} & \textbf{0.432} & \textbf{0.164} & \textbf{0.224} & \textbf{0.278} & \textbf{0.355} & \textbf{0.132} & \textbf{0.153} & \textbf{0.167} & 0.191 \\
\cmidrule(lr){2-15}
 & \multirow{2}{*}{MAE} & Original & 0.374 & 0.427 & 0.455 & 0.495 & 0.265 & 0.310 & 0.347 & 0.402 & 0.237 & 0.257 & 0.272 & \textbf{0.291} \\
 &  & \textbf{+CL} & \textbf{0.354} & \textbf{0.398} & \textbf{0.431} & \textbf{0.476} & \textbf{0.258} & \textbf{0.302} & \textbf{0.340} & \textbf{0.393} & \textbf{0.231} & \textbf{0.254} & \textbf{0.270} & 0.293 \\
\midrule
\multirow{4}{*}{PDF} & \multirow{2}{*}{MSE} & Original & 0.285 & 0.358 & 0.384 & 0.418 & 0.162 & 0.229 & 0.284 & 0.350 & 0.128 & 0.145 & 0.161 & 0.194 \\
 &  & \textbf{+CL} & \textbf{0.272} & \textbf{0.336} & \textbf{0.371} & \textbf{0.414} & \textbf{0.160} & \textbf{0.217} & \textbf{0.269} & \textbf{0.346} & \textbf{0.126} & \textbf{0.143} & \textbf{0.159} & \textbf{0.191} \\
\cmidrule(lr){2-15}
 & \multirow{2}{*}{MAE} & Original & 0.342 & 0.390 & 0.417 & \textbf{0.453} & 0.250 & 0.295 & 0.334 & 0.380 & 0.224 & 0.241 & 0.258 & 0.291 \\
 &  & \textbf{+CL} & \textbf{0.336} & \textbf{0.380} & \textbf{0.411} & 0.454 & \textbf{0.249} & \textbf{0.289} & \textbf{0.325} & \textbf{0.378} & \textbf{0.223} & \textbf{0.240} & \textbf{0.257} & \textbf{0.289} \\
\bottomrule
\end{tabular}}
\end{table*}

\begin{figure}[ht]
    \centering
    \includegraphics[width=0.49\linewidth]{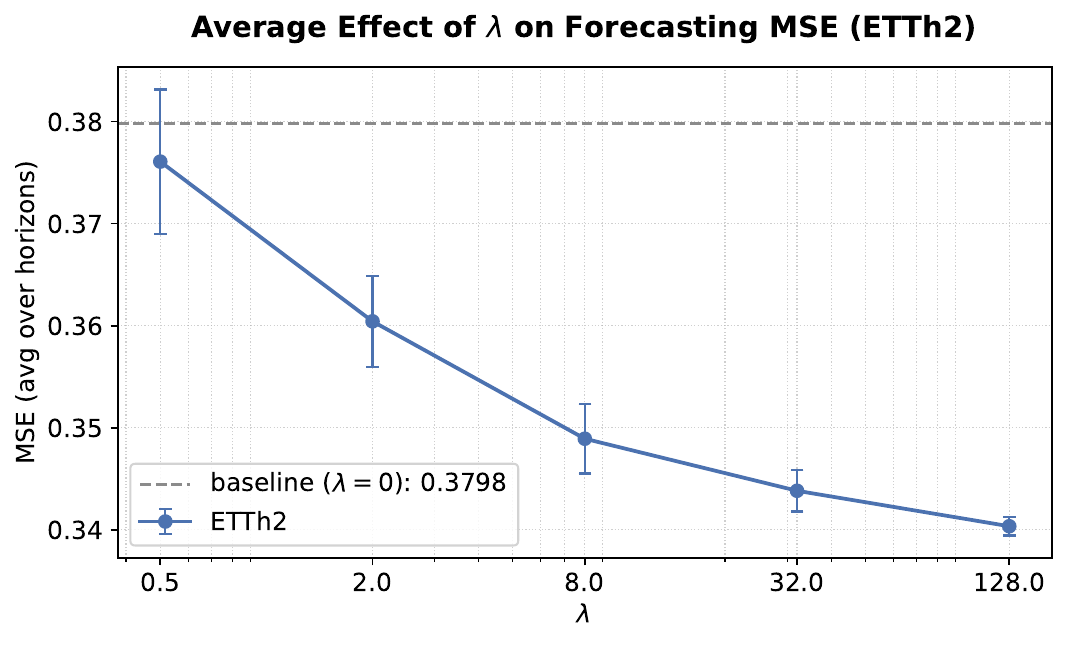}
    \includegraphics[width=0.49\linewidth]{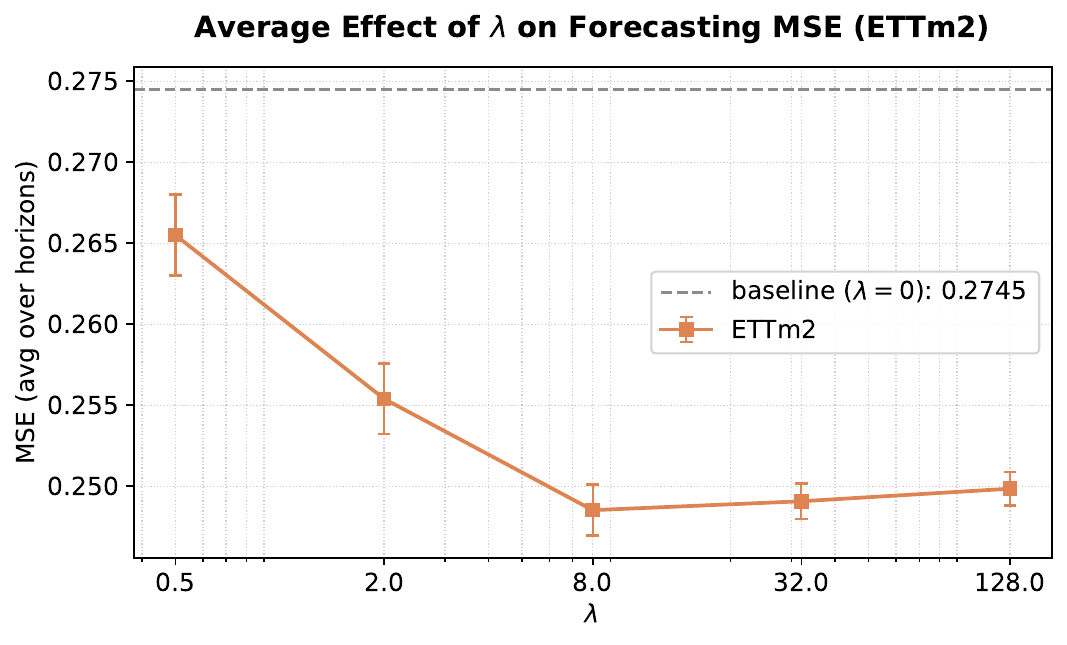}

    \caption{Test MSE versus consistency-loss coefficient $\lambda \in \{0, 0.5, 2, 8, 32, 128\}$ on ETTh2 (left) and ETTm2 (right), averaged over forecasting horizons.}
    \label{fig:lambda_ablation}
\end{figure}

\paragraph{Hyperparameter Effect.} A natural follow-up question is whether realizing this benefit requires careful tuning of $\lambda$. To answer it, we sweep $\lambda \in \{0, 0.5, 2, 8, 32, 128\}$ on ETTh2 and ETTm2 and report the test MSE in Figure~\ref{fig:lambda_ablation}. Three trends emerge: (i) any $\lambda > 0$ outperforms the MSE-only baseline ($\lambda = 0$); (ii) increasing $\lambda$ from $0$ initially improves test MSE, indicating stronger regularization aids generalization; and (iii) once $\lambda$ becomes too large, the gain saturates and eventually reverses, as excessive regularization prevents the model from fitting the data. CL therefore yields improvements over a wide range of $\lambda$, so practical gains are achievable without extensive tuning.

\begin{figure}[ht]
    \centering
    \includegraphics[width=0.98\linewidth]{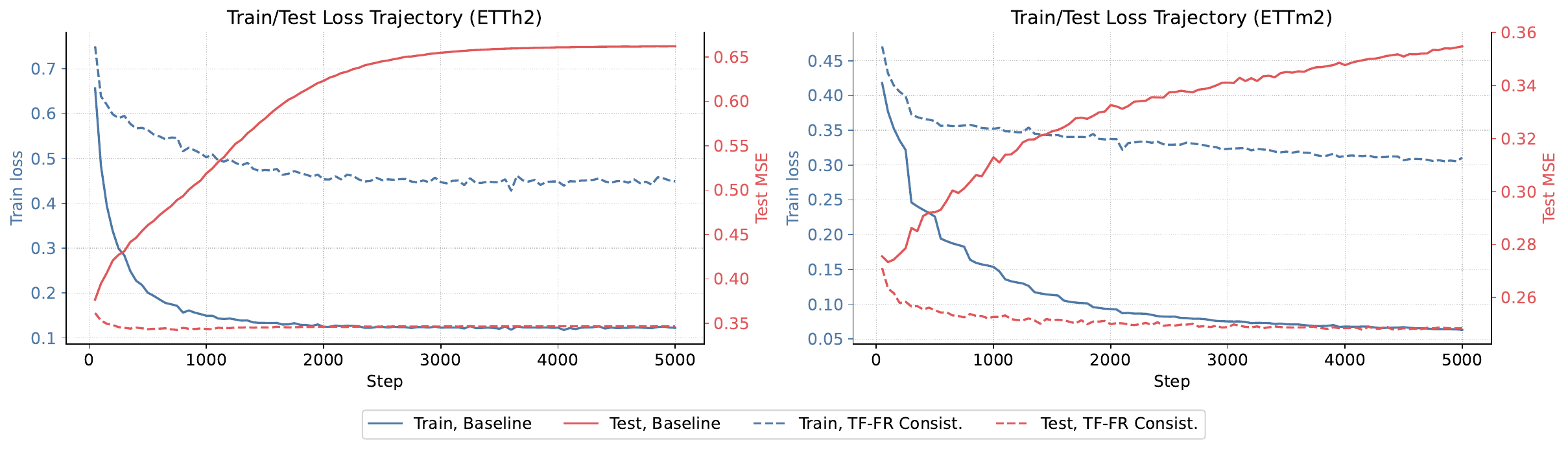}
    \caption{Training and test loss trajectories on ETTh2 and ETTm2. CL reduces the train--test gap and maintains decreasing test loss, whereas MSE-only training rapidly lowers the training loss but quickly overfits, leading to increased test loss.}
    \label{fig:train_test_curve}
\end{figure}

\paragraph{From Memorization to Dynamics Learning.} Having shown that CL helps broadly and robustly, we now turn to \emph{why} it helps. A common failure mode of MLP forecasters trained only with MSE is that they memorize off-manifold observations instead of learning the underlying dynamics; CL is designed to counter this by penalizing forecasts that drift off the predictive manifold (Section~\ref{sec:manifold-aware-consistency}). Since the manifold is not directly observable on real data, we use the train--test loss gap as a proxy: a memorizing model fits the training set rapidly while its test error grows, whereas a model that captures the dynamics improves on both. Figure~\ref{fig:train_test_curve} confirms this picture on ETTh2 and ETTm2. The MSE-trained MLP drives the training loss down sharply but its test loss soon rises, the signature of off-manifold memorization; with CL, the training loss decreases more gradually while, crucially, the test loss continues to fall. This matches exactly the behavior predicted by the off-manifold bound in Theorem~\ref{thm:manifold-global-informal} in Section~\ref{sec:manifold-aware-consistency}.

\paragraph{Rollout Stability.} Section~\ref{sec:manifold-aware-consistency} links reduced off-manifold sensitivity to free-running stability under suitable conditions. To examine this interpretation beyond one-shot forecasting, we test CL when predictions are repeatedly fed back as inputs. We train with $L=720$ and $H=96$, then evaluate recursive rollouts at horizons 336 and 720, using separately trained one-shot models as references. Table~\ref{tab:rollout-vs-oneblock} shows that CL achieves lower or equal rollout MSE than the baseline across all evaluated datasets and forecasting horizons.  For example, the 720-step rollout MSE on ETTh1 falls from $0.508$ to $0.427$. With CL, rollout performance remains close to, and sometimes better than, the corresponding one-shot results, supporting improved free-running robustness. Appendix~\ref{app:exp-rollout} provides the detailed analysis and results with standard deviations.

\begin{table}[ht]
\centering
\caption{Mean test MSE over seeds for recursive rollout and one-shot forecasting. For each (Original, +CL) pair, the better value is shown in \textbf{bold}. More details are reported in Appendix~\ref{app:exp-rollout}.}
\label{tab:rollout-vs-oneblock}
\footnotesize
\setlength{\tabcolsep}{4pt}
\renewcommand{\arraystretch}{1.05}
\begin{tabular*}{\textwidth}{@{\extracolsep{\fill}}lcl|ccccccc@{}}
\toprule
\textbf{Forecasting} & $H$ & \textbf{Method} & ETTh1 & ETTh2 & ETTm1 & ETTm2 & Weather & ECL & Traffic \\
\midrule
\multirow{4}{*}{Rollout} & \multirow{2}{*}{336} & Original & 0.461 & 0.403 & 0.383 & 0.299 & 0.263 & 0.162 & 0.392 \\
 &  & \textbf{+CL} & \textbf{0.411} & \textbf{0.361} & \textbf{0.353} & \textbf{0.265} & \textbf{0.235} & \textbf{0.161} & \textbf{0.391} \\
\cmidrule(lr){2-10}
 & \multirow{2}{*}{720} & Original & 0.508 & 0.428 & 0.452 & 0.381 & 0.345 & 0.200 & 0.428 \\
 &  & \textbf{+CL} & \textbf{0.427} & \textbf{0.399} & \textbf{0.412} & \textbf{0.346} & \textbf{0.308} & \textbf{0.200} & \textbf{0.428} \\
\midrule
\multirow{4}{*}{One-shot} & \multirow{2}{*}{336} & Original & 0.458 & 0.395 & 0.377 & 0.296 & 0.258 & 0.163 & 0.388 \\
 &  & \textbf{+CL} & \textbf{0.434} & \textbf{0.362} & \textbf{0.352} & \textbf{0.268} & \textbf{0.235} & \textbf{0.160} & \textbf{0.388} \\
\cmidrule(lr){2-10}
 & \multirow{2}{*}{720} & Original & 0.500 & 0.452 & 0.446 & 0.378 & 0.336 & 0.198 & 0.428 \\
 &  & \textbf{+CL} & \textbf{0.458} & \textbf{0.397} & \textbf{0.410} & \textbf{0.348} & \textbf{0.312} & \textbf{0.198} & \textbf{0.428} \\
\bottomrule
\end{tabular*}
\end{table}

\section{Related Work}
\label{sec:related-work}

Modern long-horizon forecasting has been driven largely by architectural design, including Informer \citep{zhou2021informer}, Autoformer and FEDformer \citep{wu2021autoformer,zhou2022fedformer}, DLinear \citep{zeng2023transformers}, PatchTST \citep{nie2023patchtst}, RevIN \citep{kim2022reversible}, TiDE \citep{das2023tide}, and iTransformer \citep{liu2024itransformer}. Our contribution is orthogonal to this line: instead of changing the backbone, we add a training objective that can be attached to linear, MLP, or Transformer forecasters.

The objective is motivated by the teacher-forcing/free-running mismatch studied in autoregressive sequence modeling and imitation learning. Scheduled sampling \citep{bengio2015scheduled} and flipped-classroom curricula \citep{teutsch2022flipped} mix teacher-forced and free-running inputs during autoregressive decoding, professor forcing \citep{lamb2016professor} aligns hidden trajectories, and DAgger, Data as Demonstrator, and stable-rollout regularization \citep{ross2011dagger,venkatraman2015improving,talvitie2014model} train on learner-visited or generated rollout states. We use the teacher-forcing/free-running comparison differently: for standard one-shot block forecasters, it serves as a local diagnostic of whether feeding back the forecast changes the next prediction. This yields an output-level consistency penalty, implemented through a residual-direction JVP, without changing the inference protocol.

Our first-order surrogate is also related to Jacobian and directional-invariance regularization, including contractive autoencoders \citep{rifai2011contractive}, supervised Jacobian regularization \citep{hoffman2019robust}, Lipschitz and robustness methods \citep{singla2021improved,zhang2022rethinking}, Tangent Prop \citep{simard1991tangent}, noise-injection and denoising views \citep{bishop1995training,vincent2010stacked}, and long-time dynamics regularization \citep{pan2018longtime}. The key distinction is directional: rather than penalizing the full Jacobian or isotropic perturbations, we suppress the model's response along the residual-induced direction through which its own forecast error is fed back. Our method therefore targets the directions that are operationally relevant for rollout mismatch. This viewpoint is related in spirit to structured shrinkage methods such as ridge/Tikhonov regularization \citep{hoerl1970ridge}, reduced-rank regression \citep{izenman1975reduced}, total least squares \citep{golub1980total}, and James--Stein shrinkage \citep{james1961estimation,efron1975stein}, but the shrinkage here is state-dependent and measured by downstream rollout sensitivity. Appendix~\ref{app:extended-related-work} provides a more detailed comparison.

\section{Conclusion}
\label{sec:conclusion}

We introduced consistency loss as a training principle for long-horizon block forecasting. The key idea is to turn the idea---the input history, the observed future target, and the model forecast are all segments of the same underlying series---into a local, differentiable regularizer: instead of only fitting the next target, the model is trained to suppress its sensitivity along the residual direction created by its own forecast error. This yields a practical Jacobian--vector-product objective that preserves one-shot inference, applies across architectures, and admits a predictive-manifold interpretation as control of off-manifold sensitivity. Empirically, this simple objective improves both a residual MLP and several strong forecasting backbones, showing that robust time series forecasting performance can be built directly into the loss rather than only into the architecture.

\begin{ack}
This work was supported by Bosch (China) Investment Co., Ltd., including computational resources. The authors declare no additional competing interests beyond their disclosed affiliations.
\end{ack}

\newpage

\bibliography{ref}

\begin{thebibliography}{10}

\bibitem{alain2014regularized}
Guillaume Alain and Yoshua Bengio.
\newblock What regularized auto-encoders learn from the data-generating distribution.
\newblock {\em The Journal of Machine Learning Research}, 15(1):3563--3593, 2014.

\bibitem{asadi2019combating}
Kavosh Asadi, Dipendra Misra, Seungchan Kim, and Michel~L. Littman.
\newblock Combating the compounding-error problem with a multi-step model.
\newblock {\em arXiv preprint arXiv:1905.13320}, 2019.

\bibitem{bengio2015scheduled}
Samy Bengio, Oriol Vinyals, Navdeep Jaitly, and Noam Shazeer.
\newblock Scheduled sampling for sequence prediction with recurrent neural networks.
\newblock {\em Advances in neural information processing systems}, 28, 2015.

\bibitem{bengio2013representation}
Yoshua Bengio, Aaron Courville, and Pascal Vincent.
\newblock Representation learning: A review and new perspectives.
\newblock {\em IEEE transactions on pattern analysis and machine intelligence}, 35(8):1798--1828, 2013.

\bibitem{bishop1995training}
Chris~M Bishop.
\newblock Training with noise is equivalent to tikhonov regularization.
\newblock {\em Neural computation}, 7(1):108--116, 1995.

\bibitem{box1975intervention}
George~EP Box and George~C Tiao.
\newblock Intervention analysis with applications to economic and environmental problems.
\newblock {\em Journal of the American Statistical association}, 70(349):70--79, 1975.

\bibitem{chen2023tsmixer}
Si-An Chen, Chun-Liang Li, Nate Yoder, Sercan~O Arik, and Tomas Pfister.
\newblock Tsmixer: An all-mlp architecture for time series forecasting.
\newblock {\em arXiv preprint arXiv:2303.06053}, 2023.

\bibitem{dai2024pdf}
Tao Dai, Beiliang Wu, Peiyuan Liu, Naiqi Li, Jigang Bao, Yong Jiang, and Shu-Tao Xia.
\newblock Periodicity decoupling framework for long-term series forecasting.
\newblock In {\em The twelfth international conference on learning representations}, 2024.

\bibitem{das2023tide}
Abhimanyu Das, Weihao Kong, Andrew Leach, Shaan Mathur, Rajat Sen, and Rose Yu.
\newblock Long-term forecasting with tide: Time-series dense encoder.
\newblock {\em arXiv preprint arXiv:2304.08424}, 2023.

\bibitem{efron1975stein}
Bradley Efron and Carl Morris.
\newblock Data analysis using stein's estimator and its generalizations.
\newblock {\em Journal of the American Statistical Association}, 70(350):311--319, 1975.

\bibitem{eldele2024tslanet}
Emadeldeen Eldele, Mohamed Ragab, Zhenghua Chen, Min Wu, and Xiaoli Li.
\newblock Tslanet: Rethinking transformers for time series representation learning.
\newblock {\em arXiv preprint arXiv:2404.08472}, 2024.

\bibitem{fefferman2016testing}
Charles Fefferman, Sanjoy Mitter, and Hariharan Narayanan.
\newblock Testing the manifold hypothesis.
\newblock {\em Journal of the American Mathematical Society}, 29(4):983--1049, 2016.

\bibitem{golub1980total}
Gene~H Golub and Charles~F Van~Loan.
\newblock An analysis of the total least squares problem.
\newblock {\em SIAM journal on numerical analysis}, 17(6):883--893, 1980.

\bibitem{hansen2018sure}
Niels~Richard Hansen.
\newblock On stein’s unbiased risk estimate for reduced rank estimators.
\newblock {\em Statistics \& Probability Letters}, 135:76--82, 2018.

\bibitem{hoerl1970ridge}
Arthur~E Hoerl and Robert~W Kennard.
\newblock Ridge regression: Biased estimation for nonorthogonal problems.
\newblock {\em Technometrics}, 12(1):55--67, 1970.

\bibitem{hoffman2019robust}
Judy Hoffman, Daniel~A Roberts, and Sho Yaida.
\newblock Robust learning with jacobian regularization.
\newblock {\em arXiv preprint arXiv:1908.02729}, 2019.

\bibitem{hu2025adaptive}
Yifan Hu, Peiyuan Liu, Peng Zhu, Dawei Cheng, and Tao Dai.
\newblock Adaptive multi-scale decomposition framework for time series forecasting.
\newblock In {\em Proceedings of the AAAI Conference on Artificial Intelligence}, volume~39, pages 17359--17367, 2025.

\bibitem{izenman1975reduced}
Alan~Julian Izenman.
\newblock Reduced-rank regression for the multivariate linear model.
\newblock {\em Journal of multivariate analysis}, 5(2):248--264, 1975.

\bibitem{james1961estimation}
William James, Charles Stein, et~al.
\newblock Estimation with quadratic loss.
\newblock In {\em Proceedings of the fourth Berkeley symposium on mathematical statistics and probability}, volume~1, pages 361--379. University of California Press, 1961.

\bibitem{kailath1970innovations}
Thomas Kailath.
\newblock The innovations approach to detection and estimation theory.
\newblock {\em Proceedings of the IEEE}, 58(5):680--695, 2005.

\bibitem{kim2025comprehensive}
Jongseon Kim, Hyungjoon Kim, HyunGi Kim, Dongjun Lee, and Sungroh Yoon.
\newblock A comprehensive survey of deep learning for time series forecasting: architectural diversity and open challenges.
\newblock {\em Artificial Intelligence Review}, 58(7):216, 2025.

\bibitem{kim2022reversible}
Taesung Kim, Jinhee Kim, Yunwon Tae, Cheonbok Park, Jang-Ho Choi, and Jaegul Choo.
\newblock Reversible instance normalization for accurate time-series forecasting against distribution shift.
\newblock In {\em International conference on learning representations}, 2021.

\bibitem{kingma2014adam}
Diederik~P Kingma and Jimmy Ba.
\newblock Adam: A method for stochastic optimization.
\newblock {\em arXiv preprint arXiv:1412.6980}, 2014.

\bibitem{kong2025deep}
Xiangjie Kong, Zhenghao Chen, Weiyao Liu, Kaili Ning, Lechao Zhang, Syauqie Muhammad~Marier, Yichen Liu, Yuhao Chen, and Feng Xia.
\newblock Deep learning for time series forecasting: a survey.
\newblock {\em International Journal of Machine Learning and Cybernetics}, 16(7):5079--5112, 2025.

\bibitem{lamb2016professor}
Alex~M Lamb, Anirudh~Goyal ALIAS PARTH~GOYAL, Ying Zhang, Saizheng Zhang, Aaron~C Courville, and Yoshua Bengio.
\newblock Professor forcing: A new algorithm for training recurrent networks.
\newblock {\em Advances in neural information processing systems}, 29, 2016.

\bibitem{lin2024sparsetsf}
Shengsheng Lin, Weiwei Lin, Wentai Wu, Haojun Chen, and Junjie Yang.
\newblock Sparsetsf: Modeling long-term time series forecasting with 1k parameters.
\newblock {\em arXiv preprint arXiv:2405.00946}, 2024.

\bibitem{liu2024itransformer}
Yong Liu, Tengge Hu, Haoran Zhang, Haixu Wu, Shiyu Wang, Lintao Ma, and Mingsheng Long.
\newblock itransformer: Inverted transformers are effective for time series forecasting.
\newblock {\em arXiv preprint arXiv:2310.06625}, 2023.

\bibitem{mccormack2022frechet}
Andrew McCormack and Peter Hoff.
\newblock The stein effect for fr{\'e}chet means.
\newblock {\em The Annals of Statistics}, 50(6):3647--3676, 2022.

\bibitem{metropolis1949monte}
Nicholas Metropolis and Stanislaw Ulam.
\newblock The monte carlo method.
\newblock {\em Journal of the American statistical association}, 44(247):335--341, 1949.

\bibitem{ni2026timedistill}
Juntong Ni, Zewen Liu, Shiyu Wang, Ming Jin, and Wei Jin.
\newblock Timedistill: Efficient long-term time series forecasting with mlp via cross-architecture distillation.
\newblock In {\em Proceedings of the 32nd ACM SIGKDD Conference on Knowledge Discovery and Data Mining V. 1}, pages 1113--1124, 2026.

\bibitem{nie2023patchtst}
Yuqi Nie, Nam~H Nguyen, Phanwadee Sinthong, and Jayant Kalagnanam.
\newblock A time series is worth 64 words: Long-term forecasting with transformers.
\newblock {\em arXiv preprint arXiv:2211.14730}, 2022.

\bibitem{oman1982contracting}
Samuel~D Oman.
\newblock Contracting towards subspaces when estimating the mean of a multivariate normal distribution.
\newblock {\em Journal of Multivariate Analysis}, 12(2):270--290, 1982.

\bibitem{pan2018longtime}
Shaowu Pan and Karthik Duraisamy.
\newblock Long-time predictive modeling of nonlinear dynamical systems using neural networks.
\newblock {\em Complexity}, 2018(1):4801012, 2018.

\bibitem{parthasarathy2025closing}
Arjun Parthasarathy, Nimit Kalra, Rohun Agrawal, Yann LeCun, Oumayma Bounou, Pavel Izmailov, and Micah Goldblum.
\newblock Closing the train-test gap in world models for gradient-based planning.
\newblock {\em arXiv preprint arXiv:2512.09929v1}, 2025.

\bibitem{paszke2017automatic}
Adam Paszke, Sam Gross, Soumith Chintala, Gregory Chanan, Edward Yang, Zachary DeVito, Zeming Lin, Alban Desmaison, Luca Antiga, and Adam Lerer.
\newblock Automatic differentiation in pytorch.
\newblock 2017.

\bibitem{perlman2012reversing}
Michael~D Perlman and Sanjay Chaudhuri.
\newblock Reversing the stein effect.
\newblock 2012.

\bibitem{qiu2024duet}
Xiangfei Qiu, Xingjian Wu, Yan Lin, Chenjuan Guo, Jilin Hu, and Bin Yang.
\newblock Duet: Dual clustering enhanced multivariate time series forecasting.
\newblock In {\em Proceedings of the 31st ACM SIGKDD Conference on Knowledge Discovery and Data Mining V. 1}, pages 1185--1196, 2025.

\bibitem{rifai2011contractive}
Salah Rifai, Pascal Vincent, Xavier Muller, Xavier Glorot, and Yoshua Bengio.
\newblock Contractive auto-encoders: Explicit invariance during feature extraction.
\newblock In {\em Proceedings of the 28th international conference on international conference on machine learning}, pages 833--840, 2011.

\bibitem{ross2011dagger}
St{\'e}phane Ross, Geoffrey Gordon, and Drew Bagnell.
\newblock A reduction of imitation learning and structured prediction to no-regret online learning.
\newblock In {\em Proceedings of the fourteenth international conference on artificial intelligence and statistics}, pages 627--635. JMLR Workshop and Conference Proceedings, 2011.

\bibitem{simard1991tangent}
Patrice Simard, Bernard Victorri, Yann LeCun, and John Denker.
\newblock Tangent prop-a formalism for specifying selected invariances in an adaptive network.
\newblock {\em Advances in neural information processing systems}, 4, 1991.

\bibitem{singla2021improved}
Sahil Singla, Surbhi Singla, and Soheil Feizi.
\newblock Improved deterministic l2 robustness on cifar-10 and cifar-100.
\newblock {\em arXiv preprint arXiv:2108.04062}, 2021.

\bibitem{stein1981mean}
Charles~M Stein.
\newblock Estimation of the mean of a multivariate normal distribution.
\newblock {\em The annals of Statistics}, pages 1135--1151, 1981.

\bibitem{taieb2012review}
Souhaib~Ben Taieb, Gianluca Bontempi, Amir~F Atiya, and Antti Sorjamaa.
\newblock A review and comparison of strategies for multi-step ahead time series forecasting based on the nn5 forecasting competition.
\newblock {\em Expert systems with applications}, 39(8):7067--7083, 2012.

\bibitem{talvitie2014model}
Erik Talvitie.
\newblock Model regularization for stable sample rollouts.
\newblock In {\em UAI}, pages 780--789, 2014.

\bibitem{teutsch2022flipped}
Philipp Teutsch and Patrick M{\"a}der.
\newblock Flipped classroom: Effective teaching for time series forecasting.
\newblock {\em arXiv preprint arXiv:2210.08959}, 2022.

\bibitem{tsay1988outliers}
Ruey~S Tsay.
\newblock Outliers, level shifts, and variance changes in time series.
\newblock {\em Journal of forecasting}, 7(1):1--20, 1988.

\bibitem{venkatraman2015improving}
Arun Venkatraman, Martial Hebert, and J~Bagnell.
\newblock Improving multi-step prediction of learned time series models.
\newblock In {\em Proceedings of the AAAI Conference on Artificial Intelligence}, volume~29, 2015.

\bibitem{vincent2010stacked}
Pascal Vincent, Hugo Larochelle, Isabelle Lajoie, Yoshua Bengio, Pierre-Antoine Manzagol, and L{\'e}on Bottou.
\newblock Stacked denoising autoencoders: Learning useful representations in a deep network with a local denoising criterion.
\newblock {\em Journal of machine learning research}, 11(12), 2010.

\bibitem{wang2024timemixer}
Shiyu Wang, Haixu Wu, Xiaoming Shi, Tengge Hu, Huakun Luo, Lintao Ma, James~Y Zhang, and Jun Zhou.
\newblock Timemixer: Decomposable multiscale mixing for time series forecasting.
\newblock {\em arXiv preprint arXiv:2405.14616}, 2024.

\bibitem{wang2025characteristic}
Zheng Wang, Kaixuan Zhang, Wanfang Chen, Xiaonan Lu, Longyuan Li, and Tobias Schlagenhauf.
\newblock Characteristic root analysis and regularization for linear time series forecasting.
\newblock {\em arXiv preprint arXiv:2509.23597}, 2025.

\bibitem{williams1989learning}
Ronald~J Williams and David Zipser.
\newblock A learning algorithm for continually running fully recurrent neural networks.
\newblock {\em Neural computation}, 1(2):270--280, 1989.

\bibitem{woo2022etsformer}
Gerald Woo, Chenghao Liu, Doyen Sahoo, Akshat Kumar, and Steven Hoi.
\newblock Etsformer: Exponential smoothing transformers for time-series forecasting.
\newblock {\em arXiv preprint arXiv:2202.01381}, 2022.

\bibitem{wu2023timesnet}
Haixu Wu, Tengge Hu, Yong Liu, Hang Zhou, Jianmin Wang, and Mingsheng Long.
\newblock Timesnet: Temporal 2d-variation modeling for general time series analysis.
\newblock {\em arXiv preprint arXiv:2210.02186}, 2022.

\bibitem{wu2021autoformer}
Haixu Wu, Jiehui Xu, Jianmin Wang, and Mingsheng Long.
\newblock Autoformer: Decomposition transformers with auto-correlation for long-term series forecasting.
\newblock {\em Advances in neural information processing systems}, 34:22419--22430, 2021.

\bibitem{xu2024fits}
Zhijian Xu, Ailing Zeng, and Qiang Xu.
\newblock Fits: Modeling time series with $10 k $ parameters.
\newblock {\em arXiv preprint arXiv:2307.03756}, 2023.

\bibitem{yi2024filternet}
Kun Yi, Jingru Fei, Qi~Zhang, Hui He, Shufeng Hao, Defu Lian, and Wei Fan.
\newblock Filternet: Harnessing frequency filters for time series forecasting.
\newblock {\em Advances in Neural Information Processing Systems}, 37:55115--55140, 2024.

\bibitem{zeng2023transformers}
Ailing Zeng, Muxi Chen, Lei Zhang, and Qiang Xu.
\newblock Are transformers effective for time series forecasting?
\newblock In {\em Proceedings of the AAAI conference on artificial intelligence}, volume~37, pages 11121--11128, 2023.

\bibitem{zhang2022rethinking}
Bohang Zhang, Du~Jiang, Di~He, and Liwei Wang.
\newblock Rethinking lipschitz neural networks and certified robustness: A boolean function perspective.
\newblock {\em Advances in neural information processing systems}, 35:19398--19413, 2022.

\bibitem{zhang2022lightts}
Tianping Zhang, Yizhuo Zhang, Wei Cao, Jiang Bian, Xiaohan Yi, Shun Zheng, and Jian Li.
\newblock Less is more: Fast multivariate time series forecasting with light sampling-oriented mlp structures, 2022.

\bibitem{zhou2021informer}
Haoyi Zhou, Shanghang Zhang, Jieqi Peng, Shuai Zhang, Jianxin Li, Hui Xiong, and Wancai Zhang.
\newblock Informer: Beyond efficient transformer for long sequence time-series forecasting.
\newblock In {\em Proceedings of the AAAI conference on artificial intelligence}, volume~35, pages 11106--11115, 2021.

\bibitem{zhou2022fedformer}
Tian Zhou, Ziqing Ma, Qingsong Wen, Xue Wang, Liang Sun, and Rong Jin.
\newblock Fedformer: Frequency enhanced decomposed transformer for long-term series forecasting.
\newblock In {\em International conference on machine learning}, pages 27268--27286. PMLR, 2022.

\end{thebibliography}
\bibliographystyle{plain}

\newpage

\appendix
\section{Theories and Proofs}
\label{app:thm-and-proof}
\subsection{Proofs for Proposition~\ref{thm:linear-nonlinear-consistency}}
\label{app:proofs-rollout-consistency}
Recall that in Section~\ref{subsec:linear-nonlinear-reduction}, we state Proposition~\ref{thm:linear-nonlinear-consistency} as
\begin{nonproposition}[Local consistency as directional sensitivity]
\label{appx:thm:linear-nonlinear-consistency}
Let \(\delta_t=x_{t+1}^{\mathrm{TF}}-x_{t+1}^{\mathrm{FR}}\) be the next-state perturbation defined in Equation~\eqref{eq:delta-next-state}, and consider the exact consistency functional \(\mathcal C_\theta(x_t,y_t)\) in Equation~\eqref{eq:exact-consistency}. Then the following statements hold.

\begin{enumerate}
    \item[\textup{(i)}] \textbf{Linear case.}
    If \(g_\theta(x)=Gx\) is linear, then
    \begin{equation}
    \label{eq:app-linear-exact-consistency}
    \mathcal C_\theta(x_t,y_t)
    =
    \|G\delta_t\|_2^2.
    \end{equation}

    \item[\textup{(ii)}] \textbf{Nonlinear case.}
    Suppose \(g_\theta\) is \(C^2\) in a neighborhood of \(x_{t+1}^{\mathrm{FR}}\). Then, as \(\|\delta_t\|_2 \to 0\),
    \begin{equation}
    \label{eq:nonlinear-first-order-consistency}
    \mathcal C_\theta(x_t,y_t)
    =
    \left\|
    Jg_\theta(x_{t+1}^{\mathrm{FR}})\,\delta_t
    \right\|_2^2
    +
    O(\|\delta_t\|_2^3),
    \end{equation}
    where \(\delta_t=x_{t+1}^{\mathrm{TF}}-x_{t+1}^{\mathrm{FR}}\), and where \(Jg_\theta(x)\) denotes the Jacobian of \(g_\theta\) at \(x\). 
\end{enumerate}
\end{nonproposition}

\begin{proof}
Let
\[
\delta_t := x_{t+1}^{\mathrm{TF}}-x_{t+1}^{\mathrm{FR}},
\qquad
q_t := x_{t+1}^{\mathrm{FR}}.
\]
Therefore,
\begin{equation}
\label{eq:appendix-C-as-increment}
\mathcal C_\theta(x_t,y_t)
=
\|g_\theta(q_t+\delta_t)-g_\theta(q_t)\|_2^2.
\end{equation}

\paragraph{Linear case.}
If \(g_\theta(x)=Gx\), then
\[
g_\theta(q_t+\delta_t)-g_\theta(q_t)=G(q_t+\delta_t)-Gq_t=G\delta_t.
\]
which proves \eqref{eq:app-linear-exact-consistency}.

\paragraph{Nonlinear case.}
Assume \(g_\theta\in C^2\) in a neighborhood of $q_t$. Since \(q_t\) is fixed and \(g_\theta\) is twice continuously differentiable, there exists a neighborhood of the line segment
\[
\{q_t+s\delta_t: s\in[0,1]\}
\]
on which the Hessian of \(g_\theta\) is bounded. Hence, by Taylor's theorem with remainder,
\begin{equation}
\label{eq:appendix-taylor}
g_\theta(q_t+\delta_t)
=
g_\theta(q_t)
+
Jg_\theta(q_t)\,\delta_t
+
\varepsilon_t,
\qquad
\|\varepsilon_t\|_2 \le c_t\|\delta_t\|_2^2
\end{equation}
for some local constant \(c_t<\infty\).
Substituting \eqref{eq:appendix-taylor} into \eqref{eq:appendix-C-as-increment} gives
\[
\mathcal C_\theta(x_t,y_t)
=
\|Jg_\theta(q_t)\,\delta_t + \varepsilon_t\|_2^2.
\]
Expanding the square,
\[
\mathcal C_\theta(x_t,y_t)
=
\|Jg_\theta(q_t)\,\delta_t\|_2^2
+
2\langle Jg_\theta(q_t)\,\delta_t,\varepsilon_t\rangle
+
\|\varepsilon_t\|_2^2.
\]
Because \(Jg_\theta(q_t)\) is bounded locally, we have
\[
\|Jg_\theta(q_t)\,\delta_t\|_2 = O(\|\delta_t\|_2),
\qquad
\|\varepsilon_t\|_2 = O(\|\delta_t\|_2^2).
\]
Therefore,
\[
2\langle Jg_\theta(q_t)\,\delta_t,\varepsilon_t\rangle = O(\|\delta_t\|_2^3),
\qquad
\|\varepsilon_t\|_2^2 = O(\|\delta_t\|_2^4).
\]
It follows that
\[
\mathcal C_\theta(x_t,y_t)
=
\|Jg_\theta(q_t)\,\delta_t\|_2^2 + O(\|\delta_t\|_2^3).
\]
Recalling that \(\delta_t=x_{t+1}^{\mathrm{TF}}-x_{t+1}^{\mathrm{FR}}\) and \(q_t=x_{t+1}^{\mathrm{FR}}\), we obtain
\[
\mathcal C_\theta(x_t,y_t)
=
\left\|
Jg_\theta(x_{t+1}^{\mathrm{FR}})\,(x_{t+1}^{\mathrm{TF}}-x_{t+1}^{\mathrm{FR}})
\right\|_2^2
+
O(\|x_{t+1}^{\mathrm{TF}}-x_{t+1}^{\mathrm{FR}}\|_2^3),
\]
which proves \eqref{eq:nonlinear-first-order-consistency}. 
\end{proof}

\subsection{Random residual insertion removes positional bias}
\label{app:positional-bias}

\paragraph{Random insertion removes fixed-location bias.}
The residual-insertion operator should not privilege a particular location in the
history window. The following lemma formalizes this point. A deterministic
insertion only tests the residual response on one coordinate subspace, whereas
randomized insertion is an unbiased Monte Carlo estimate of the position-averaged
directional response and gives every admissible location nonzero control.

\begin{lemma}[Random residual insertion removes fixed-location bias]
\label{lem:random-insertion-bias}
Let $\Omega$ be a finite set of admissible insertion positions, and let
$P_\omega:\mathcal Y\to\mathcal X$ be the zero-padding insertion operator
associated with position $\omega\in\Omega$. For a local Jacobian $J$ and residual
$r\in\mathcal Y$, define the position-wise directional response
\[
R_\omega(J,r):=\|J P_\omega r\|_2^2 .
\]
Let $q$ be a distribution on $\Omega$ with full support, and define
\[
R_q(J,r):=\mathbb E_{\omega\sim q} R_\omega(J,r)
       = \sum_{\omega\in\Omega} q_\omega \|J P_\omega r\|_2^2 .
\]
Then:
\[
\mathbb E_{\omega\sim q}\big[\|J P_\omega r\|_2^2\big] = R_q(J,r),
\]
so a single sampled insertion gives an unbiased Monte Carlo estimate of the
position-averaged sensitivity. Moreover, for every $\omega\in\Omega$,
\[
R_\omega(J,r)
\le \frac{1}{q_\omega} R_q(J,r)
\le \frac{1}{q_{\min}} R_q(J,r),
\qquad
q_{\min}:=\min_{\omega\in\Omega} q_\omega .
\]
Thus controlling $R_q$ controls the residual sensitivity at every admissible
insertion position up to the sampling weight.

In contrast, for any fixed insertion position $\omega_0$, if there exists another
position $\omega_1$ such that
$\operatorname{Im}(P_{\omega_1})\not\subseteq \operatorname{Im}(P_{\omega_0})$,
then for any $M>0$ there exist a Jacobian $J$ and a residual $r$ such that
\[
R_{\omega_0}(J,r)=0,
\qquad
R_{\omega_1}(J,r)\ge M .
\]
Therefore a deterministic insertion can report zero consistency penalty while
leaving arbitrarily large residual sensitivity at another valid position.
\end{lemma}

\begin{proof}
The unbiasedness claim follows directly from the definition of $R_q$. Since all
terms in the sum defining $R_q$ are nonnegative, for every $\omega\in\Omega$,
\[
R_q(J,r)=\sum_{\omega'\in\Omega} q_{\omega'}R_{\omega'}(J,r)
\ge q_\omega R_\omega(J,r),
\]
which proves the full-support control bound.

For the deterministic counterexample, choose $r$ such that
$u:=P_{\omega_1}r\notin \operatorname{Im}(P_{\omega_0})$. Such an $r$ exists by
the assumption
$\operatorname{Im}(P_{\omega_1})\not\subseteq \operatorname{Im}(P_{\omega_0})$.
Since $u$ is not in $\operatorname{Im}(P_{\omega_0})$, there exists a linear
functional $a$ that vanishes on $\operatorname{Im}(P_{\omega_0})$ but satisfies
$a^\top u\neq 0$. Let $J=\alpha e_1 a^\top$ for a scalar $\alpha>0$ and a unit
vector $e_1$ in the output space. Then $Jv=0$ for all
$v\in\operatorname{Im}(P_{\omega_0})$, so $R_{\omega_0}(J,r)=0$. However,
\[
R_{\omega_1}(J,r)
= \|J u\|_2^2
= \alpha^2 |a^\top u|^2 .
\]
Choosing $\alpha$ large enough makes this quantity at least $M$. This proves the
claim.
\end{proof}

Lemma~\ref{lem:random-insertion-bias} justifies the design in Eq.~\eqref{eq:first-order-consistency-loss-readable}. We are not
trying to construct a globally valid randomly padded next window. Instead, the
training loss estimates the averaged residual-direction sensitivity over insertion
positions. Deterministic history padding corresponds to optimizing only
$R_{\omega_0}$, which can completely miss high-gain residual directions at other
positions; randomized insertion optimizes $R_q$ and therefore removes this
fixed-location bias.

\subsection{Full Statement of the Local-to-Global Predictive-Tube Theorem and Proof}
\label{app:manifold-global-theorem}

For completeness, we restate the full version of the local-to-global guarantee summarized informally in Theorem~\ref{thm:manifold-global-informal}.

\begin{theorem}[Full residual consistency implies predictive-tube stability]
\label{thm:full-residual-manifold-global}
Let \(\mathcal M\subset \mathcal X\cong\mathbb R^m\) be a compact \(C^2\) embedded predictive manifold. Let \(U_{\rho_*}(\mathcal M)\) be a tubular neighborhood on which the nearest-point projection
\[
\pi:U_{\rho_*}(\mathcal M)\to \mathcal M
\]
is well-defined, and set \(\Pi^\perp(x):=\Pi_{N_{\pi(x)}\mathcal M}\). Let \(0<\rho\le \rho_*\). Assume \(\Phi_\theta\in C^2\) on \(U_{\rho_*}(\mathcal M)\) and assume the following conditions.

\begin{enumerate}
    \item[\textup{A1)}] \textbf{Approximate manifold invariance.}
    \begin{equation}
    \label{eq:full-epsM}
    \varepsilon_{\mathcal M}
    :=
    \sup_{z\in\mathcal M}
    d(\Phi_\theta(z),\mathcal M)
    <\rho_* .
    \end{equation}
    In particular, \(\Phi_\theta(z)\in U_{\rho_*}(\mathcal M)\) for all \(z\in\mathcal M\), so \(\Pi^\perp(\Phi_\theta(z))\) is well-defined.

    \item[\textup{A2)}] \textbf{Normal excitation by full residual directions.}
    For each training pair, define
    \[
    q_t:=\Phi_\theta(x_t),
    \qquad
    \delta_t:=P(y_t-g_\theta(x_t)).
    \]
    Let \(\mathcal Q\subset U_\rho(\mathcal M)\) denote the set of visited states on which the directional penalty is controlled. For every \(q\in\mathcal Q\), define the conditional second moment
    \begin{equation}
    \label{eq:full-residual-second-moment}
    M_\delta(q)
    :=
    \mathbb E[\delta_t\delta_t^\top\mid q_t=q].
    \end{equation}
    There exists \(\mu>0\) such that
    \begin{equation}
    \label{eq:full-normal-excitation}
    M_\delta(q)
    \succeq
    \mu\,\Pi^\perp(q),
    \qquad q\in\mathcal Q .
    \end{equation}

    \item[\textup{A3)}] \textbf{Bounded full directional penalty.}
    There exists \(\eta_0>0\) such that, for every \(q\in\mathcal Q\),
    \begin{equation}
    \label{eq:full-directional-bound}
    \mathbb E
    \left[
    \|Jg_\theta(q)\delta_t\|_2^2
    \mid q_t=q
    \right]
    \le
    \eta_0 .
    \end{equation}

    \item[\textup{A4)}] \textbf{Coverage and smoothness.}
    The function
    \begin{equation}
    \label{eq:full-hN}
    h_N(x)
    :=
    \|Jg_\theta(x)\Pi^\perp(x)\|_F
    \end{equation}
    is \(L_h\)-Lipschitz on \(U_\rho(\mathcal M)\), and \(\mathcal Q\) is a \(\tau\)-net of \(U_\rho(\mathcal M)\).

    \item[\textup{A5)}] \textbf{Normal gains of the state-update operator.}
    Define
    \begin{equation}
    \label{eq:full-aN}
    a_N
    :=
    \sup_{z\in\mathcal M}
    \left\|
    \Pi^\perp(\Phi_\theta(z))
    D_1S(z,g_\theta(z))
    \Pi^\perp(z)
    \right\|,
    \end{equation}
    and
    \begin{equation}
    \label{eq:full-bN}
    b_N
    :=
    \sup_{z\in\mathcal M}
    \left\|
    \Pi^\perp(\Phi_\theta(z))
    D_2S(z,g_\theta(z))
    \right\|.
    \end{equation}
\end{enumerate}

Then the following statements hold.

\begin{enumerate}
    \item[\textup{(i)}] \textbf{Normal operator control from full residual probes.}
    \begin{equation}
    \label{eq:full-normal-operator-control}
    \sup_{x\in U_\rho(\mathcal M)}
    \|Jg_\theta(x)\Pi^\perp(x)\|_F
    \le
    \gamma_N,
    \qquad
    \gamma_N
    :=
    \sqrt{\eta_0/\mu}+L_h\tau .
    \end{equation}

    \item[\textup{(ii)}] \textbf{Local transverse recursion.}
    There exists a constant \(C_\rho>0\) such that, for every \(x\in U_\rho(\mathcal M)\),
    \begin{equation}
    \label{eq:full-local-recursion}
    d(\Phi_\theta(x),\mathcal M)
    \le
    \varepsilon_{\mathcal M}
    +
    \kappa_N d(x,\mathcal M)
    +
    C_\rho d(x,\mathcal M)^2,
    \end{equation}
    where
    \begin{equation}
    \label{eq:full-kappa}
    \kappa_N
    :=
    a_N+b_N\gamma_N .
    \end{equation}

    \item[\textup{(iii)}] \textbf{Global-in-time predictive-tube stability.}
    If
    \begin{equation}
    \label{eq:full-bar-kappa}
    \bar\kappa
    :=
    \kappa_N+C_\rho\rho
    <1,
    \qquad
    \varepsilon_{\mathcal M}\le (1-\bar\kappa)\rho,
    \end{equation}
    then \(U_\rho(\mathcal M)\) is forward invariant under \(\Phi_\theta\), and every rollout sequence
    \[
    x_{k+1}=\Phi_\theta(x_k),
    \qquad x_0\in U_\rho(\mathcal M),
    \]
    satisfies
    \begin{equation}
    \label{eq:full-global-tube}
    d(x_k,\mathcal M)
    \le
    \bar\kappa^k d(x_0,\mathcal M)
    +
    \frac{1-\bar\kappa^k}{1-\bar\kappa}
    \varepsilon_{\mathcal M},
    \qquad
    k\ge 0 .
    \end{equation}
\end{enumerate}
\end{theorem}

We now prove Theorem~\ref{thm:full-residual-manifold-global}. The proof separates the local-to-global mechanism into three steps: full residual probes imply normal operator control on visited states; coverage extends this control to the entire predictive tube; and transverse contraction yields global-in-time tube stability.

\begin{lemma}[Full residual probes imply normal operator control on visited states]
\label{lem:full-residual-to-normal-operator}
Let \(q\in U_\rho(\mathcal M)\), let \(\Pi^\perp(q)\) be an orthogonal projector, and let \(\delta\) be a random vector with conditional second moment
\[
M_\delta(q):=\mathbb E[\delta\delta^\top\mid q].
\]
If
\[
M_\delta(q)\succeq \mu\Pi^\perp(q)
\qquad \text{for some }\mu>0,
\]
then, for every matrix \(J\),
\begin{equation}
\label{eq:lem-full-residual-to-normal-operator}
\mathbb E[\|J\delta\|_2^2\mid q]
\ge
\mu\|J\Pi^\perp(q)\|_F^2 .
\end{equation}
\end{lemma}

\begin{proof}
We compute
\[
\mathbb E[\|J\delta\|_2^2\mid q]
=
\mathbb E[\delta^\top J^\top J\delta\mid q]
=
\operatorname{tr}(J^\top J M_\delta(q)).
\]
Since \(J^\top J\succeq 0\) and \(M_\delta(q)\succeq \mu\Pi^\perp(q)\),
\[
\operatorname{tr}(J^\top J M_\delta(q))
\ge
\mu\operatorname{tr}(J^\top J\Pi^\perp(q)).
\]
Because \(\Pi^\perp(q)\) is an orthogonal projector,
\[
\operatorname{tr}(J^\top J\Pi^\perp(q))
=
\operatorname{tr}\bigl((J\Pi^\perp(q))^\top(J\Pi^\perp(q))\bigr)
=
\|J\Pi^\perp(q)\|_F^2.
\]
This proves \eqref{eq:lem-full-residual-to-normal-operator}.
\end{proof}

\begin{lemma}[Coverage lifts visited-state control to the whole tube]
\label{lem:coverage-lift-full-residual}
Let \(U\subset\mathbb R^m\), let \(h:U\to\mathbb R\) be \(L_h\)-Lipschitz, and let \(\mathcal Q\subset U\) be a \(\tau\)-net of \(U\). Then
\begin{equation}
\label{eq:lem-coverage-lift-full-residual}
\sup_{x\in U}h(x)
\le
\sup_{q\in\mathcal Q}h(q)+L_h\tau .
\end{equation}
\end{lemma}

\begin{proof}
Fix \(x\in U\). Since \(\mathcal Q\) is a \(\tau\)-net, there exists \(q\in\mathcal Q\) such that \(\|x-q\|_2\le \tau\). By Lipschitz continuity,
\[
h(x)\le h(q)+L_h\|x-q\|_2\le h(q)+L_h\tau .
\]
Taking the supremum over \(x\in U\) proves the claim.
\end{proof}

\begin{lemma}[Quadratic tangential lifting]
\label{lem:quadratic-tangential-lifting-full-residual}
Let \(\mathcal M\subset\mathbb R^m\) be a compact \(C^2\) embedded submanifold. Then there exist constants \(\rho_T>0\) and \(C_T>0\) such that for every \(m\in\mathcal M\) and every \(t\in T_m\mathcal M\) with \(\|t\|_2\le \rho_T\), there exists \(\widetilde m\in\mathcal M\) satisfying
\begin{equation}
\label{eq:lem-tangential-lifting-full-residual}
\|\widetilde m-(m+t)\|_2\le C_T\|t\|_2^2 .
\end{equation}
\end{lemma}

\begin{proof}
Since \(\mathcal M\) is compact and \(C^2\), it admits a finite atlas with uniformly bounded second derivatives. Equivalently, after a rigid motion depending smoothly on \(m\), a neighborhood of each \(m\in\mathcal M\) can be represented as the graph of a \(C^2\) map
\[
\psi_m:B_{\rho_T}^d(0)\to\mathbb R^{m-d}
\]
over \(T_m\mathcal M\), with
\[
\psi_m(0)=0,
\qquad
D\psi_m(0)=0,
\qquad
\sup_m\|D^2\psi_m\|\le K_T
\]
for uniform constants \(\rho_T>0\) and \(K_T<\infty\). For \(t\in T_m\mathcal M\) with \(\|t\|_2\le \rho_T\), identify \(t\) with its coordinate vector \(u\in\mathbb R^d\) and define
\[
\widetilde m:=m+u+\psi_m(u).
\]
Then \(\widetilde m\in\mathcal M\), and Taylor's theorem gives
\[
\|\psi_m(u)\|_2\le \frac{K_T}{2}\|u\|_2^2.
\]
Therefore
\[
\|\widetilde m-(m+t)\|_2=\|\psi_m(u)\|_2\le \frac{K_T}{2}\|t\|_2^2.
\]
Taking \(C_T:=K_T/2\) proves the result.
\end{proof}

\begin{proof}[Proof of Theorem~\ref{thm:full-residual-manifold-global}]
We prove the three claims in order.

\paragraph{Proof of \textup{(i)}.}
Fix a visited state \(q\in\mathcal Q\) and write \(J=Jg_\theta(q)\). By Assumptions~A2--A3 and Lemma~\ref{lem:full-residual-to-normal-operator},
\[
\eta_0
\ge
\mathbb E[\|J\delta_t\|_2^2\mid q_t=q]
\ge
\mu\|J\Pi^\perp(q)\|_F^2.
\]
Thus every visited state satisfies
\begin{equation}
\label{eq:visited-normal-bound-full-residual}
\|Jg_\theta(q)\Pi^\perp(q)\|_F
\le
\sqrt{\eta_0/\mu}.
\end{equation}
Now define \(h_N(x):=\|Jg_\theta(x)\Pi^\perp(x)\|_F\). By Assumption~A4, \(h_N\) is \(L_h\)-Lipschitz on \(U_\rho(\mathcal M)\), and \(\mathcal Q\) is a \(\tau\)-net of \(U_\rho(\mathcal M)\). Lemma~\ref{lem:coverage-lift-full-residual} and \eqref{eq:visited-normal-bound-full-residual} imply
\[
\sup_{x\in U_\rho(\mathcal M)} h_N(x)
\le
\sup_{q\in\mathcal Q} h_N(q)+L_h\tau
\le
\sqrt{\eta_0/\mu}+L_h\tau
=\gamma_N.
\]
This proves \eqref{eq:full-normal-operator-control}.

\paragraph{Proof of \textup{(ii)}.}
Fix \(x\in U_\rho(\mathcal M)\), and let
\[
z:=\pi(x)\in\mathcal M,
\qquad
n:=x-z\in N_z\mathcal M.
\]
Since \(x\) lies in the tubular neighborhood, \(\|n\|_2=d(x,\mathcal M)\). Set
\[
y:=\Phi_\theta(z),
\qquad
m:=\pi(y)\in\mathcal M.
\]
The point \(m\) is well-defined by Assumption~A1.

Because \(\Phi_\theta\in C^2\) on the compact tube, Taylor's theorem gives
\begin{equation}
\label{eq:taylor-full-residual-proof}
\Phi_\theta(x)
=
\Phi_\theta(z)+D\Phi_\theta(z)n+R_z(n),
\qquad
\|R_z(n)\|_2\le C_1\|n\|_2^2,
\end{equation}
for some constant \(C_1<\infty\) independent of \(x\).

Decompose the linear term relative to \(T_m\mathcal M\oplus N_m\mathcal M\):
\[
D\Phi_\theta(z)n=t+s,
\qquad
 t\in T_m\mathcal M,
\quad
 s\in N_m\mathcal M,
\]
where
\[
t=(I-\Pi^\perp(y))D\Phi_\theta(z)n,
\qquad
s=\Pi^\perp(y)D\Phi_\theta(z)n.
\]
Since \(\Phi_\theta\in C^1\) on the compact tube,
\[
M_\Phi:=\sup_{u\in U_\rho(\mathcal M)}\|D\Phi_\theta(u)\|<\infty,
\]
and hence \(\|t\|_2\le M_\Phi\|n\|_2\). Shrinking \(\rho\) if necessary, we may assume \(\|t\|_2\le \rho_T\), so Lemma~\ref{lem:quadratic-tangential-lifting-full-residual} applies. Thus there exists \(\widetilde m\in\mathcal M\) such that
\begin{equation}
\label{eq:lift-full-residual-proof}
\|\widetilde m-(m+t)\|_2\le C_T\|t\|_2^2.
\end{equation}
Using \eqref{eq:taylor-full-residual-proof} and \eqref{eq:lift-full-residual-proof},
\begin{align}
 d(\Phi_\theta(x),\mathcal M)
 &\le \|\Phi_\theta(x)-\widetilde m\|_2 \notag\\
 &\le \|y-m\|_2+\|s\|_2+\|R_z(n)\|_2+\|\widetilde m-(m+t)\|_2 .
\label{eq:distance-split-full-residual-proof}
\end{align}
The first term is bounded by the manifold defect:
\[
\|y-m\|_2=d(\Phi_\theta(z),\mathcal M)\le \varepsilon_{\mathcal M}.
\]

We next control \(\|s\|_2\). Since \(n\in N_z\mathcal M\), \(n=\Pi^\perp(z)n\). By the chain rule,
\[
D\Phi_\theta(z)
=
D_1S(z,g_\theta(z))+D_2S(z,g_\theta(z))Jg_\theta(z).
\]
Therefore,
\begin{align*}
\|s\|_2
&=
\|\Pi^\perp(y)D\Phi_\theta(z)n\|_2 \\
&\le
\left\|
\Pi^\perp(y)D_1S(z,g_\theta(z))\Pi^\perp(z)
\right\|\|n\|_2 \\
&\quad +
\left\|
\Pi^\perp(y)D_2S(z,g_\theta(z))
\right\|
\|Jg_\theta(z)\Pi^\perp(z)n\|_2 \\
&\le
\left(a_N+b_N\|Jg_\theta(z)\Pi^\perp(z)\|\right)\|n\|_2 .
\end{align*}
By part~\textup{(i)},
\[
\|Jg_\theta(z)\Pi^\perp(z)\|
\le
\|Jg_\theta(z)\Pi^\perp(z)\|_F
\le
\gamma_N.
\]
Hence
\begin{equation}
\label{eq:s-bound-full-residual-proof}
\|s\|_2
\le
(a_N+b_N\gamma_N)\|n\|_2
=
\kappa_N d(x,\mathcal M).
\end{equation}
Finally, \eqref{eq:taylor-full-residual-proof}, \eqref{eq:lift-full-residual-proof}, and \(\|t\|_2\le M_\Phi\|n\|_2\) imply
\[
\|R_z(n)\|_2+\|\widetilde m-(m+t)\|_2
\le
(C_1+C_T M_\Phi^2)\|n\|_2^2.
\]
Combining this with \eqref{eq:distance-split-full-residual-proof} and \eqref{eq:s-bound-full-residual-proof} gives
\[
d(\Phi_\theta(x),\mathcal M)
\le
\varepsilon_{\mathcal M}
+
\kappa_N d(x,\mathcal M)
+
C_\rho d(x,\mathcal M)^2,
\]
where one may take \(C_\rho:=C_1+C_T M_\Phi^2\). This proves \eqref{eq:full-local-recursion}.

\paragraph{Proof of \textup{(iii)}.}
Let \(x_{k+1}=\Phi_\theta(x_k)\) and define
\[
d_k:=d(x_k,\mathcal M).
\]
Suppose \(d_k\le \rho\). By part~\textup{(ii)},
\[
d_{k+1}
\le
\varepsilon_{\mathcal M}+
\kappa_N d_k+C_\rho d_k^2
\le
\varepsilon_{\mathcal M}+(
\kappa_N+C_\rho\rho)d_k
=
\varepsilon_{\mathcal M}+\bar\kappa d_k.
\]
Since \(d_k\le\rho\) and \(\varepsilon_{\mathcal M}\le(1-\bar\kappa)\rho\),
\[
d_{k+1}
\le
\varepsilon_{\mathcal M}+\bar\kappa\rho
\le
(1-\bar\kappa)\rho+\bar\kappa\rho
=
\rho.
\]
Thus \(d_k\le\rho\) implies \(d_{k+1}\le\rho\). Since \(x_0\in U_\rho(\mathcal M)\), induction shows that \(U_\rho(\mathcal M)\) is forward invariant under \(\Phi_\theta\).

The same one-step bound gives the affine recursion
\[
d_{k+1}\le \varepsilon_{\mathcal M}+\bar\kappa d_k.
\]
Iterating it yields
\[
d_k
\le
\bar\kappa^k d_0+
\sum_{j=0}^{k-1}\bar\kappa^j\varepsilon_{\mathcal M}
=
\bar\kappa^k d_0+
\frac{1-\bar\kappa^k}{1-\bar\kappa}\varepsilon_{\mathcal M},
\]
which proves \eqref{eq:full-global-tube}. This completes the proof.
\end{proof}

\begin{remark}[Role of the assumptions]
\label{rem:full-residual-assumptions}
The proof separates the guarantee into three bridges. Assumptions~A2--A3 convert the full residual-direction penalty into operator control on the normal response. Assumption~A4 extends this control from visited states to the entire predictive tube. Assumptions~A1 and A5 then convert the normal operator bound into a transverse error recursion, which bootstraps to global-in-time predictive-tube stability.
\end{remark}

\subsection{Local rollout consistency controls the transverse Lyapunov exponent}

\begin{corollary}[Negative transverse Lyapunov exponent]
\label{cor:transverse_lyapunov}
Assume the conditions of Theorem~\ref{thm:full-residual-manifold-global}.
In addition, suppose that \(\mathcal M\) is invariant under the free-running map
\(\Phi_\theta\), so that \(\epsilon_M=0\).
For \(z\in\mathcal M\), define the normal linear cocycle
\[
A_\theta(z)
:=
\Pi^\perp(\Phi_\theta(z))
D\Phi_\theta(z)
\Pi^\perp(z).
\]
Let \(z_{k+1}=\Phi_\theta(z_k)\), and define the largest transverse Lyapunov
exponent by
\[
\lambda_\perp(z_0)
:=
\limsup_{k\to\infty}
\frac{1}{k}
\log
\left\|
A_\theta(z_{k-1})\cdots A_\theta(z_0)
\right\|.
\]
If
\[
\kappa_N := a_N + b_N\gamma_N < 1,
\]
then
\[
\lambda_\perp(z_0)
\leq
\log \kappa_N
<0
\]
for every \(z_0\in\mathcal M\).
Consequently, the predictive manifold is transversely exponentially stable.
\end{corollary}

\begin{proof}
By the chain rule,
\[
D\Phi_\theta(z)
=
D_1S(z,g_\theta(z))
+
D_2S(z,g_\theta(z))Jg_\theta(z).
\]
Projecting onto the normal bundle gives
\[
\|A_\theta(z)\|
\leq
a_N + b_N
\|Jg_\theta(z)\Pi^\perp(z)\|.
\]
The residual-excitation and consistency assumptions imply
\[
\sup_{z\in\mathcal M}
\|Jg_\theta(z)\Pi^\perp(z)\|
\leq
\gamma_N.
\]
Therefore
\[
\sup_{z\in\mathcal M}
\|A_\theta(z)\|
\leq
a_N+b_N\gamma_N
=
\kappa_N.
\]
For any trajectory \(z_{k+1}=\Phi_\theta(z_k)\) on \(\mathcal M\),
\[
\left\|
A_\theta(z_{k-1})\cdots A_\theta(z_0)
\right\|
\leq
\prod_{i=0}^{k-1}\|A_\theta(z_i)\|
\leq
\kappa_N^k.
\]
Taking \(k^{-1}\log(\cdot)\) and then \(\limsup\) yields
\[
\lambda_\perp(z_0)
\leq
\log \kappa_N
<0.
\]
\end{proof}

\begin{remark}[Practical transverse stability]
When \(\epsilon_M>0\), the same argument gives practical rather than exact
transverse stability.  In a tube \(U_\rho(\mathcal M)\), the distance recursion
\[
d(x_{k+1},\mathcal M)
\leq
\epsilon_M+\bar\kappa d(x_k,\mathcal M),
\qquad
\bar\kappa<1,
\]
implies
\[
\limsup_{k\to\infty}
d(x_k,\mathcal M)
\leq
\frac{\epsilon_M}{1-\bar\kappa}.
\]
Thus approximate invariance produces an attracting tube around the predictive
manifold, rather than convergence exactly onto \(\mathcal M\).
\end{remark}

\subsection{Composite versus Branchwise Rollout Consistency}
\label{sec:composite-branchwise-consistency}

Many forecasting architectures can be written as a sum of complementary components. A common example is a hybrid linear--nonlinear model
\begin{equation}
\label{eq:hybrid-two-branch}
g_\theta(x)
=
g_{\mathrm{lin}}(x)
+
g_{\mathrm{nl}}(x),
\end{equation}
where the linear branch captures low-frequency or trend-like structure and the nonlinear branch captures residual interactions. More generally, we consider a \(B\)-branch additive forecaster
\begin{equation}
\label{eq:multi-branch-model}
g_\theta(x)
=
\sum_{b=1}^{B} g_b(x).
\end{equation}
The rollout-consistency penalty can then be applied in two different ways.

Let
\[
q_t := S(x_t,g_\theta(x_t)),
\qquad
\delta_t := P(y_t-g_\theta(x_t))
\]
be the free-running next state and the residual-induced perturbation. Define the local response of branch \(b\) by
\begin{equation}
\label{eq:branch-response}
z_b(q_t,\delta_t)
:=
Jg_b(q_t)\delta_t.
\end{equation}
For brevity, we write \(z_b=z_b(q_t,\delta_t)\). The \emph{joint} consistency penalty treats the whole model as a single map:
\begin{equation}
\label{eq:joint-penalty}
\mathcal R_{\mathrm{joint}}
:=
\left\|
\sum_{b=1}^{B} z_b
\right\|_2^2
=
\left\|
Jg_\theta(q_t)\delta_t
\right\|_2^2.
\end{equation}
In contrast, the \emph{branchwise} consistency penalty suppresses the response of each branch separately:
\begin{equation}
\label{eq:branchwise-penalty}
\mathcal R_{\mathrm{br}}
:=
\sum_{b=1}^{B}
\|z_b\|_2^2.
\end{equation}
Although both penalties are motivated by rollout consistency, they impose different inductive biases. The joint penalty controls only the final composite response of the model. The branchwise penalty is stronger: it prevents the composite response from becoming small solely through cancellation among branches.

\begin{proposition}[Cancellation decomposition]
\label{prop:cancellation-decomposition}
Let \(z_1,\dots,z_B\in\mathbb R^p\), and define
\[
\bar z := \frac{1}{B}\sum_{b=1}^{B} z_b.
\]
Then
\begin{equation}
\label{eq:cancellation-decomposition}
\sum_{b=1}^{B}\|z_b\|_2^2
=
\frac{1}{B}
\left\|
\sum_{b=1}^{B}z_b
\right\|_2^2
+
\sum_{b=1}^{B}
\|z_b-\bar z\|_2^2.
\end{equation}
Equivalently,
\begin{equation}
\label{eq:cancellation-pairwise}
\mathcal R_{\mathrm{br}}
=
\frac{1}{B}\mathcal R_{\mathrm{joint}}
+
\frac{1}{B}
\sum_{1\le i<j\le B}
\|z_i-z_j\|_2^2.
\end{equation}
\end{proposition}

\begin{proof}
The standard bias--variance identity gives
\[
\sum_{b=1}^{B}\|z_b\|_2^2
=
B\|\bar z\|_2^2
+
\sum_{b=1}^{B}\|z_b-\bar z\|_2^2.
\]
Since \(B\bar z=\sum_{b=1}^{B}z_b\),
\[
B\|\bar z\|_2^2
=
\frac{1}{B}
\left\|
\sum_{b=1}^{B}z_b
\right\|_2^2,
\]
which proves \eqref{eq:cancellation-decomposition}. The pairwise form follows from
\[
\sum_{1\le i<j\le B}\|z_i-z_j\|_2^2
=
B\sum_{b=1}^{B}\|z_b-\bar z\|_2^2.
\]
\end{proof}

Proposition~\ref{prop:cancellation-decomposition} shows that branchwise consistency is not merely a rescaled version of joint consistency. It equals a joint consistency term plus a \emph{branch disagreement} term. Thus, branchwise regularization suppresses both the average response of the composite model and the spread of the branch responses around that average.

For the two-branch linear--nonlinear case in \eqref{eq:hybrid-two-branch}, define
\[
z_{\mathrm{lin}}
:=
Jg_{\mathrm{lin}}(q_t)\delta_t,
\qquad
z_{\mathrm{nl}}
:=
Jg_{\mathrm{nl}}(q_t)\delta_t.
\]
Then Proposition~\ref{prop:cancellation-decomposition} yields
\begin{equation}
\label{eq:two-branch-identity}
\|z_{\mathrm{lin}}\|_2^2+\|z_{\mathrm{nl}}\|_2^2
=
\frac12
\|z_{\mathrm{lin}}+z_{\mathrm{nl}}\|_2^2
+
\frac12
\|z_{\mathrm{lin}}-z_{\mathrm{nl}}\|_2^2.
\end{equation}
Therefore, separately penalizing the linear and nonlinear branches is equivalent, up to a factor of \(1/2\), to penalizing the composite rollout response together with an additional term that discourages the two branches from producing opposite residual responses.

\begin{corollary}[Joint consistency does not control branchwise consistency]
\label{cor:joint-does-not-control-branchwise}
For \(B\ge 2\),
\begin{equation}
\label{eq:joint-upper-bound}
\mathcal R_{\mathrm{joint}}
\le
B\,\mathcal R_{\mathrm{br}}.
\end{equation}
However, there is no finite constant \(C\) such that
\begin{equation}
\label{eq:no-reverse-bound}
\mathcal R_{\mathrm{br}}
\le
C\,\mathcal R_{\mathrm{joint}}
\end{equation}
holds uniformly over all branch responses \(z_1,\dots,z_B\).
\end{corollary}

\begin{proof}
The upper bound \eqref{eq:joint-upper-bound} follows from Cauchy's inequality:
\[
\left\|\sum_{b=1}^{B}z_b\right\|_2^2
\le
B\sum_{b=1}^{B}\|z_b\|_2^2.
\]
To see that no reverse bound can hold, take \(B=2\), \(z_1=u\), and \(z_2=-u\) for any nonzero \(u\). Then
\[
\mathcal R_{\mathrm{joint}}
=
\|u-u\|_2^2
=
0,
\qquad
\mathcal R_{\mathrm{br}}
=
2\|u\|_2^2
>
0.
\]
Thus \(\mathcal R_{\mathrm{br}}\) cannot be bounded by any finite multiple of \(\mathcal R_{\mathrm{joint}}\).
\end{proof}

Corollary~\ref{cor:joint-does-not-control-branchwise} formalizes the cancellation issue. A small joint penalty guarantees that the final composite model is locally insensitive to the residual-induced perturbation, but it does not guarantee that the individual branches are insensitive. The composite response may be small only because two large branch responses cancel.

\subsection{Branchwise robustness under perturbations}
\label{subsec:branchwise-robustness}

The distinction between joint and branchwise penalties becomes important when branch responses are not perfectly reproducible. This can happen because of optimization noise, distribution shift, local changes in the nonlinear Jacobian, normalization effects, or architectural gates. To make this precise, consider multiplicative branch perturbations
\[
z_b \mapsto (1+\varepsilon_b)z_b,
\qquad
|\varepsilon_b|\le \rho.
\]
Define the worst-case perturbed response
\begin{equation}
\label{eq:robust-branch-response}
\mathcal R_{\rho}^{\mathrm{rob}}
:=
\sup_{|\varepsilon_b|\le \rho}
\left\|
\sum_{b=1}^{B}(1+\varepsilon_b)z_b
\right\|_2^2.
\end{equation}

\begin{proposition}[Branchwise penalty controls perturbed composite responses]
\label{prop:branchwise-robustness}
For every \(\rho\ge 0\),
\begin{equation}
\label{eq:robust-bound}
\mathcal R_{\rho}^{\mathrm{rob}}
\le
\left(
\sqrt{\mathcal R_{\mathrm{joint}}}
+
\rho\sqrt{B\,\mathcal R_{\mathrm{br}}}
\right)^2.
\end{equation}
In particular, branchwise control bounds the sensitivity of the composite response to branch-level perturbations.
\end{proposition}

\begin{proof}
For any \(\varepsilon_b\) satisfying \(|\varepsilon_b|\le\rho\),
\begin{align*}
\left\|
\sum_{b=1}^{B}(1+\varepsilon_b)z_b
\right\|_2
&\le
\left\|
\sum_{b=1}^{B}z_b
\right\|_2
+
\left\|
\sum_{b=1}^{B}\varepsilon_b z_b
\right\|_2 \\
&\le
\sqrt{\mathcal R_{\mathrm{joint}}}
+
\sum_{b=1}^{B}|\varepsilon_b|\,\|z_b\|_2 \\
&\le
\sqrt{\mathcal R_{\mathrm{joint}}}
+
\rho\sum_{b=1}^{B}\|z_b\|_2 \\
&\le
\sqrt{\mathcal R_{\mathrm{joint}}}
+
\rho\sqrt{B\sum_{b=1}^{B}\|z_b\|_2^2} \\
&=
\sqrt{\mathcal R_{\mathrm{joint}}}
+
\rho\sqrt{B\,\mathcal R_{\mathrm{br}}}.
\end{align*}
Taking the supremum over all admissible \(\varepsilon_b\) and squaring proves the result.
\end{proof}

The bound in Proposition~\ref{prop:branchwise-robustness} explains why branchwise penalties may be more stable than joint penalties when cancellation is fragile. For example, in the two-branch case with \(z_1=u\) and \(z_2=-u\), the unperturbed joint response is zero. However, choosing \(\varepsilon_1=\rho\) and \(\varepsilon_2=-\rho\) gives
\[
\left\|
(1+\rho)u-(1-\rho)u
\right\|_2^2
=
4\rho^2\|u\|_2^2.
\]
Thus a model may appear perfectly consistent under the joint penalty while remaining highly sensitive to small branch-level mismatches.

\subsection{Operator-level consequences under residual excitation}
\label{subsec:operator-consequences-branchwise}

The same distinction also appears at the operator level. Let
\[
J_b(q):=Jg_b(q),
\qquad
J_\Sigma(q):=\sum_{b=1}^{B}J_b(q)=Jg_\theta(q).
\]
Let \(\Pi^\perp(q)\) denote the normal projector associated with a predictive manifold, used here only for analysis. Suppose the residual-induced perturbation \(\delta_t\) has conditional second moment
\begin{equation}
\label{eq:branchwise-second-moment}
M_\delta(q)
:=
\mathbb E[\delta_t\delta_t^\top\mid q_t=q].
\end{equation}
The following result compares what joint and branchwise penalties control under the same excitation condition.

\begin{proposition}[Composite versus branchwise normal-operator control]
\label{prop:operator-joint-vs-branchwise}
Assume that for some \(\mu>0\),
\begin{equation}
\label{eq:branchwise-normal-excitation}
M_\delta(q)\succeq \mu\,\Pi^\perp(q).
\end{equation}
Then the following hold.

\begin{enumerate}
    \item[\textup{(i)}] If the joint directional penalty satisfies
    \begin{equation}
    \label{eq:joint-directional-bound}
    \mathbb E\!\left[
    \|J_\Sigma(q)\delta_t\|_2^2
    \,\middle|\, q_t=q
    \right]
    \le
    \eta_{\mathrm{joint}},
    \end{equation}
    then
    \begin{equation}
    \label{eq:joint-operator-bound}
    \|J_\Sigma(q)\Pi^\perp(q)\|_F^2
    \le
    \eta_{\mathrm{joint}}/\mu.
    \end{equation}

    \item[\textup{(ii)}] If the branchwise directional penalty satisfies
    \begin{equation}
    \label{eq:branchwise-directional-bound}
    \mathbb E\!\left[
    \sum_{b=1}^{B}
    \|J_b(q)\delta_t\|_2^2
    \,\middle|\, q_t=q
    \right]
    \le
    \eta_{\mathrm{br}},
    \end{equation}
    then
    \begin{equation}
    \label{eq:branchwise-operator-bound}
    \sum_{b=1}^{B}
    \|J_b(q)\Pi^\perp(q)\|_F^2
    \le
    \eta_{\mathrm{br}}/\mu.
    \end{equation}
\end{enumerate}
Moreover, the joint bound \eqref{eq:joint-operator-bound} does not imply any uniform bound on the individual quantities \(\|J_b(q)\Pi^\perp(q)\|_F\).
\end{proposition}

\begin{proof}
For part~\textup{(i)}, write \(J=J_\Sigma(q)\). Then
\begin{align*}
\mathbb E\!\left[
\|J\delta_t\|_2^2
\,\middle|\, q_t=q
\right]
&=
\operatorname{tr}\!\left(J^\top J M_\delta(q)\right) \\
&\ge
\mu\operatorname{tr}\!\left(J^\top J \Pi^\perp(q)\right) \\
&=
\mu\|J\Pi^\perp(q)\|_F^2,
\end{align*}
where the inequality uses \eqref{eq:branchwise-normal-excitation}. Combining this with \eqref{eq:joint-directional-bound} proves \eqref{eq:joint-operator-bound}.

For part~\textup{(ii)}, applying the same argument to each \(J_b(q)\) and summing over \(b\) gives
\[
\mathbb E\!\left[
\sum_{b=1}^{B}
\|J_b(q)\delta_t\|_2^2
\,\middle|\, q_t=q
\right]
\ge
\mu
\sum_{b=1}^{B}
\|J_b(q)\Pi^\perp(q)\|_F^2.
\]
Combining this with \eqref{eq:branchwise-directional-bound} proves \eqref{eq:branchwise-operator-bound}.

Finally, the absence of an individual bound from joint control follows by cancellation. For \(B=2\), take \(J_1(q)=A\) and \(J_2(q)=-A\), where \(A\) is arbitrary. Then
\[
J_\Sigma(q)=J_1(q)+J_2(q)=0,
\]
so the joint quantity is identically zero, while
\[
\|J_1(q)\Pi^\perp(q)\|_F
=
\|A\Pi^\perp(q)\|_F
\]
can be arbitrarily large.
\end{proof}

Proposition~\ref{prop:operator-joint-vs-branchwise} clarifies the theoretical role of the two penalties. Joint consistency controls the normal response of the \emph{composite} rollout map, which is the quantity needed for stability of the final predictor. Branchwise consistency provides a stronger sufficient condition: it controls the normal response of every component and therefore rules out stability that depends on exact cancellation among branches.

\subsection{An interpolation between joint and branchwise control}
\label{subsec:interpolated-branchwise-control}

The preceding results suggest a continuous family of penalties interpolating between composite consistency and branchwise robustness. Define the branch disagreement energy
\begin{equation}
\label{eq:branch-disagreement-energy}
\mathcal D_{\mathrm{br}}
:=
\sum_{b=1}^{B}\|z_b-\bar z\|_2^2.
\end{equation}
For \(\beta\ge 0\), consider
\begin{equation}
\label{eq:interpolated-penalty}
\mathcal R_{\beta}
:=
\mathcal R_{\mathrm{joint}}
+
\beta B\,\mathcal D_{\mathrm{br}}.
\end{equation}
When \(\beta=0\), \(\mathcal R_\beta\) reduces to the joint penalty. When \(\beta=1\),
\[
\mathcal R_{\beta}
=
B\,\mathcal R_{\mathrm{br}},
\]
by Proposition~\ref{prop:cancellation-decomposition}. Thus \(\beta\) controls how strongly the training objective discourages cancellation among branches.

For the two-branch linear--nonlinear model, this becomes
\begin{equation}
\label{eq:two-branch-interpolation}
\mathcal R_{\beta}
=
\|z_{\mathrm{lin}}+z_{\mathrm{nl}}\|_2^2
+
\beta
\|z_{\mathrm{lin}}-z_{\mathrm{nl}}\|_2^2.
\end{equation}
The first term enforces local rollout consistency of the final predictor, while the second term controls the disagreement between the linear and nonlinear residual responses. This yields a practical tradeoff: small \(\beta\) prioritizes the stability of the composite model, whereas large \(\beta\) discourages fragile cancellation and promotes branchwise robustness.

\begin{remark}[Exact finite-difference version]
\label{rem:finite-difference-branchwise}
All identities above hold verbatim for exact finite-difference consistency. One simply replaces
\[
z_b=Jg_b(q_t)\delta_t
\]
by
\[
\Delta_b
:=
g_b(q_t+\delta_t)-g_b(q_t).
\]
Then the joint finite-difference penalty is
\[
\left\|
\sum_{b=1}^{B}\Delta_b
\right\|_2^2,
\]
and the branchwise finite-difference penalty is
\[
\sum_{b=1}^{B}\|\Delta_b\|_2^2.
\]
The same cancellation decomposition applies. Thus the distinction between joint and branchwise consistency is not an artifact of the first-order Jacobian approximation; it is a structural property of additive multi-branch models.
\end{remark}

\paragraph{Implication.}
The choice between joint and branchwise penalties depends on the desired inductive bias. If the goal is to control the actual rollout behavior of the final forecaster, the joint penalty is the most direct objective. If the goal is to prevent the linear and nonlinear components from hiding large and opposite residual responses, the branchwise penalty is stronger. The interpolated objective in \eqref{eq:interpolated-penalty} provides a compromise between these two regimes.

\section{Detailed Setup and Additional Experiments}
\label{app:exp}

\begingroup
\raggedbottom
\setlength{\intextsep}{8pt plus 2pt minus 2pt}
\setlength{\textfloatsep}{10pt plus 2pt minus 2pt}
\setlength{\floatsep}{8pt plus 2pt minus 2pt}

\subsection{Dataset.}
\label{app:exp-dataset}
We provide a summary of the datasets used in Table~\ref{tab:dataset_summary}. For ETT \citep{zhou2021informer} data, we follow previous works \citep{lin2024sparsetsf,wu2021autoformer} and divide the dataset into training, validation, and test sets with a $6:2:2$ ratio. The remaining datasets are divided into training, validation, and test sets with a $7:1:2$ ratio.

\begin{table}[!htbp]
\caption{Detailed information about the datasets used in this paper.}
\centering
\scriptsize
\begin{tabular}{@{}lcccccccc@{}}
\toprule
Datasets & ETTh1 & ETTh2 & ETTm1 & ETTm2  & Weather & Traffic & Electricity \\
\midrule
No. of Channels & 7 & 7 & 7 & 7 & 21 & 862 & 321 \\
Timesteps & 17420 & 17420 & 69680 & 69680  & 52696 & 17544 & 26304 \\
Frequency & Hourly & Hourly & 15min & 15min  & 10min & Hourly & Hourly \\
Application Domain & Electricity & Electricity & Electricity & Electricity & Weather & Traffic & Electricity \\
\bottomrule
\end{tabular}

\label{tab:dataset_summary}
\end{table}

\subsection{Implementation Details}
\label{app:exp_details}
\paragraph{Main Experiments.}
In our main experiments with consistency loss, we set the lookback window length to 720. This choice allows the model to exploit as much historical information as possible from the time series, provided that overfitting is effectively controlled. It is also consistent with prior work such as \citep{dai2024pdf, wang2025characteristic, lin2024sparsetsf}. We use Adam \citep{kingma2014adam} as the optimizer with a learning rate of $0.0005$. Each experiment is repeated five times, and we report the average performance.

We use a residual MLP as the time-series backbone with a hidden size of 2048. The MLP branch contains three layers with $\text{ReLU}(\cdot)$ activations and uses neither normalization nor bias terms. The linear shortcut is implemented as a learnable linear layer to match the dimensionality between the lookback window and the forecasting horizon. Figure~\ref{fig:app-res-mlp} visualizes the model architecture.

\begin{figure}[!htbp]
  \centering
    \begin{minipage}[c]{.5\linewidth}
    \centering  \includegraphics[width=\linewidth]{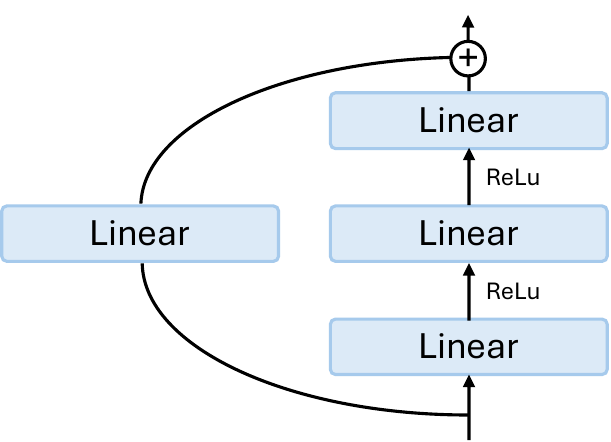}
  \end{minipage}

    \caption{The ResMLP architecture used in our experiments.}
    \label{fig:app-res-mlp}
\end{figure}

For the residual MLP, we apply the consistency loss branch-wise, using separate coefficients for the linear shortcut and the MLP branch; see Appendix~\ref{sec:composite-branchwise-consistency} for a theoretical analysis of this implementation. For the linear shortcut, we fix $\lambda=0.5$. For the MLP branch, we search over $\lambda\in\{0, 0.5, 2, 8, 32, 128\}$.

\paragraph{Backbone-Agnostic Experiments.}
To further test whether consistency loss generalizes to existing time-series backbones, we apply it to FilterNet \citep{yi2024filternet}, TSLANet \citep{eldele2024tslanet}, iTransformer \citep{liu2024itransformer}, and PDF \citep{dai2024pdf} as case studies. We used a lookback window of 720 for all these backbones, similar to our main experiment. The learning rate and optimizer settings also follow from our main experiment.

For PDF, BN uses statistics from the current JVP batch (the full batch or sampled subset), and differentiation includes their dependence on the inputs. The JVP pass does not update running statistics; only the ordinary MSE forward updates them using train-mode BN.

Since not all backbones have an explicit branch-based structure, we apply the consistency loss to the entire model and search over $\lambda\in\{0, 0.125, 0.25, 0.5, 2, 8\}$.

\paragraph{Computing Resources.}
All experiments are implemented in PyTorch \citep{paszke2017automatic} and conducted on a single NVIDIA A100 80GB GPU. We use mean squared error (MSE) as both the training loss and the evaluation metric. Additional details on the experimental settings are provided in the code and running scripts included in the supplementary material.

\paragraph{Baselines.}
We provide details on how the baseline results are obtained. Some baseline repositories are affected by a testing-data bug caused by setting \texttt{drop\_last=True}\footnote{See more details in \texttt{https://github.com/VEWOXIC/FITS/}.}. Therefore, when reproducing PatchTST \citep{nie2023patchtst}, FilterNet \citep{yi2024filternet}, and TSLANet \citep{yi2024filternet} with their official implementations, we first fix this bug. For TSLANet and FilterNet, whose default lookback windows are substantially shorter than ours, we run both their original settings and a 720-lookback-window setting for a fairer comparison. For DUET \citep{qiu2024duet}, AMD \citep{hu2025adaptive}, PDF \citep{dai2024pdf}, SparseTSF \citep{lin2024sparsetsf}, RootPurge \citep{wang2025characteristic}, and TimeDistill \citep{ni2026timedistill}, we report results reproduced from their official implementations.

\subsection{Run-to-run Variance of Main Experiment}
\label{app:exp-r2r-var-main}

Table~\ref{tab:app-r2r-var} reports the mean and standard deviation of the main MLP+CL results over repeated runs. The variances are small across datasets and forecasting horizons, showing that the improvements reported in Table~\ref{table:main_res} are stable rather than driven by a favorable random seed. 

\begin{table}[!htbp]
\centering
\caption{Main results with run-to-run variance included.}
\label{tab:app-r2r-var}
\small
\setlength{\tabcolsep}{5pt}
\renewcommand{\arraystretch}{1.08}
\begin{tabular}{@{}l|cccc@{}}
\toprule
\multirow{2}{*}{\textbf{Dataset}} & \multicolumn{4}{c}{\textbf{Forecasting horizon}} \\
\cmidrule(lr){2-5}
& \textbf{96} & \textbf{192} & \textbf{336} & \textbf{720} \\
\midrule

	\textbf{ETTh1} & \(0.358{\pm}0.002\) & \(0.395{\pm}0.003\) & \(0.434{\pm}0.001\) & \(0.458{\pm}0.005\) \\
	\textbf{ETTh2} & \(0.269{\pm}0.001\) & \(0.331{\pm}0.002\) & \(0.362{\pm}0.001\) & \(0.397{\pm}0.000\) \\
	\textbf{ETTm1} & \(0.282{\pm}0.002\) & \(0.317{\pm}0.002\) & \(0.352{\pm}0.001\) & \(0.410{\pm}0.002\) \\
	\textbf{ETTm2} & \(0.158{\pm}0.001\) & \(0.215{\pm}0.001\) & \(0.268{\pm}0.002\) & \(0.348{\pm}0.001\) \\
    \textbf{Weather} & \(0.142{\pm}0.000 \) & \(0.185{\pm}0.000\) & \(0.235{\pm}0.001\) & \(0.312{\pm}0.001\) \\
    \textbf{ECL} & \(0.127{\pm}0.000\) & \(0.144{\pm}0.001\) & \(0.160{\pm}0.001\) & \(0.198{\pm}0.001\) \\
	\textbf{Traffic} & \(0.358{\pm}0.000\) & \(0.375{\pm}0.000\) & \(0.388{\pm}0.000\) & \(0.428{\pm}0.001\) \\

\bottomrule
\end{tabular}
\end{table}

\subsection{Controlled MLP Baseline}
\label{app:exp-mlp-zero}

To isolate the contribution of consistency loss from that of the MLP architecture, we evaluate the same residual MLP used in Table~\ref{table:main_res} with $\lambda_{\mathrm{MLP}}=0$. The model architecture and all other experimental settings are kept unchanged. In particular, the consistency-loss coefficient for the linear shortcut remains fixed at $0.5$.

Table~\ref{tab:mlp-zero-baseline} reports the test MSE as mean and standard deviation. Compared with this baseline, MLP+CL in Table~\ref{tab:app-r2r-var} achieves a lower or equal MSE in all 28 dataset--horizon settings. These results support the benefit of consistency regularization while keeping model capacity fixed.

\begin{table}[!htbp]
\centering
\caption{Test MSE (mean$\pm$std) of the controlled MLP baseline with $\lambda_{\mathrm{MLP}}=0$. The lookback window is $L=720$.}
\label{tab:mlp-zero-baseline}
\small
\setlength{\tabcolsep}{5pt}
\renewcommand{\arraystretch}{1.08}
\begin{tabular}{@{}l|cccc@{}}
\toprule
\multirow{2}{*}{\textbf{Dataset}} & \multicolumn{4}{c}{\textbf{Forecasting horizon}} \\
\cmidrule(lr){2-5}
& \textbf{96} & \textbf{192} & \textbf{336} & \textbf{720} \\
\midrule
\textbf{ETTh1}   & \(0.400{\pm}0.004\) & \(0.434{\pm}0.001\) & \(0.463{\pm}0.006\) & \(0.510{\pm}0.010\) \\
\textbf{ETTh2}   & \(0.301{\pm}0.003\) & \(0.367{\pm}0.010\) & \(0.395{\pm}0.001\) & \(0.455{\pm}0.009\) \\
\textbf{ETTm1}   & \(0.307{\pm}0.004\) & \(0.349{\pm}0.008\) & \(0.379{\pm}0.004\) & \(0.446{\pm}0.006\) \\
\textbf{ETTm2}   & \(0.182{\pm}0.003\) & \(0.238{\pm}0.004\) & \(0.297{\pm}0.004\) & \(0.381{\pm}0.008\) \\
\textbf{Weather} & \(0.158{\pm}0.003\) & \(0.202{\pm}0.001\) & \(0.258{\pm}0.002\) & \(0.336{\pm}0.003\) \\
\textbf{ECL}     & \(0.129{\pm}0.000\) & \(0.148{\pm}0.000\) & \(0.163{\pm}0.001\) & \(0.199{\pm}0.001\) \\
\textbf{Traffic} & \(0.358{\pm}0.001\) & \(0.375{\pm}0.000\) & \(0.388{\pm}0.001\) & \(0.428{\pm}0.001\) \\
\bottomrule
\end{tabular}
\end{table}

\subsection{Full Impact of Hyperparameter}
\label{app:exp-hyperparameter}
In this section, we show the full set of results for how the choice of $\lambda$ affects the performance of the ResMLP used in our main experiments. As shown in Figure~\ref{fig:app-hyperparameter-full} in most of the settings, a broad value of $\lambda$ value yeilds performance improvement and the improvements are most prominent in smaller datasets.

\begin{figure}[!htb]
  \centering
  \begin{minipage}{0.44\textwidth}
    \centering
    \includegraphics[width=\linewidth]{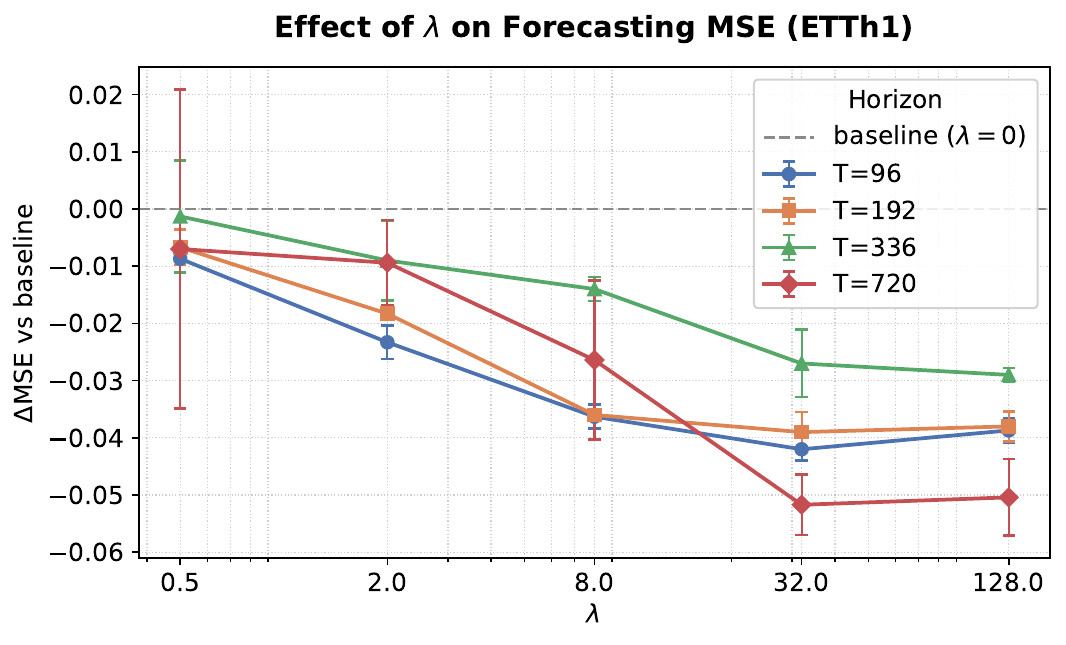}
  \end{minipage}\hspace{0.04\textwidth}
  \begin{minipage}{0.44\textwidth}
    \centering
    \includegraphics[width=\linewidth]{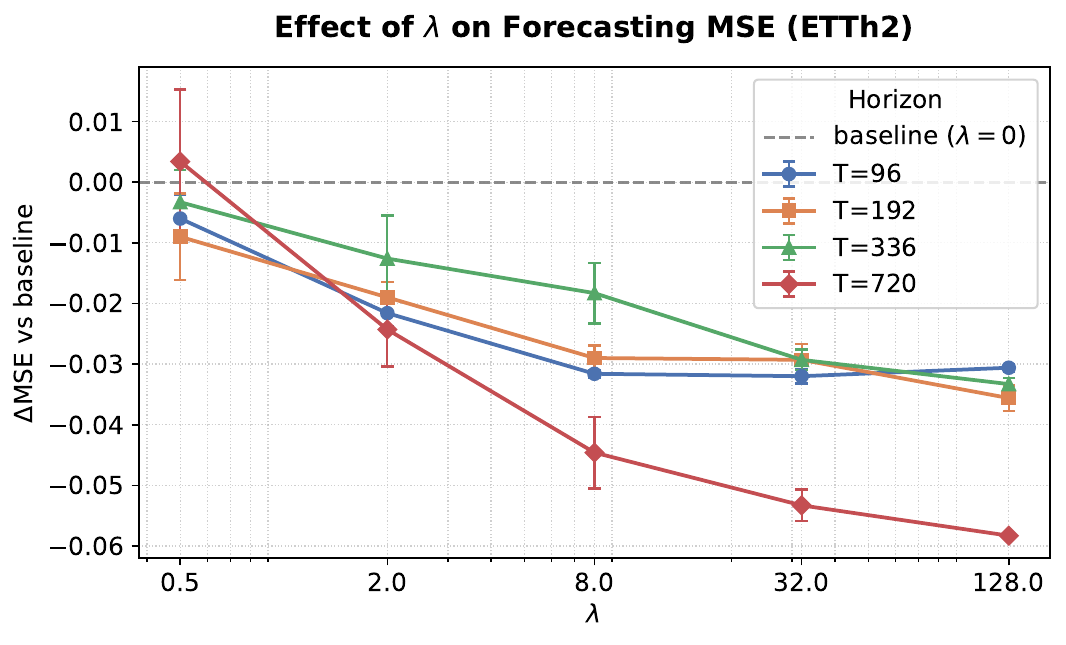}
  \end{minipage}

  \vspace{0.6em}

  \begin{minipage}{0.44\textwidth}
    \centering
    \includegraphics[width=\linewidth]{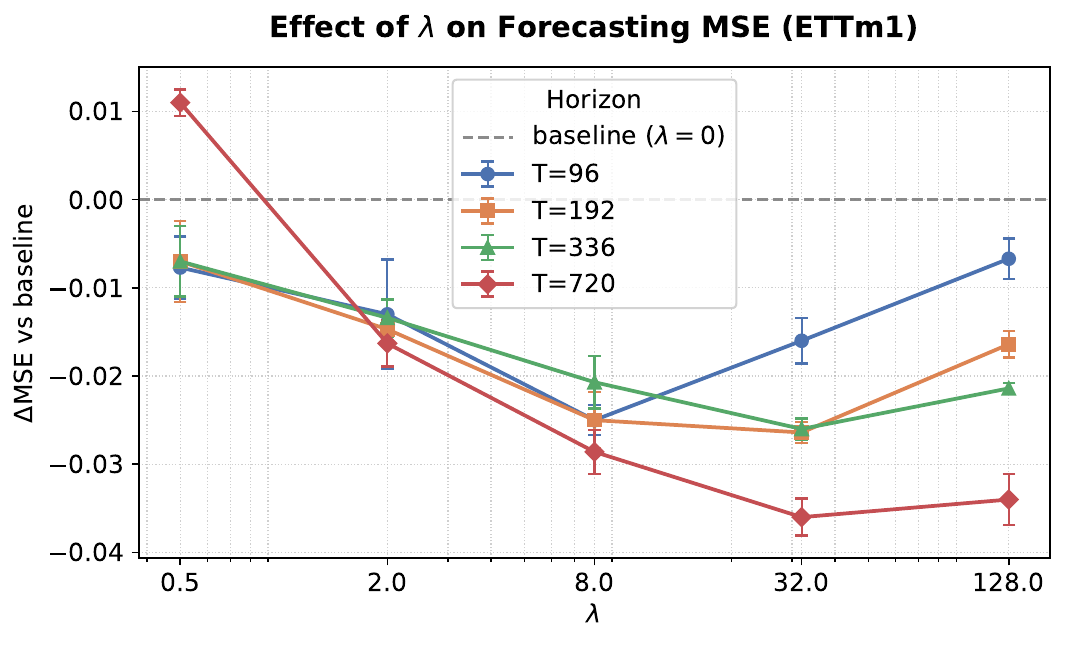}
  \end{minipage}\hspace{0.04\textwidth}
  \begin{minipage}{0.44\textwidth}
    \centering
    \includegraphics[width=\linewidth]{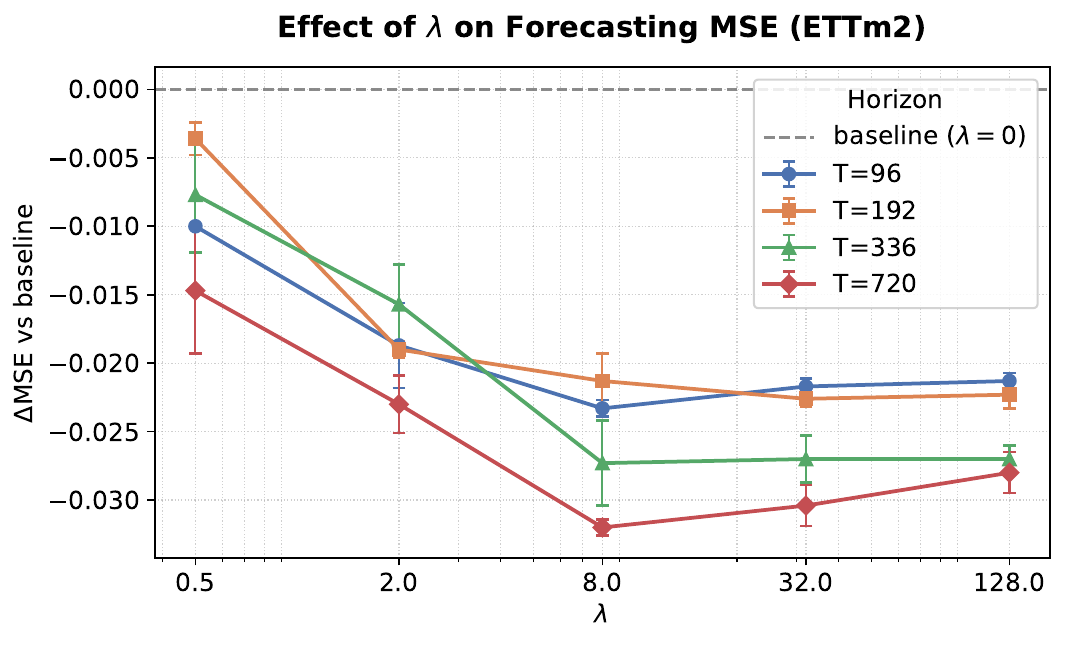}
  \end{minipage}

  \vspace{0.6em}

  \begin{minipage}{0.44\textwidth}
    \centering
    \includegraphics[width=\linewidth]{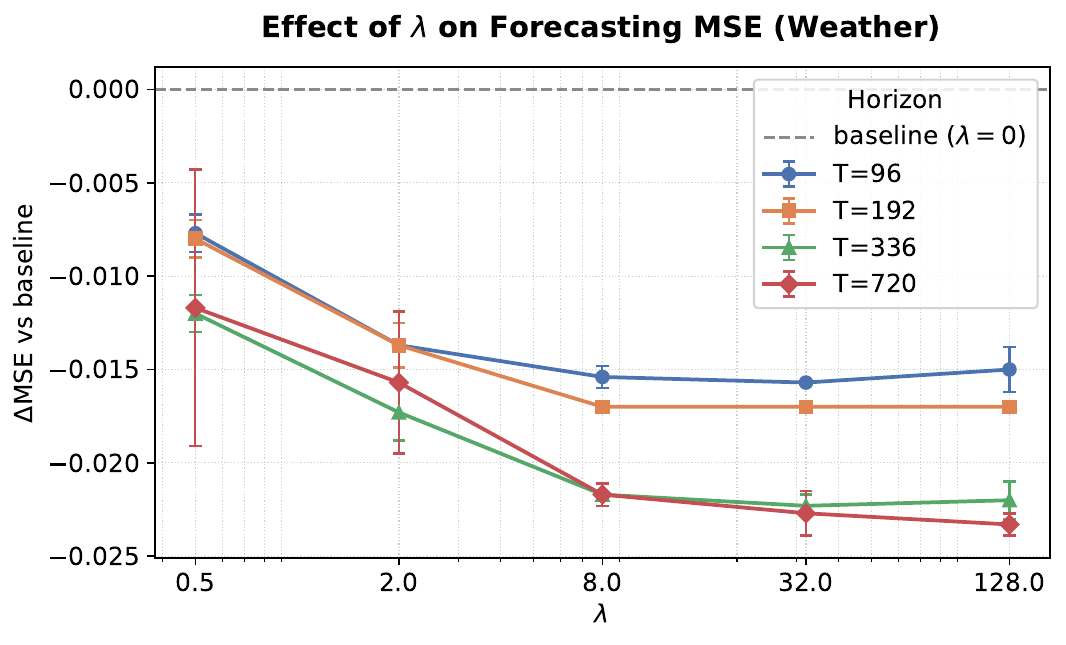}
  \end{minipage}\hspace{0.04\textwidth}
  \begin{minipage}{0.44\textwidth}
    \centering
    \includegraphics[width=\linewidth]{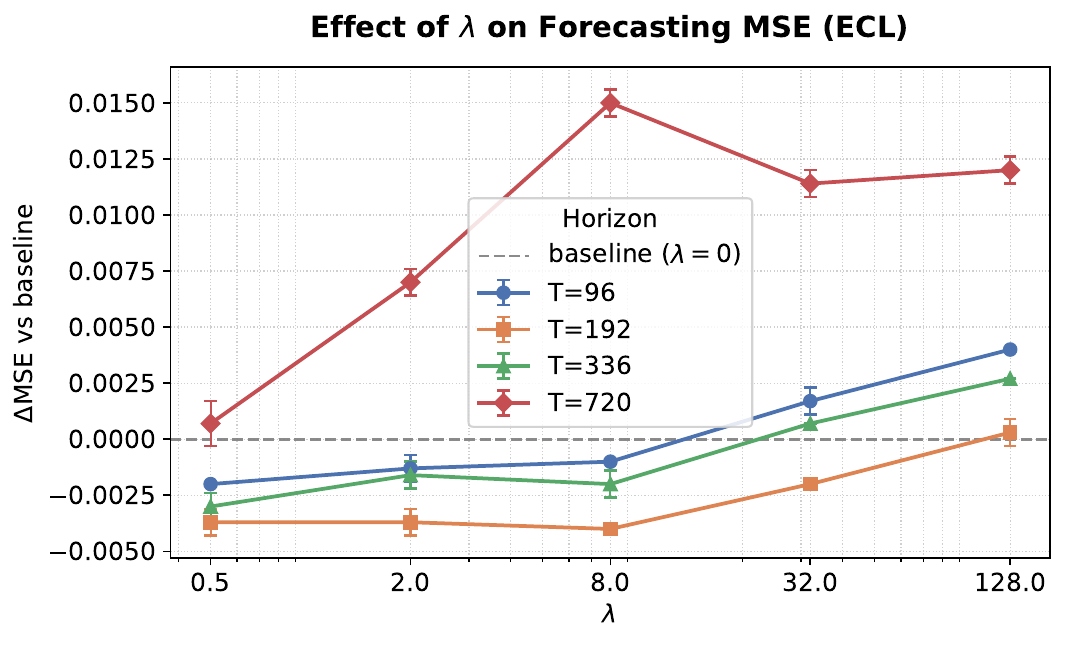}
  \end{minipage}

  \vspace{0.6em}

  \begin{minipage}{0.44\textwidth}
    \centering
    \includegraphics[width=\linewidth]{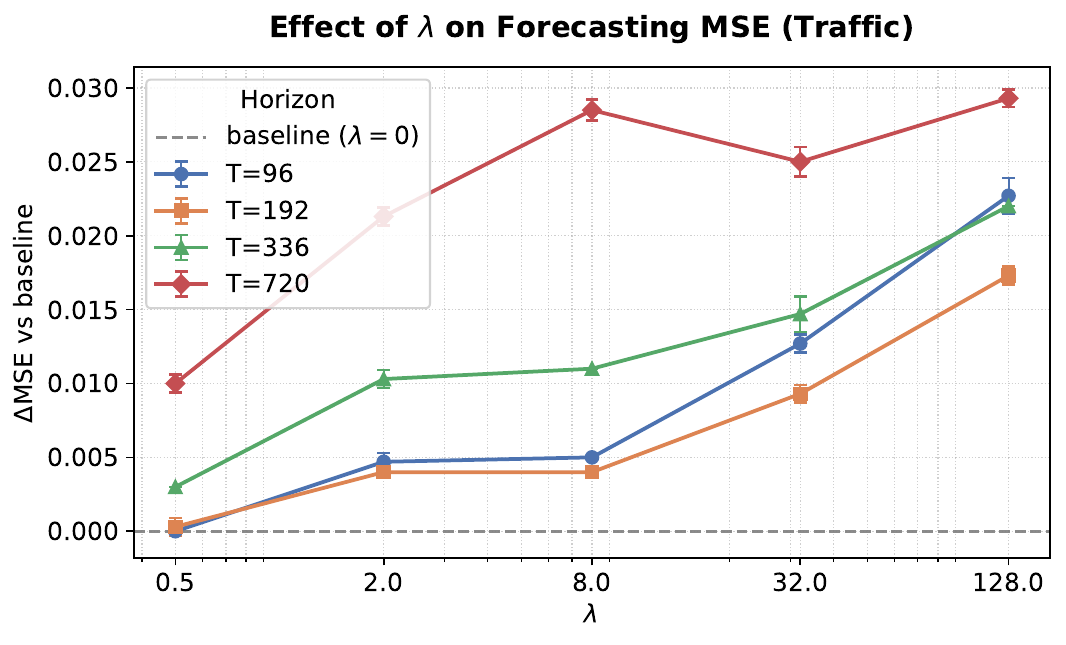}
  \end{minipage}
  \caption{Full hyperparameter sensitivity of the consistency-loss coefficient $\lambda$ for the ResMLP used in the main experiments. Each panel reports test MSE as a function of $\lambda$ for one dataset across forecasting horizons.}
  \label{fig:app-hyperparameter-full}
\end{figure}

\subsection{Consistency Loss - Jacobian Vector Product Alignment Check}
\label{app:exp-cl-jvp-alignment}

The method section motivates the practical consistency loss as a first-order surrogate for the exact teacher-forced/free-running consistency objective. In particular, Proposition~\ref{thm:linear-nonlinear-consistency} shows that the exact finite-difference consistency term
\begin{equation}
\label{eq:app-fd-consistency-check}
\mathcal C_{\mathrm{Full}}(x_t,y_t)
:=
\left\|
g_\theta(x_{t+1}^{\mathrm{TF}})
-
g_\theta(x_{t+1}^{\mathrm{FR}})
\right\|_2^2
\end{equation}
is locally approximated by the Jacobian-vector-product (JVP) quantity
\begin{equation}
\label{eq:app-jvp-consistency-check}
\mathcal C_{\mathrm{JVP}}(x_t,y_t)
:=
\left\|
Jg_\theta(x_{t+1}^{\mathrm{FR}})\,
\delta_t
\right\|_2^2,
\qquad
\delta_t=x_{t+1}^{\mathrm{TF}}-x_{t+1}^{\mathrm{FR}}.
\end{equation}
The practical objective further replaces the full next-window perturbation by the residual-insertion surrogate and evaluates the directional response of $Jg_\theta(x_{t+1}^{\mathrm{FR}})$ with $Jg_\theta(x_t)$,
\begin{equation}
\label{eq:app-practical-jvp-check}
\widehat{\mathcal C}_{\mathrm{JVP}}(x_t,y_t)
:=
\left\|
Jg_\theta(x_t)\,P_{\omega_t}r_t
\right\|_2^2,
\qquad
\omega_t\sim q.
\end{equation}

To assess this alignment, we use ETTh2 and ETTm2 as case studies. At the best-validation checkpoint, we compute, for each training example, the exact consistency loss, the ideal next-state JVP in Equation~\eqref{eq:app-jvp-consistency-check}, and the practical residual-insertion JVP in Equation~\eqref{eq:app-practical-jvp-check}. Because these quantities require matched lookback and forecast lengths, we use only the $L=720, H=720$ setting.

\begin{figure}[!htb]
    \centering
    \includegraphics[width=0.76\textwidth]{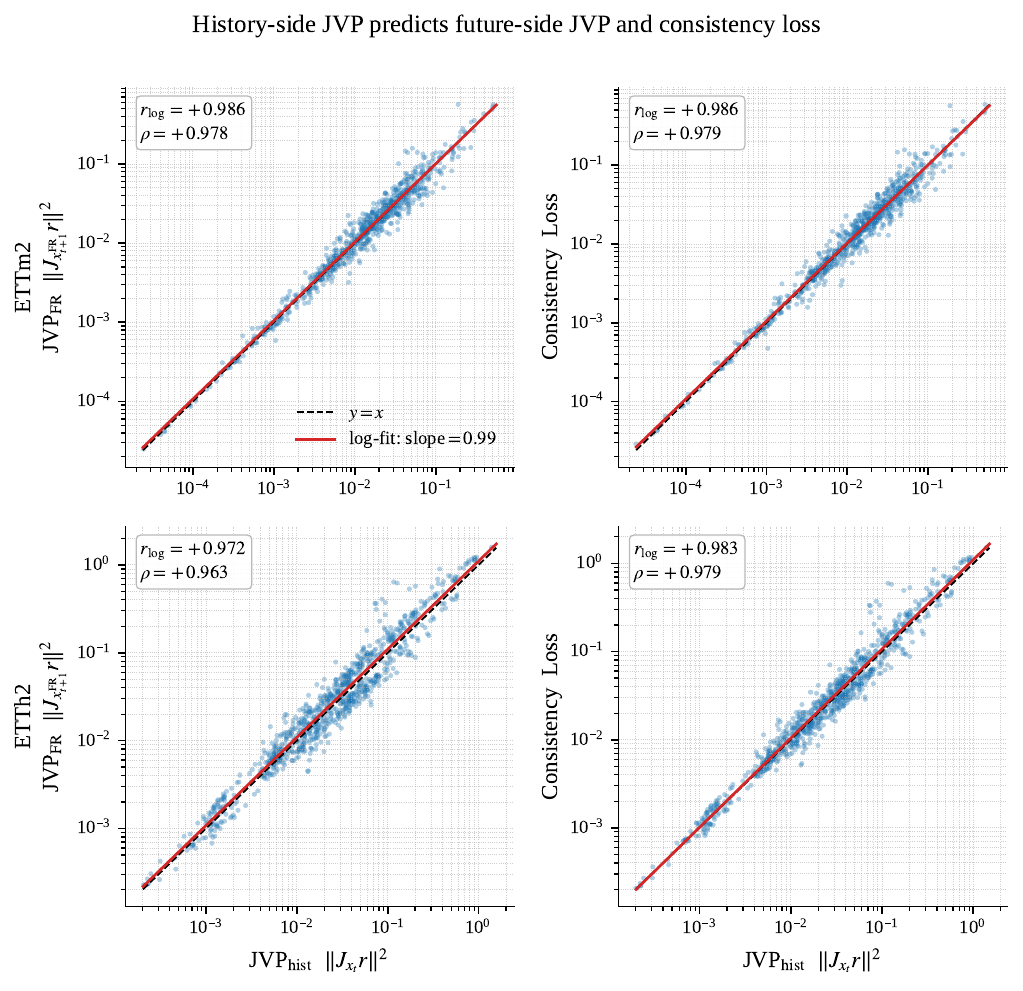}
    \caption{Instance-level JVP alignment in the $L=720, H=720$ setting. Rows show ETTm2 and ETTh2; columns compare the practical JVP at $x_t$ with the ideal next-state JVP and the exact teacher-forced/free-running consistency loss, respectively.}
    \label{fig:app-jvp-alignment}
\end{figure}

Figure~\ref{fig:app-jvp-alignment} shows that the practical JVP is strongly associated with both the exact teacher-forced/free-running discrepancy and the ideal next-state JVP, supporting its use as a training surrogate.

\Needspace{6\baselineskip}
\subsection{Padding Strategy}
\label{app:exp-padding}

Remark~\ref{rmk:residual-domain-alignment} explains the practical consistency loss as a local way to insert the forecast residual back into the input window. In this local view, only the coordinates occupied by the inserted forecast block should change. The residual is therefore placed at the sampled position, and all other coordinates are treated as unchanged, i.e., padded with zeros. This is the reason we use zero-padding in the main experiments. To check whether this design choice matters, we compare it with two alternatives that fill the remaining coordinates with nonzero values: \emph{average padding}, which uses the mean residual value, and \emph{repeat padding}, which repeats the residual block. These alternatives are less directly aligned with the local-insertion argument, but they help test whether the empirical gains depend on the particular padding rule.

\begin{table}[!htbp]
\centering
\caption{Best test MSE achieved by consistency loss under average and repeat padding.}
\label{tab:padding-best-coef-combined}
\scriptsize
\setlength{\tabcolsep}{3pt}
\renewcommand{\arraystretch}{1.08}
\resizebox{\textwidth}{!}{%
\begin{tabular}{@{}lcccc|cccc@{}}
\toprule
\multirow{2}{*}{\textbf{Dataset}} & \multicolumn{4}{c}{\textbf{Avg Padding}} & \multicolumn{4}{c}{\textbf{Repeat Padding}} \\
\cmidrule(lr){2-5} \cmidrule(lr){6-9}
& \textbf{96} & \textbf{192} & \textbf{336} & \textbf{720} & \textbf{96} & \textbf{192} & \textbf{336} & \textbf{720} \\
\midrule
\textbf{ETTh1}   & \(0.363{\pm}0.001\) & \(0.398{\pm}0.002\) & \(0.440{\pm}0.004\) & \(0.458{\pm}0.005\) & \(0.410{\pm}0.005\) & \(0.440{\pm}0.003\) & \(0.453{\pm}0.003\) & \(0.458{\pm}0.005\) \\
\textbf{ETTh2}   & \(0.274{\pm}0.001\) & \(0.332{\pm}0.001\) & \(0.358{\pm}0.001\) & \(0.397{\pm}0.000\) & \(0.308{\pm}0.002\) & \(0.364{\pm}0.003\) & \(0.381{\pm}0.002\) & \(0.397{\pm}0.000\) \\
\textbf{ETTm1}   & \(0.291{\pm}0.003\) & \(0.319{\pm}0.002\) & \(0.351{\pm}0.001\) & \(0.410{\pm}0.002\) & \(0.315{\pm}0.003\) & \(0.351{\pm}0.008\) & \(0.377{\pm}0.002\) & \(0.410{\pm}0.002\) \\
\textbf{ETTm2}   & \(0.162{\pm}0.001\) & \(0.216{\pm}0.001\) & \(0.268{\pm}0.002\) & \(0.349{\pm}0.002\) & \(0.184{\pm}0.003\) & \(0.239{\pm}0.000\) & \(0.278{\pm}0.001\) & \(0.349{\pm}0.002\) \\
\textbf{Weather} & \(0.143{\pm}0.001\) & \(0.185{\pm}0.001\) & \(0.235{\pm}0.001\) & \(0.313{\pm}0.001\) & \(0.145{\pm}0.001\) & \(0.189{\pm}0.001\) & \(0.243{\pm}0.001\) & \(0.313{\pm}0.001\) \\
\textbf{ECL}     & \(0.128{\pm}0.001\) & \(0.145{\pm}0.000\) & \(0.162{\pm}0.001\) & \(0.199{\pm}0.000\) & \(0.130{\pm}0.000\) & \(0.158{\pm}0.001\) & \(0.178{\pm}0.000\) & \(0.199{\pm}0.000\) \\
\textbf{Traffic} & \(0.359{\pm}0.000\) & \(0.375{\pm}0.000\) & \(0.389{\pm}0.001\) & \(0.428{\pm}0.001\) & \(0.363{\pm}0.001\) & \(0.378{\pm}0.001\) & \(0.391{\pm}0.002\) & \(0.428{\pm}0.001\) \\
\bottomrule
\end{tabular}
}
\end{table}

Table~\ref{tab:padding-best-coef-combined} shows a clear ordering. Repeat padding is consistently the weakest choice, especially on the ETT datasets: for example, on ETTh2 it increases the best MSE from the zero-padding result in Table~\ref{table:main_res} by roughly $0.039$, $0.033$, and $0.019$ at horizons 96, 192, and 336, respectively. Average padding is much closer to zero-padding, and in several large-data or low-variance settings the difference is small; for instance, Weather and Traffic change only marginally at short horizons. However, average padding still rarely improves over zero-padding, and it can be noticeably worse on settings such as ETTh1 at horizon 336. At horizon 720 the padding choices become nearly indistinguishable because the forecast block already spans the full lookback window, leaving little or no missing region to fill. Overall, the results support the theoretical interpretation: the most reliable padding rule is the one that treats uninserted coordinates as zero perturbations rather than injecting artificial residual structure into them.

\subsection{Consistency Loss Positional Bias}
\label{app:exp-pos-bias}

To evaluate the effect of using a deterministic insertion location, we test consistency loss with history padding and select the best coefficient \(\lambda\) for each dataset--horizon pair. Table~\ref{tab:history-padding-best-coef} reports the best test MSE obtained under settings similar to those of the main experiments (see Appendix~\ref{app:exp_details}). The results show that, except at the 720-step forecasting horizon where no positional bias is introduced, the performance of the full consistency loss is adversely affected by positional bias.

\begin{table}[!htbp]
\centering
\caption{Best test MSE achieved by consistency loss with history padding.}
\label{tab:history-padding-best-coef}
\small
\setlength{\tabcolsep}{5pt}
\renewcommand{\arraystretch}{1.08}
\begin{tabular}{@{}l|cccc@{}}
\toprule
\multirow{2}{*}{\textbf{Dataset}} & \multicolumn{4}{c}{\textbf{Forecasting horizon}} \\
\cmidrule(lr){2-5}
& \textbf{96} & \textbf{192} & \textbf{336} & \textbf{720} \\
\midrule

	\textbf{ETTh1} & \(0.369{\pm}0.005\) & \(0.402{\pm}0.008\) & \(0.432{\pm}0.001\) & \(0.446{\pm}0.019\) \\
	\textbf{ETTh2} & \(0.288{\pm}0.002\) & \(0.337{\pm}0.004\) & \(0.362{\pm}0.002\) & \(0.396{\pm}0.009\) \\
	\textbf{ETTm1} & \(0.310{\pm}0.001\) & \(0.335{\pm}0.002\) & \(0.361{\pm}0.003\) & \(0.414{\pm}0.002\) \\
	\textbf{ETTm2} & \(0.168{\pm}0.003\) & \(0.225{\pm}0.003\) & \(0.278{\pm}0.008\) & \(0.369{\pm}0.006\) \\
    \textbf{Weather} & \(0.145{\pm}0.001\) & \(0.186{\pm}0.001\) & \(0.236{\pm}0.001\) & \(0.312{\pm}0.001\) \\
    \textbf{ECL} & \(0.129{\pm}0.001\) & \(0.148{\pm}0.000\) & \(0.163{\pm}0.000\) & \(0.198{\pm}0.001\) \\
	\textbf{Traffic} & \(0.359{\pm}0.000\) & \(0.375{\pm}0.000\) & \(0.389{\pm}0.001\) & \(0.428{\pm}0.001\) \\

\bottomrule
\end{tabular}
\end{table}

\subsection{Directional vs. Isotropic Consistency Target}
\label{app:exp-isotropic-cl}

This experiment tests whether the gains of consistency loss come merely from generic Jacobian smoothing. We replace the residual-direction JVP penalty with a Jacobian Frobenius penalty, which suppresses sensitivity in all input directions rather than only the forecast-error direction. Table~\ref{tab:fnorm-best-coef} shows that isotropic Jacobian regularization can help, but its results are generally weaker than the directional consistency results in Table~\ref{table:main_res}. This supports the main claim in Section~\ref{sec:related-work}: the useful regularization is not simply smaller Jacobian norm, but selective control along the directions through which the model's own errors are fed back.

\begin{table}[!htbp]
\centering
\caption{Best test MSE achieved by Jacobian Frobenius regularization.}
\label{tab:fnorm-best-coef}
\small
\setlength{\tabcolsep}{5pt}
\renewcommand{\arraystretch}{1.08}
\begin{tabular}{@{}l|cccc@{}}
\toprule
\multirow{2}{*}{\textbf{Dataset}} & \multicolumn{4}{c}{\textbf{Forecasting horizon}} \\
\cmidrule(lr){2-5}
& \textbf{96} & \textbf{192} & \textbf{336} & \textbf{720} \\
\midrule
\textbf{ETTh1}   & \(0.379{\pm}0.001\) & \(0.402{\pm}0.001\) & \(0.426{\pm}0.004\) & \(0.441{\pm}0.001\) \\
\textbf{ETTh2}   & \(0.285{\pm}0.001\) & \(0.335{\pm}0.002\) & \(0.356{\pm}0.002\) & \(0.386{\pm}0.001\) \\
\textbf{ETTm1}   & \(0.304{\pm}0.001\) & \(0.339{\pm}0.002\) & \(0.366{\pm}0.003\) & \(0.413{\pm}0.002\) \\
\textbf{ETTm2}   & \(0.178{\pm}0.003\) & \(0.224{\pm}0.000\) & \(0.271{\pm}0.001\) & \(0.352{\pm}0.002\) \\
\textbf{Weather} & \(0.156{\pm}0.001\) & \(0.199{\pm}0.001\) & \(0.244{\pm}0.001\) & \(0.310{\pm}0.001\) \\
\textbf{ECL}     & \(0.129{\pm}0.000\) & \(0.148{\pm}0.000\) & \(0.163{\pm}0.001\) & \(0.199{\pm}0.001\) \\
\textbf{Traffic} & \(0.358{\pm}0.000\) & \(0.375{\pm}0.001\) & \(0.388{\pm}0.001\) & \(0.426{\pm}0.001\) \\
\bottomrule
\end{tabular}
\end{table}

\subsection{Rollout Stability}
\label{app:exp-rollout}

Section~\ref{sec:manifold-aware-consistency} gives a conditional geometric interpretation of consistency loss: by reducing sensitivity to residual-induced perturbations, the model can better control off-manifold error propagation and keep free-running trajectories inside a stable predictive tube. We test this implication directly with a rollout experiment. In both the CL and non-CL settings, we train a short-horizon model with lookback length $L=720$ and forecast length $H=96$, then recursively feed predictions back into the model and evaluate the resulting rollout MSE at horizons 336 and 720. As a reference, we also report the one-shot block-forecasting results, where separate models are trained directly for $H=336$ and $H=720$.

\begin{table}[!htbp]
\centering
\caption{Test MSE comparison between rollout forecasting (ours vs.\ baseline) and one-shot block forecasting (mean$\pm$std over seeds).}
\label{tab:app-rollout-vs-oneblock}
\setlength{\tabcolsep}{3pt}
\renewcommand{\arraystretch}{1.08}
\begin{tabular}{@{}llcc|cc@{}}
\toprule
\multirow{2}{*}{\textbf{Dataset}} & \multirow{2}{*}{\textbf{Method}} & \multicolumn{2}{c}{\textbf{+ Consistency Loss}} & \multicolumn{2}{c}{\textbf{Baseline (without CL)}} \\
\cmidrule(lr){3-4} \cmidrule(lr){5-6}
& & \textbf{336} & \textbf{720} & \textbf{336} & \textbf{720} \\
\midrule
\multirow{2}{*}{\textbf{ETTh1}}
  & Rollout   & \(0.411{\pm}0.003\) & \(0.427{\pm}0.009\) & \(0.461{\pm}0.007\) & \(0.508{\pm}0.017\) \\
  & One-shot  & \(0.434{\pm}0.001\) & \(0.458{\pm}0.005\) & \(0.458{\pm}0.006\) & \(0.500{\pm}0.009\) \\
\midrule
\multirow{2}{*}{\textbf{ETTh2}}
  & Rollout   & \(0.361{\pm}0.002\) & \(0.399{\pm}0.002\) & \(0.403{\pm}0.018\) & \(0.428{\pm}0.016\) \\
  & One-shot  & \(0.362{\pm}0.001\) & \(0.397{\pm}0.000\) & \(0.395{\pm}0.001\) & \(0.452{\pm}0.009\) \\
\midrule
\multirow{2}{*}{\textbf{ETTm1}}
  & Rollout   & \(0.353{\pm}0.003\) & \(0.412{\pm}0.005\) & \(0.383{\pm}0.008\) & \(0.452{\pm}0.020\) \\
  & One-shot  & \(0.352{\pm}0.001\) & \(0.410{\pm}0.002\) & \(0.377{\pm}0.003\) & \(0.446{\pm}0.006\) \\
\midrule
\multirow{2}{*}{\textbf{ETTm2}}
  & Rollout   & \(0.265{\pm}0.002\) & \(0.346{\pm}0.002\) & \(0.299{\pm}0.006\) & \(0.381{\pm}0.011\) \\
  & One-shot  & \(0.268{\pm}0.001\) & \(0.348{\pm}0.001\) & \(0.296{\pm}0.004\) & \(0.378{\pm}0.008\) \\
\midrule
\multirow{2}{*}{\textbf{Weather}}
  & Rollout   & \(0.235{\pm}0.002\) & \(0.308{\pm}0.001\) & \(0.263{\pm}0.012\) & \(0.345{\pm}0.022\) \\
  & One-shot  & \(0.235{\pm}0.001\) & \(0.312{\pm}0.001\) & \(0.258{\pm}0.002\) & \(0.336{\pm}0.003\) \\
\midrule
\multirow{2}{*}{\textbf{ECL}}
  & Rollout   & \(0.161{\pm}0.000\) & \(0.200{\pm}0.000\) & \(0.162{\pm}0.000\) & \(0.200{\pm}0.001\) \\
  & One-shot  & \(0.160{\pm}0.001\) & \(0.198{\pm}0.001\) & \(0.163{\pm}0.001\) & \(0.198{\pm}0.001\) \\
\midrule
\multirow{2}{*}{\textbf{Traffic}}
  & Rollout   & \(0.391{\pm}0.000\) & \(0.428{\pm}0.000\) & \(0.392{\pm}0.000\) & \(0.428{\pm}0.000\) \\
  & One-shot  & \(0.388{\pm}0.000\) & \(0.428{\pm}0.001\) & \(0.388{\pm}0.000\) & \(0.428{\pm}0.001\) \\
\bottomrule
\end{tabular}
\end{table}

Table~\ref{tab:app-rollout-vs-oneblock} shows that CL consistently improves rollout forecasting over the baseline without CL. The improvement is most pronounced on the ETT and Weather datasets, where rollout errors are more sensitive to accumulated prediction mistakes. For example, on ETTh1 the 720-step rollout MSE drops from $0.508$ to $0.427$, and on Weather it drops from $0.345$ to $0.308$. Comparing rollout with one-shot forecasting also reveals a useful contrast. For the baseline, recursive rollout is usually worse than training a separate one-shot model directly for the target horizon, indicating that errors may be amplified when the model repeatedly consumes its own predictions. With CL, this rollout degradation is much smaller: the $H=96$ model remains close to, and sometimes better than, the corresponding one-shot $H=336$ or $H=720$ model. On larger datasets such as ECL and Traffic, both CL and baseline are already close, so the rollout gap is correspondingly small. Overall, these observations support the stability interpretation in Section~\ref{sec:manifold-aware-consistency}: consistency loss does not only improve one-shot accuracy, but also makes the model less prone to error amplification during free-running rollout.

\subsection{Computational Analysis}
\label{app:exp-compute}

\paragraph{Training cost.}
The JVP evaluates the directional response without materializing the dense $d_y\times d_x$ Jacobian. This avoids storing $d_xd_y$ Jacobian entries, but still incurs additional training cost. Let $F$ and $B$ denote the costs of an ordinary forward and backward pass, respectively, on the same minibatch. Ignoring optimizer-update costs, a coarse pass-count model gives
\begin{equation}
\label{eq:app-cl-compute-full}
T_{\mathrm{MSE}}\approx F+B,
\qquad
T_{\mathrm{JVP}}\approx 3F+2B.
\end{equation}
The full-JVP implementation first performs an ordinary forward pass to obtain the prediction and stopped-gradient residual direction. The forward-mode JVP then propagates both primal and tangent values, and the combined loss is differentiated backward. Equation~\eqref{eq:app-cl-compute-full} is an approximate accounting of these operations.

\paragraph{Subsampling the consistency loss.}
For an example-wise forecaster and a minibatch of size $N$, let $c_i$ be the consistency penalty for example $i$, with the same stopped-gradient residual convention as in the main method. Sample a subset $\mathcal I\subset\{1,\ldots,N\}$ of size $m$ uniformly without replacement and average the penalties over that subset. If each $c_i$ is unchanged by the choice of $\mathcal I$, then, conditional on the batch,
\begin{equation}
\label{eq:app-cl-subsample}
\mathbb E_{\mathcal I}\!\left[\frac{1}{m}\sum_{i\in\mathcal I}c_i\right]
=\frac{1}{m}\sum_{i=1}^{N}\frac{m}{N}c_i
=\frac{1}{N}\sum_{i=1}^{N}c_i.
\end{equation}
Subsampling preserves the expected batch-mean consistency objective and its gradient while being much cheaper to compute. The prediction loss still uses the full minibatch. With $\rho=m/N$, the corresponding cost model is
\begin{equation}
\label{eq:app-cl-compute-sampled}
T(\rho)\approx F+B+\rho(2F+B).
\end{equation}
For example, if $B\approx2F$, the model predicts a full-JVP multiplier of $7/3$ and a $1/16$-JVP multiplier of approximately $1.083$. The experiments below measure the resulting trade-off between computation and forecasting accuracy.

\paragraph{Measured time and memory.}
Table~\ref{tab:app-compute-resources} reports wall time for 100 training iterations and peak reserved GPU memory on ECL (Electricity; 321 variables) with lookback $L=720$ and forecast length $H=720$, using an NVIDIA A100 after 100 warm-up iterations. We average three runs; timing standard deviations are 1--2\,ms and are omitted. The workload and training batch size are identical across the three configurations within each backbone.

\begin{table}[!htbp]
\centering
\caption{Training time (seconds per 100 iterations) and peak reserved memory (GiB) on ECL ($L=H=720$). Parentheses show multipliers over ``MSE-only" setting.}
\label{tab:app-compute-resources}
\footnotesize
\setlength{\tabcolsep}{4pt}
\renewcommand{\arraystretch}{1.08}
\begin{tabular}{@{}lclccc@{}}
\toprule
\textbf{Backbone} & $N$ & \textbf{Metric} & \textbf{MSE only} & \textbf{Full JVP} & \textbf{$1/16$ JVP} \\
\midrule
\multirow{2}{*}{ResMLP} & \multirow{2}{*}{128} & Time (s) & 17.564 ($1.00\times$) & 38.397 ($2.19\times$) & 19.009 ($1.08\times$) \\
 &  & Memory (GiB) & 1.918 ($1.00\times$) & 3.931 ($2.05\times$) & 1.837 ($\approx1.00\times$) \\
\midrule
\multirow{2}{*}{FilterNet} & \multirow{2}{*}{64} & Time (s) & 8.847 ($1.00\times$) & 9.146 ($1.03\times$) & 9.154 ($1.03\times$) \\
 &  & Memory (GiB) & 0.610 ($1.00\times$) & 1.055 ($1.73\times$) & 0.638 ($1.04\times$) \\
\midrule
\multirow{2}{*}{iTransformer} & \multirow{2}{*}{64} & Time (s) & 9.437 ($1.00\times$) & 29.380 ($3.11\times$) & 11.925 ($1.26\times$) \\
 &  & Memory (GiB) & 3.258 ($1.00\times$) & 11.205 ($3.44\times$) & 3.756 ($1.15\times$) \\
\midrule
\multirow{2}{*}{TSLANet} & \multirow{2}{*}{32} & Time (s) & 16.363 ($1.00\times$) & 61.863 ($3.78\times$) & 19.587 ($1.20\times$) \\
 &  & Memory (GiB) & 13.677 ($1.00\times$) & 38.708 ($2.83\times$) & 13.649 ($\approx1.00\times$) \\
\bottomrule
\end{tabular}
\end{table}

Full JVP increases wall time by $1.03$--$3.78\times$ across the measured backbones, whereas $1/16$ JVP uses $1.03$--$1.26\times$ the MSE-only wall time. Equivalently, their relative throughputs are $0.26$--$0.97\times$ and $0.79$--$0.97\times$, respectively. For ResMLP, subsampling reduces the time overhead from approximately $119\%$ to $8\%$, with peak reserved memory close to the baseline. 

\paragraph{Accuracy with sampled JVPs: ResMLP.}
Table~\ref{tab:app-sampled-mlp-full} gives all 28 dataset--horizon results for each JVP configuration, and Table~\ref{tab:app-sampled-mlp-summary} summarizes MSE over the seven datasets. Full-JVP results are from Table~\ref{tab:app-r2r-var}. The training batch size is $N=128$, so $m=8$ and $m=2$ correspond to sampling ratios of $1/16$ and $1/64$. The controlled $\lambda_{\mathrm{MLP}}=0$ reference is the baseline in Appendix~\ref{app:exp-mlp-zero}. With $m=8$, the overall MSE is $0.3000$, compared with $0.3002$ for full JVP; with $m=2$, it is $0.3013$. Both sampled configurations achieve lower or equal mean MSE than the controlled baseline in all 28 settings. Thus, the substantial reduction in JVP computation preserves comparable forecasting accuracy in these experiments. CL adds no model parameters or inference-time computation.

\begin{table}[!htbp]
\centering
\caption{ResMLP test MSE with full and sampled JVPs ($N=128$). Best values are \textbf{bold}.}
\label{tab:app-sampled-mlp-full}
\small
\setlength{\tabcolsep}{6pt}
\renewcommand{\arraystretch}{1.08}
\begin{tabular}{@{}llcccc@{}}
\toprule
\multirow{2}{*}{\textbf{Dataset}} & \multirow{2}{*}{\textbf{Training Setting}} & \multicolumn{4}{c}{\textbf{Forecasting horizon}} \\
\cmidrule(lr){3-6}
 &  & 96 & 192 & 336 & 720 \\
\midrule
\multirow{3}{*}{ETTh1} & Full JVP & \textbf{0.358} & \textbf{0.395} & \textbf{0.434} & 0.458 \\
 & $m=8$ JVP ($1/16$) & 0.359 & 0.396 & 0.435 & \textbf{0.449} \\
 & $m=2$ JVP ($1/64$) & 0.361 & 0.398 & 0.437 & 0.456 \\
\midrule
\multirow{3}{*}{ETTh2} & Full JVP & \textbf{0.269} & 0.331 & \textbf{0.362} & \textbf{0.397} \\
 & $m=8$ JVP ($1/16$) & \textbf{0.269} & \textbf{0.330} & \textbf{0.362} & 0.399 \\
 & $m=2$ JVP ($1/64$) & 0.271 & 0.332 & \textbf{0.362} & 0.401 \\
\midrule
\multirow{3}{*}{ETTm1} & Full JVP & \textbf{0.282} & \textbf{0.317} & \textbf{0.352} & \textbf{0.410} \\
 & $m=8$ JVP ($1/16$) & \textbf{0.282} & 0.318 & 0.353 & 0.411 \\
 & $m=2$ JVP ($1/64$) & 0.284 & 0.320 & 0.354 & 0.412 \\
\midrule
\multirow{3}{*}{ETTm2} & Full JVP & \textbf{0.158} & \textbf{0.215} & \textbf{0.268} & \textbf{0.348} \\
 & $m=8$ JVP ($1/16$) & 0.160 & \textbf{0.215} & \textbf{0.268} & 0.349 \\
 & $m=2$ JVP ($1/64$) & 0.160 & 0.216 & 0.269 & 0.350 \\
\midrule
\multirow{3}{*}{Weather} & Full JVP & \textbf{0.142} & \textbf{0.185} & \textbf{0.235} & 0.312 \\
 & $m=8$ JVP ($1/16$) & 0.143 & \textbf{0.185} & \textbf{0.235} & \textbf{0.311} \\
 & $m=2$ JVP ($1/64$) & 0.144 & 0.186 & 0.236 & 0.312 \\
\midrule
\multirow{3}{*}{ECL} & Full JVP & 0.127 & 0.144 & 0.160 & 0.198 \\
 & $m=8$ JVP ($1/16$) & \textbf{0.126} & \textbf{0.143} & \textbf{0.159} & \textbf{0.197} \\
 & $m=2$ JVP ($1/64$) & \textbf{0.126} & \textbf{0.143} & \textbf{0.159} & 0.198 \\
\midrule
\multirow{3}{*}{Traffic} & Full JVP & \textbf{0.358} & 0.375 & 0.388 & \textbf{0.428} \\
 & $m=8$ JVP ($1/16$) & \textbf{0.358} & \textbf{0.374} & \textbf{0.387} & \textbf{0.428} \\
 & $m=2$ JVP ($1/64$) & \textbf{0.358} & 0.375 & 0.388 & \textbf{0.428} \\
\bottomrule
\end{tabular}
\end{table}

\begin{table}[!htbp]
\centering
\caption{ResMLP test MSE averaged across seven datasets ($N=128$); Overall MSE averages all 28 experiment settings. The controlled-baseline row averages the reported means in Table~\ref{tab:mlp-zero-baseline}.}
\label{tab:app-sampled-mlp-summary}
\small
\setlength{\tabcolsep}{5pt}
\renewcommand{\arraystretch}{1.08}
\begin{tabular}{@{}lccccc@{}}
\toprule
Training Setting & $H=96$ & $H=192$ & $H=336$ & $H=720$ & Overall \\
\midrule
No CL on MLP branch ($\lambda_{\mathrm{MLP}}=0$) & 0.2621 & 0.3019 & 0.3347 & 0.3936 & 0.3231 \\
Full JVP & 0.2420 & 0.2803 & 0.3141 & 0.3644 & 0.3002 \\
$m=8$ JVP ($1/16$) & 0.2424 & 0.2801 & 0.3141 & 0.3634 & 0.3000 \\
$m=2$ JVP ($1/64$) & 0.2434 & 0.2814 & 0.3150 & 0.3653 & 0.3013 \\
\bottomrule
\end{tabular}
\end{table}

\Needspace{6\baselineskip}
\paragraph{Accuracy with sampled JVPs: other backbones.}
We evaluate a fixed JVP subset of $m=2$ on ETTh2, ETTm2, and ECL at all four forecasting horizons. Tables~\ref{tab:app-sampled-backbone-summary} and~\ref{tab:app-sampled-backbone-full} report the mean and per-setting MSE, respectively. MSE-only and full-JVP values are from Table~\ref{tab:cl_results}. Full- and sampled-JVP means use the unrounded results; per-setting values are displayed to three decimals. The BN for PDF strategy follows Appendix~\ref{app:exp_details}.

\begin{table}[!htbp]
\centering
\caption{Mean test MSE across 12 settings per backbone. $\Delta$ is the relative change from full JVP.}
\label{tab:app-sampled-backbone-summary}
\small
\setlength{\tabcolsep}{6pt}
\renewcommand{\arraystretch}{1.08}
\begin{tabular}{@{}lcccc@{}}
\toprule
Backbone & MSE only & Full JVP & $m=2$ JVP & $\Delta$ (\%) \\
\midrule
FilterNet & 0.2723 & 0.2590 & \textbf{0.2583} & $-0.26$ \\
TSLANet & 0.2672 & 0.2561 & \textbf{0.2548} & $-0.50$ \\
iTransformer & 0.2730 & \textbf{0.2602} & 0.2626 & $+0.89$ \\
PDF & 0.2582 & \textbf{0.2503} & 0.2504 & $+0.06$ \\
\bottomrule
\end{tabular}
\end{table}

\begin{table}[!htbp]
\centering
\caption{Full versus sampled ($m=2$) JVP: test MSE by dataset and horizon.}
\label{tab:app-sampled-backbone-full}
\footnotesize
\setlength{\tabcolsep}{2.2pt}
\renewcommand{\arraystretch}{1.08}
\setlength{\aboverulesep}{1pt}
\setlength{\belowrulesep}{1pt}
\resizebox{\textwidth}{!}{%
\begin{tabular}{@{}ll|cccc|cccc|cccc@{}}
\toprule
\multirow{2}{*}{\textbf{Model}} & \multirow{2}{*}{\textbf{Training Setting}} & \multicolumn{4}{c|}{ETTh2} & \multicolumn{4}{c|}{ETTm2} & \multicolumn{4}{c}{ECL} \\
\cmidrule(lr){3-6}\cmidrule(lr){7-10}\cmidrule(lr){11-14}
 & & 96 & 192 & 336 & 720 & 96 & 192 & 336 & 720 & 96 & 192 & 336 & 720 \\
\midrule
\multirow{2}{*}{FilterNet} & Full JVP & \textbf{0.278} & \textbf{0.341} & 0.382 & 0.428 & \textbf{0.164} & \textbf{0.224} & \textbf{0.273} & \textbf{0.352} & \textbf{0.137} & \textbf{0.153} & \textbf{0.169} & \textbf{0.206} \\
 & $m=2$ JVP & 0.280 & \textbf{0.341} & \textbf{0.375} & \textbf{0.420} & \textbf{0.164} & 0.225 & 0.277 & \textbf{0.352} & \textbf{0.137} & \textbf{0.153} & \textbf{0.169} & \textbf{0.206} \\
\midrule
\multirow{2}{*}{TSLANet} & Full JVP & 0.270 & 0.332 & 0.364 & 0.411 & 0.190 & \textbf{0.239} & \textbf{0.282} & \textbf{0.358} & 0.128 & \textbf{0.144} & 0.160 & 0.194 \\
 & $m=2$ JVP & \textbf{0.269} & \textbf{0.331} & \textbf{0.362} & \textbf{0.399} & \textbf{0.189} & 0.240 & 0.284 & 0.359 & \textbf{0.127} & \textbf{0.144} & \textbf{0.159} & \textbf{0.193} \\
\midrule
\multirow{2}{*}{iTransformer} & Full JVP & \textbf{0.284} & \textbf{0.351} & \textbf{0.391} & 0.432 & \textbf{0.164} & \textbf{0.224} & \textbf{0.278} & \textbf{0.355} & \textbf{0.132} & \textbf{0.153} & \textbf{0.167} & 0.191 \\
 & $m=2$ JVP & 0.294 & 0.361 & 0.399 & \textbf{0.428} & 0.165 & \textbf{0.224} & 0.280 & 0.357 & \textbf{0.132} & 0.154 & \textbf{0.167} & \textbf{0.190} \\
\midrule
\multirow{2}{*}{PDF} & Full JVP & \textbf{0.272} & 0.336 & 0.371 & 0.414 & 0.160 & \textbf{0.217} & \textbf{0.269} & \textbf{0.346} & \textbf{0.126} & \textbf{0.143} & \textbf{0.159} & \textbf{0.191} \\
 & $m=2$ JVP & 0.274 & \textbf{0.335} & \textbf{0.368} & \textbf{0.409} & \textbf{0.159} & 0.219 & 0.273 & 0.348 & 0.127 & 0.144 & \textbf{0.159} & 0.192 \\
\bottomrule
\end{tabular}}
\end{table}

Across all four backbones, the mean MSE over 12 settings differs by less than $0.9\%$ between $m=2$ and full JVP. At three-decimal precision, sampled JVP is no worse than MSE-only in all 48 settings.

\FloatBarrier
\endgroup
\section{Extended Related Work}
\label{app:extended-related-work}

\paragraph{Teacher forcing, consistency, and rollout mismatch.}
The mismatch between teacher forcing and free-running rollout has been studied in autoregressive sequence prediction, imitation learning, and learned dynamics. Scheduled Sampling \citep{bengio2015scheduled} reduces this mismatch by mixing ground-truth and model-generated histories during training, while Professor Forcing \citep{lamb2016professor} adversarially aligns hidden-state dynamics between teacher-forced and sampling regimes. In structured prediction and imitation learning, DAgger \citep{ross2011dagger} addresses the same induced-distribution problem by aggregating learner-visited states. Closer to forecasting and learned dynamics, Data as Demonstrator (DaD) \citep{venkatraman2015improving} augments training with model-generated states to improve multi-step prediction, Talvitie's stable-rollout regularization \citep{talvitie2014model} explicitly targets rollout stability, and the Flipped Classroom curricula of Teutsch and M\"ader \citep{teutsch2022flipped} interpolate between teacher forcing and free running for long-horizon time-series forecasting. Our work is closest in spirit to this line, but differs in that we derive an explicit \emph{local output-level consistency} penalty from the exact discrepancy between teacher-forced and free-running next states, rather than modifying the sampling policy, adversarially matching hidden dynamics, or relying on a curriculum to expose the model to its own predictions.

\paragraph{Jacobian regularization, directional invariance, and noise stability.}
Jacobian-based regularization has been used to encourage invariance, robustness, and stable dynamics. Contractive autoencoders penalize the full Frobenius norm of the encoder Jacobian \citep{rifai2011contractive}; supervised Jacobian regularization improves robustness and classification margins \citep{hoffman2019robust}; certified-robustness and Lipschitz-learning approaches control Jacobians through operator constraints or architectural design \citep{singla2021improved,zhang2022rethinking}; and Pan and Duraisamy \citep{pan2018longtime} use Jacobian regularization to improve long-time prediction of nonlinear dynamical systems. Earlier directional variants include Tangent Prop \citep{simard1991tangent}, which penalizes Jacobian action on hand-specified invariance directions, while denoising and noise-injection viewpoints connect isotropic perturbations to Jacobian/Tikhonov penalties \citep{vincent2010stacked,bishop1995training}. In contrast, our regularizer is neither isotropic nor hand-designed: it penalizes \(Jg_\theta(x)v\) only along the \emph{model-induced perturbation direction} \(v = \delta_t := x_{t+1}^{\mathrm{TF}}-x_{t+1}^{\mathrm{FR}}\), the actual next-state displacement caused by the current prediction error. This makes our objective a data-adaptive directional Jacobian penalty rather than a generic smoothness or Lipschitz regularizer.

\paragraph{Time-series forecasting backbones and training under shift.}
Modern long-horizon time-series forecasting has been driven largely by architectural innovations, including efficient Transformer variants such as Informer \citep{zhou2021informer}, decomposition-based designs such as Autoformer and FEDformer \citep{wu2021autoformer,zhou2022fedformer}, simple linear baselines such as DLinear \citep{zeng2023transformers}, patch-based and channel-independent models such as PatchTST \citep{nie2023patchtst}, reversible normalization for non-stationarity such as RevIN \citep{kim2022reversible}, MLP encoder--decoder models such as TiDE \citep{das2023tide}, and inverted-tokenization Transformers such as iTransformer \citep{liu2024itransformer}. These works mostly modify the backbone architecture or the input normalization scheme. Our contribution is orthogonal: it is a training criterion that can be attached to linear, MLP, or Transformer forecasters. This also clarifies the role of the linear case in our analysis: DLinear makes residual propagation especially transparent, whereas the strong performance of PatchTST, TiDE, and iTransformer suggests that rollout-aware regularization should be understood as complementary to architecture choice rather than tied to any particular model family. Finally, unlike RevIN, which addresses marginal distribution shift through reversible normalization, our objective addresses the \emph{endogenous} distribution shift created by autoregressive feedback.

\paragraph{Classical estimators and structured shrinkage.}
From a statistical viewpoint, our method is related to structured shrinkage and biased estimation. Ridge/Tikhonov regularization \citep{hoerl1970ridge}, reduced-rank regression \citep{izenman1975reduced}, and total least squares \citep{golub1980total} all introduce structure to reduce risk under ill-conditioning or covariate corruption, while James--Stein and empirical-Bayes estimators \citep{james1961estimation,efron1975stein} shrink toward structured targets to improve aggregate mean-squared error; later extensions consider shrinkage toward subspaces \citep{oman1982contracting} and caution that misspecified shrinkage targets can reverse the advantage \citep{perlman2012reversing}. Our method is related in spirit but not identical to these estimators. The shrinkage is applied in \emph{residual space} rather than parameter space, the shrinkage metric is induced by downstream rollout sensitivity, and the target is state-dependent---a low-gain subspace or local predictive manifold---rather than a fixed point or static subspace. For this reason, we view the method as a learned, dynamical subspace-shrinkage rule on \emph{propagated} error.

\paragraph{World model training.}
Related concepts for addressing compounding error and train-test mismatch have also been explored in world-model training. Asadi et al.~\citep{asadi2019combating} address compounding error through multi-step models that directly predict the state reached after an action sequence, avoiding recursive one-step composition. Parthasarathy et al.~\citep{parthasarathy2025closing} study the mismatch between predictive training and gradient-based planning. Their Online World Modeling (OWM) augments training with simulator-corrected planner trajectories, while Adversarial World Modeling (AWM) uses adversarial perturbations of latent states and actions to improve the planning optimization landscape. However, insights from these rollout and planning settings need not transfer unchanged to time-series forecasting, where the prediction protocols and training signals differ. Our formulation focuses on forecast-induced residuals in observed sequences, without requiring simulator feedback or adversarial search.

\section{Discussion}
\label{sec:discussion}

Beyond its dynamical and geometric interpretations, the proposed objective also admits a statistical reading as a \emph{task-aware shrinkage estimator on propagated error}. This viewpoint is useful because it clarifies both what is being regularized and why the regularization can improve free-running performance even when it introduces local bias.

\paragraph{Residual-space generalized ridge and spectral shrinkage.}
In the linear case, the consistency-augmented objective can be written as
\begin{equation}
\label{eq:discussion-linear-metric}
\|r_t\|_2^2 + \lambda \|GPr_t\|_2^2
=
r_t^\top
\left(I+\lambda P^\top G^\top G P\right)
r_t.
\end{equation}
Equation~\eqref{eq:discussion-linear-metric} shows that the method is naturally interpreted as a generalized ridge or Tikhonov-type penalty in \emph{residual space}, rather than in parameter space. Classical ridge regression introduces bias in the regression coefficients in order to reduce quadratic risk under collinearity \citep{hoerl1970ridge}; by contrast, \eqref{eq:discussion-linear-metric} introduces bias against residual directions that are strongly amplified by downstream rollout. If \(GP = U\Sigma V^\top\) is a singular value decomposition, then
\begin{equation}
\label{eq:discussion-spectral}
\|GPr_t\|_2^2
=
\sum_i \sigma_i^2 \langle v_i,r_t\rangle^2,
\end{equation}
which makes explicit that the method performs anisotropic spectral shrinkage: directions with large downstream gain are penalized more heavily than directions with small gain. In this sense, the linear objective is closer in spirit to reduced-rank and spectral shrinkage estimators than to isotropic smoothing \citep{izenman1975reduced}. The nonlinear Jacobian form inherits the same structure locally through the state-dependent operator
\[
M_\theta(x_t):=Jg_\theta(x_{t+1}^{\mathrm{FR}})P,
\]
so that the first-order penalty \(\|M_\theta(x_t)r_t\|_2^2\) may be viewed as a local, state-dependent quadratic form on propagated residuals. This also sharpens the distinction from generic Jacobian regularization for dynamical prediction: rather than shrinking all local sensitivities uniformly, the present objective shrinks the \emph{residual-induced} directions that are actually realized by teacher-forced/free-running mismatch \citep{pan2018longtime}.

\paragraph{An endogenous errors-in-variables correction.}
A second interpretation is as a dynamic analogue of errors-in-variables estimation. By construction,
\[
x_{t+1}^{\mathrm{FR}}
=
x_{t+1}^{\mathrm{TF}} - Pr_t,
\]
so the regressor seen by the model at the next step is contaminated by the current prediction error. Classical total least squares and errors-in-variables methods are designed precisely for regression settings in which both predictors and responses are noisy \citep{golub1980total}. The analogy is not exact: in our case the covariate noise is endogenous, state-dependent, and generated by the model itself rather than by an external measurement process. Nevertheless, the statistical role is similar. Free-running rollout replaces a clean teacher-forced covariate with a contaminated one, and the consistency penalty acts as a downstream-risk-aware correction for this model-induced covariate error. From this perspective, the method may be viewed as a sequential, task-dependent form of total least squares, specialized to the self-generated regressors that arise in autoregressive prediction.

\paragraph{A Stein-like view: shrinkage toward low-gain structure.}
The same mechanism is also Stein-like, although not literally equivalent to the classical James--Stein estimator. James and Stein showed that, under quadratic loss, a biased shrinkage estimator can dominate the unbiased estimator for multivariate normal means; Stein's later analysis and the empirical Bayes treatment of Efron and Morris made explicit the broader principle that structured shrinkage can improve overall risk \citep{james1961estimation,stein1981mean,efron1975stein}. Our objective shares this principle. It accepts local bias in order to reduce a larger risk, here the free-running rollout risk rather than a one-shot squared-error loss. At the same time, the differences are important. First, the relevant risk is multi-step and dynamical. Second, the shrinkage target is not a fixed point such as the origin, but a state-dependent low-gain subspace associated with the local predictive geometry. Third, the shrinkage rule is not closed form, but learned implicitly through the training objective. For these reasons, the closest analogy is not classical James--Stein shrinkage toward a constant target, but a learned, state-dependent \emph{subspace} shrinkage rule on propagated error. This analogy is reinforced by Stein-type extensions that shrink toward subspaces rather than a single point \citep{oman1982contracting}. It also highlights a potential failure mode: if the local predictive geometry or the implied shrinkage target is poorly estimated, then shrinkage may be applied in the wrong directions, echoing the reverse-Stein caveat identified by \cite{perlman2012reversing}.

\paragraph{Self-induced denoising and innovation suppression.}
In the nonlinear setting, the Jacobian-based penalty also connects naturally to regularized denoising estimators. Contractive auto-encoders penalize Jacobian sensitivity in order to induce local contraction, while regularized denoising auto-encoders have been shown to preserve variation along high-density manifolds while suppressing perturbations orthogonal to them \citep{rifai2011contractive,alain2014regularized}. Our method is similar in spirit but differs in a crucial respect: the corruption distribution is not exogenous and user-specified, but endogenous and generated online by the model's own rollout mismatch. The objective is therefore better described as \emph{self-induced denoising}. This viewpoint is closely related to the innovations interpretation as well. In the classical innovations framework, the useful residual is the component of the signal that is genuinely new relative to the past \citep{kailath1970innovations}. The proposed penalty pushes reinjected residuals toward such innovation-like directions: once fed back into the state, they should carry as little predictable structure as possible into future forecasts. In this sense, the method does not merely reduce residual magnitude; it suppresses residual components that remain dynamically observable after rollout.

\paragraph{Outlook: risk estimation and geometric shrinkage.}
These connections suggest at least two statistical extensions. First, since Stein's unbiased risk estimate provides a way to evaluate shrinkage and spectral estimators, and has recently been shown to apply to reduced-rank estimators as well, it is natural to ask whether an analogous \emph{rollout-SURE} could be developed to tune the regularization strength or the effective off-manifold control in a principled way \citep{stein1981mean,hansen2018sure}. Second, the present geometric argument is phrased through local linearization of a predictive manifold. Recent extensions of the Stein effect to Fr\'echet means in non-Euclidean spaces suggest that shrinkage toward structured geometric targets need not be restricted to linear subspaces \citep{mccormack2022frechet}. This raises the possibility of replacing local linear geometric approximations by learned latent manifolds or manifold-valued shrinkage targets, leading to a more intrinsically geometric version of the present framework.

Overall, these comparisons suggest that the proposed method is best understood not as a generic smoothness regularizer, but as a learned shrinkage procedure on \emph{propagated} error: generalized ridge in residual space, spectral shrinkage in rollout-sensitive directions, endogenous correction for self-generated covariate noise, and, under its predictive-manifold interpretation, selective control of off-manifold drift.

\section{Limitations, Broader Impact, and Future Work}
\label{sec:limitations-future-work}

\subsection{Limitations}
\label{app:limitation}
The proposed consistency loss has several limitations. First, although JVPs avoid materializing the full Jacobian matrix, they introduce additional training cost. We quantify this cost in Appendix~\ref{app:exp-compute}. Full-batch JVP computation can be expensive, so we also provide subsampled JVPs when computational efficiency is critical. With a $1/16$ sampling ratio, the measured training time is $1.03$--$1.26\times$ that of MSE-only training across the profiled backbones.

Second, consistency loss introduces an additional regularization strength \(\lambda\). While our experiments show that the method improves performance over a broad range of \(\lambda\) values, overly large coefficients can over-regularize the model. In such cases, the model may suppress residual-induced directions too aggressively and lose useful predictive variation, especially when the data contain genuinely sensitive dynamics or high-gain but informative components. Thus, the consistency coefficient still needs to be selected with validation data, as with other regularization methods.

Third, although the objective is architecture-agnostic in principle, different forecasting backbones may respond differently to the same consistency penalty. Architectural choices such as residual connections, normalization layers, channel mixing, attention modules, patching strategies, and linear shortcuts can affect how residual-direction sensitivity is represented and regularized. Our experiments show that consistency loss improves several representative backbones, but a complete understanding of its interaction with architecture design remains an open question.

\subsection{Broader Impact and Future Directions}
\label{app:broader-impact}

The perspective developed in this paper suggests a broader way to design objectives for sequential models. The training loss need not only reward accurate one-step prediction; it can also specify which errors the model is allowed to remember and propagate. This is particularly relevant in settings where small self-generated deviations can accumulate over long horizons, such as scientific forecasting, control-adjacent prediction, and neural or physical dynamical systems.

At the same time, the present approach should not be read as advocating indiscriminate contraction. In many sequential domains, meaningful variability lives along low-dimensional state sets or predictive manifolds, and overly aggressive stabilization could suppress the very dynamics that make the model useful. Our predictive-manifold interpretation addresses this tension by targeting the directions that push rollouts away from the data-supported set while leaving along-manifold evolution unconstrained.

Several extensions are natural. One direction is to replace the abstract predictive-manifold assumption by learned latent manifolds or adaptive normal-bundle estimation. Another is to move from uniform tube bounds to average-case guarantees based on quantities such as transverse Lyapunov exponents. A third is empirical: to study how rollout-consistency penalties interact with modern foundation time-series models and large latent-state predictors. More broadly, the underlying principle is simple: a forecasting model should not merely reduce its errors; it should learn not to amplify its own mistakes.

\newpage

\section*{NeurIPS Paper Checklist}

\begin{enumerate}

\item {\bf Claims}
    \item[] Question: Do the main claims made in the abstract and introduction accurately reflect the paper's contributions and scope?
    \item[] Answer: \answerYes{} %
    \item[] Justification: The abstract and introduction accurately describe the paper’s main contributions, methodology, and intended scope, without overstating the empirical findings or generality of the proposed approach.

    \item[] Guidelines:
    \begin{itemize}
        \item The answer \answerNA{} means that the abstract and introduction do not include the claims made in the paper.
        \item The abstract and/or introduction should clearly state the claims made, including the contributions made in the paper and important assumptions and limitations. A \answerNo{} or \answerNA{} answer to this question will not be perceived well by the reviewers. 
        \item The claims made should match theoretical and experimental results, and reflect how much the results can be expected to generalize to other settings. 
        \item It is fine to include aspirational goals as motivation as long as it is clear that these goals are not attained by the paper. 
    \end{itemize}

\item {\bf Limitations}
    \item[] Question: Does the paper discuss the limitations of the work performed by the authors?
    \item[] Answer: \answerYes{} %
    \item[] Justification: The paper explicitly discusses the main limitations of the proposed method, including its assumptions, practical constraints, and potential failure cases in Appendix~\ref{app:limitation}
    \item[] Guidelines:
    \begin{itemize}
        \item The answer \answerNA{} means that the paper has no limitation while the answer \answerNo{} means that the paper has limitations, but those are not discussed in the paper. 
        \item The authors are encouraged to create a separate ``Limitations'' section in their paper.
        \item The paper should point out any strong assumptions and how robust the results are to violations of these assumptions (e.g., independence assumptions, noiseless settings, model well-specification, asymptotic approximations only holding locally). The authors should reflect on how these assumptions might be violated in practice and what the implications would be.
        \item The authors should reflect on the scope of the claims made, e.g., if the approach was only tested on a few datasets or with a few runs. In general, empirical results often depend on implicit assumptions, which should be articulated.
        \item The authors should reflect on the factors that influence the performance of the approach. For example, a facial recognition algorithm may perform poorly when image resolution is low or images are taken in low lighting. Or a speech-to-text system might not be used reliably to provide closed captions for online lectures because it fails to handle technical jargon.
        \item The authors should discuss the computational efficiency of the proposed algorithms and how they scale with dataset size.
        \item If applicable, the authors should discuss possible limitations of their approach to address problems of privacy and fairness.
        \item While the authors might fear that complete honesty about limitations might be used by reviewers as grounds for rejection, a worse outcome might be that reviewers discover limitations that aren't acknowledged in the paper. The authors should use their best judgment and recognize that individual actions in favor of transparency play an important role in developing norms that preserve the integrity of the community. Reviewers will be specifically instructed to not penalize honesty concerning limitations.
    \end{itemize}

\item {\bf Theory assumptions and proofs}
    \item[] Question: For each theoretical result, does the paper provide the full set of assumptions and a complete (and correct) proof?
    \item[] Answer: \answerYes{} %
    \item[] Justification: All theoretical results are stated together with their required assumptions, and complete proofs are provided in Appendix~\ref{app:thm-and-proof}.
    \item[] Guidelines:
    \begin{itemize}
        \item The answer \answerNA{} means that the paper does not include theoretical results. 
        \item All the theorems, formulas, and proofs in the paper should be numbered and cross-referenced.
        \item All assumptions should be clearly stated or referenced in the statement of any theorems.
        \item The proofs can either appear in the main paper or the supplemental material, but if they appear in the supplemental material, the authors are encouraged to provide a short proof sketch to provide intuition. 
        \item Inversely, any informal proof provided in the core of the paper should be complemented by formal proofs provided in appendix or supplemental material.
        \item Theorems and Lemmas that the proof relies upon should be properly referenced. 
    \end{itemize}

    \item {\bf Experimental result reproducibility}
    \item[] Question: Does the paper fully disclose all the information needed to reproduce the main experimental results of the paper to the extent that it affects the main claims and/or conclusions of the paper (regardless of whether the code and data are provided or not)?
    \item[] Answer: \answerYes{} %
    \item[] Justification: The paper provides details about the experimental setup, datasets, evaluation protocols, model configurations in Appendix~\ref{app:exp_details}. We also provide source code in the supplementary material in Appendix~\ref{app:exp_details}. We also provide source code in the supplementary materials.
    \item[] Guidelines:
    \begin{itemize}
        \item The answer \answerNA{} means that the paper does not include experiments.
        \item If the paper includes experiments, a \answerNo{} answer to this question will not be perceived well by the reviewers: Making the paper reproducible is important, regardless of whether the code and data are provided or not.
        \item If the contribution is a dataset and\slash or model, the authors should describe the steps taken to make their results reproducible or verifiable. 
        \item Depending on the contribution, reproducibility can be accomplished in various ways. For example, if the contribution is a novel architecture, describing the architecture fully might suffice, or if the contribution is a specific model and empirical evaluation, it may be necessary to either make it possible for others to replicate the model with the same dataset, or provide access to the model. In general. releasing code and data is often one good way to accomplish this, but reproducibility can also be provided via detailed instructions for how to replicate the results, access to a hosted model (e.g., in the case of a large language model), releasing of a model checkpoint, or other means that are appropriate to the research performed.
        \item While NeurIPS does not require releasing code, the conference does require all submissions to provide some reasonable avenue for reproducibility, which may depend on the nature of the contribution. For example
        \begin{enumerate}
            \item If the contribution is primarily a new algorithm, the paper should make it clear how to reproduce that algorithm.
            \item If the contribution is primarily a new model architecture, the paper should describe the architecture clearly and fully.
            \item If the contribution is a new model (e.g., a large language model), then there should either be a way to access this model for reproducing the results or a way to reproduce the model (e.g., with an open-source dataset or instructions for how to construct the dataset).
            \item We recognize that reproducibility may be tricky in some cases, in which case authors are welcome to describe the particular way they provide for reproducibility. In the case of closed-source models, it may be that access to the model is limited in some way (e.g., to registered users), but it should be possible for other researchers to have some path to reproducing or verifying the results.
        \end{enumerate}
    \end{itemize}

\item {\bf Open access to data and code}
    \item[] Question: Does the paper provide open access to the data and code, with sufficient instructions to faithfully reproduce the main experimental results, as described in supplemental material?
    \item[] Answer: \answerYes{} %
    \item[] Justification: The experiments are conducted on publicly available datasets (see Appendix~\ref{app:exp-dataset}) (see Appendix~\ref{app:exp-dataset}), and the code together with reproduction instructions will be provided in the supplementary material to facilitate faithful reproduction of the main results.

    \item[] Guidelines:
    \begin{itemize}
        \item The answer \answerNA{} means that paper does not include experiments requiring code.
        \item Please see the NeurIPS code and data submission guidelines (\url{https://neurips.cc/public/guides/CodeSubmissionPolicy}) for more details.
        \item While we encourage the release of code and data, we understand that this might not be possible, so \answerNo{} is an acceptable answer. Papers cannot be rejected simply for not including code, unless this is central to the contribution (e.g., for a new open-source benchmark).
        \item The instructions should contain the exact command and environment needed to run to reproduce the results. See the NeurIPS code and data submission guidelines (\url{https://neurips.cc/public/guides/CodeSubmissionPolicy}) for more details.
        \item The authors should provide instructions on data access and preparation, including how to access the raw data, preprocessed data, intermediate data, and generated data, etc.
        \item The authors should provide scripts to reproduce all experimental results for the new proposed method and baselines. If only a subset of experiments are reproducible, they should state which ones are omitted from the script and why.
        \item At submission time, to preserve anonymity, the authors should release anonymized versions (if applicable).
        \item Providing as much information as possible in supplemental material (appended to the paper) is recommended, but including URLs to data and code is permitted.
    \end{itemize}

\item {\bf Experimental setting/details}
    \item[] Question: Does the paper specify all the training and test details (e.g., data splits, hyperparameters, how they were chosen, type of optimizer) necessary to understand the results?
    \item[] Answer: \answerYes{} %
    \item[] Justification: The paper clearly specifies all key training and evaluation details, including data splits, hyperparameter settings and selection strategy, optimizer configuration, and other implementation choices required to fully understand and interpret the reported results. The related sections are Section~\ref{subsec:exp-setup}, Appendix~\ref{app:exp_details}, and Appendix~\ref{app:exp-dataset} The related sections are Section~\ref{subsec:exp-setup}, Appendix~\ref{app:exp_details}, and Appendix~\ref{app:exp-dataset}
    \item[] Guidelines:
    \begin{itemize}
        \item The answer \answerNA{} means that the paper does not include experiments.
        \item The experimental setting should be presented in the core of the paper to a level of detail that is necessary to appreciate the results and make sense of them.
        \item The full details can be provided either with the code, in appendix, or as supplemental material.
    \end{itemize}

\item {\bf Experiment statistical significance}
    \item[] Question: Does the paper report error bars suitably and correctly defined or other appropriate information about the statistical significance of the experiments? Refer to Appendix~\ref{app:exp-r2r-var-main} for details.
    \item[] Answer: \answerYes{} %
    \item[] Justification: The paper includes appropriate measures of variability such as error bars or multiple-run statistics where applicable, and reports results in a way that reflects the statistical significance and robustness of the experimental findings. Refer to Appendix~\ref{app:exp-r2r-var-main} for details.
    \item[] Guidelines:
    \begin{itemize}
        \item The answer \answerNA{} means that the paper does not include experiments.
        \item The authors should answer \answerYes{} if the results are accompanied by error bars, confidence intervals, or statistical significance tests, at least for the experiments that support the main claims of the paper.
        \item The factors of variability that the error bars are capturing should be clearly stated (for example, train/test split, initialization, random drawing of some parameter, or overall run with given experimental conditions).
        \item The method for calculating the error bars should be explained (closed form formula, call to a library function, bootstrap, etc.)
        \item The assumptions made should be given (e.g., Normally distributed errors).
        \item It should be clear whether the error bar is the standard deviation or the standard error of the mean.
        \item It is OK to report 1-sigma error bars, but one should state it. The authors should preferably report a 2-sigma error bar than state that they have a 96\% CI, if the hypothesis of Normality of errors is not verified.
        \item For asymmetric distributions, the authors should be careful not to show in tables or figures symmetric error bars that would yield results that are out of range (e.g., negative error rates).
        \item If error bars are reported in tables or plots, the authors should explain in the text how they were calculated and reference the corresponding figures or tables in the text.
    \end{itemize}

\item {\bf Experiments compute resources}
    \item[] Question: For each experiment, does the paper provide sufficient information on the computer resources (type of compute workers, memory, time of execution) needed to reproduce the experiments? Details in Appendix~\ref{app:exp_details}
    \item[] Answer: \answerYes{} %
    \item[] Justification: The paper reports the key computational details required for reproduction, including the type of hardware used, training time, and relevant resource constraints, allowing readers to reasonably estimate the compute requirements of the experiments. Details in Appendix~\ref{app:exp_details}
    \item[] Guidelines:
    \begin{itemize}
        \item The answer \answerNA{} means that the paper does not include experiments.
        \item The paper should indicate the type of compute workers CPU or GPU, internal cluster, or cloud provider, including relevant memory and storage.
        \item The paper should provide the amount of compute required for each of the individual experimental runs as well as estimate the total compute. 
        \item The paper should disclose whether the full research project required more compute than the experiments reported in the paper (e.g., preliminary or failed experiments that didn't make it into the paper). 
    \end{itemize}
    
\item {\bf Code of ethics}
    \item[] Question: Does the research conducted in the paper conform, in every respect, with the NeurIPS Code of Ethics \url{https://neurips.cc/public/EthicsGuidelines}?
    \item[] Answer: \answerYes{} %
    \item[] Justification: The research has been conducted in accordance with the NeurIPS Code of Ethics, adhering to its requirements on research integrity, responsible reporting, and consideration of potential ethical and societal impacts.
    \item[] Guidelines:
    \begin{itemize}
        \item The answer \answerNA{} means that the authors have not reviewed the NeurIPS Code of Ethics.
        \item If the authors answer \answerNo, they should explain the special circumstances that require a deviation from the Code of Ethics.
        \item The authors should make sure to preserve anonymity (e.g., if there is a special consideration due to laws or regulations in their jurisdiction).
    \end{itemize}

\item {\bf Broader impacts}
    \item[] Question: Does the paper discuss both potential positive societal impacts and negative societal impacts of the work performed?
    \item[] Answer: \answerNo{} %
    \item[] Justification: This work is methodological research on training objectives for time-series forecasting and does not introduce a deployed system, collect new human-subject data, or target a specific application domain. Any societal impacts would arise only through downstream uses of forecasting models in particular domains, which are outside the scope of this paper.
    \item[] Guidelines:
    \begin{itemize}
        \item The answer \answerNA{} means that there is no societal impact of the work performed.
        \item If the authors answer \answerNA{} or \answerNo, they should explain why their work has no societal impact or why the paper does not address societal impact.
        \item Examples of negative societal impacts include potential malicious or unintended uses (e.g., disinformation, generating fake profiles, surveillance), fairness considerations (e.g., deployment of technologies that could make decisions that unfairly impact specific groups), privacy considerations, and security considerations.
        \item The conference expects that many papers will be foundational research and not tied to particular applications, let alone deployments. However, if there is a direct path to any negative applications, the authors should point it out. For example, it is legitimate to point out that an improvement in the quality of generative models could be used to generate Deepfakes for disinformation. On the other hand, it is not needed to point out that a generic algorithm for optimizing neural networks could enable people to train models that generate Deepfakes faster.
        \item The authors should consider possible harms that could arise when the technology is being used as intended and functioning correctly, harms that could arise when the technology is being used as intended but gives incorrect results, and harms following from (intentional or unintentional) misuse of the technology.
        \item If there are negative societal impacts, the authors could also discuss possible mitigation strategies (e.g., gated release of models, providing defenses in addition to attacks, mechanisms for monitoring misuse, mechanisms to monitor how a system learns from feedback over time, improving the efficiency and accessibility of ML).
    \end{itemize}
    
\item {\bf Safeguards}
    \item[] Question: Does the paper describe safeguards that have been put in place for responsible release of data or models that have a high risk for misuse (e.g., pre-trained language models, image generators, or scraped datasets)?
    \item[] Answer: \answerNA{} %
    \item[] Justification: The paper poses no such risks.
    \item[] Guidelines:
    \begin{itemize}
        \item The answer \answerNA{} means that the paper poses no such risks.
        \item Released models that have a high risk for misuse or dual-use should be released with necessary safeguards to allow for controlled use of the model, for example by requiring that users adhere to usage guidelines or restrictions to access the model or implementing safety filters. 
        \item Datasets that have been scraped from the Internet could pose safety risks. The authors should describe how they avoided releasing unsafe images.
        \item We recognize that providing effective safeguards is challenging, and many papers do not require this, but we encourage authors to take this into account and make a best faith effort.
    \end{itemize}

\item {\bf Licenses for existing assets}
    \item[] Question: Are the creators or original owners of assets (e.g., code, data, models), used in the paper, properly credited and are the license and terms of use explicitly mentioned and properly respected?
    \item[] Answer: \answerYes{} %
    \item[] Justification: The paper properly credits the creators or original sources of all external assets used, and it explicitly states and respects the corresponding licenses and terms of use for code, data, and models. Refer to Section~\ref{subsec:exp-setup}, Appendix~\ref{app:exp_details}, and Appendix~\ref{app:exp-dataset}.
    \item[] Guidelines:
    \begin{itemize}
        \item The answer \answerNA{} means that the paper does not use existing assets.
        \item The authors should cite the original paper that produced the code package or dataset.
        \item The authors should state which version of the asset is used and, if possible, include a URL.
        \item The name of the license (e.g., CC-BY 4.0) should be included for each asset.
        \item For scraped data from a particular source (e.g., website), the copyright and terms of service of that source should be provided.
        \item If assets are released, the license, copyright information, and terms of use in the package should be provided. For popular datasets, \url{paperswithcode.com/datasets} has curated licenses for some datasets. Their licensing guide can help determine the license of a dataset.
        \item For existing datasets that are re-packaged, both the original license and the license of the derived asset (if it has changed) should be provided.
        \item If this information is not available online, the authors are encouraged to reach out to the asset's creators.
    \end{itemize}

\item {\bf New assets}
    \item[] Question: Are new assets introduced in the paper well documented and is the documentation provided alongside the assets?
    \item[] Answer: \answerNA{} %
    \item[] Justification: The paper does not introduce any new assets such as datasets, models, or tools that require accompanying documentation.
    \item[] Guidelines:
    \begin{itemize}
        \item The answer \answerNA{} means that the paper does not release new assets.
        \item Researchers should communicate the details of the dataset\slash code\slash model as part of their submissions via structured templates. This includes details about training, license, limitations, etc. 
        \item The paper should discuss whether and how consent was obtained from people whose asset is used.
        \item At submission time, remember to anonymize your assets (if applicable). You can either create an anonymized URL or include an anonymized zip file.
    \end{itemize}

\item {\bf Crowdsourcing and research with human subjects}
    \item[] Question: For crowdsourcing experiments and research with human subjects, does the paper include the full text of instructions given to participants and screenshots, if applicable, as well as details about compensation (if any)? 
    \item[] Answer: \answerNA{} %
    \item[] Justification: The paper does not involve crowdsourcing studies or human subject experiments; therefore, this requirement is not applicable.
    \item[] Guidelines:
    \begin{itemize}
        \item The answer \answerNA{} means that the paper does not involve crowdsourcing nor research with human subjects.
        \item Including this information in the supplemental material is fine, but if the main contribution of the paper involves human subjects, then as much detail as possible should be included in the main paper. 
        \item According to the NeurIPS Code of Ethics, workers involved in data collection, curation, or other labor should be paid at least the minimum wage in the country of the data collector. 
    \end{itemize}

\item {\bf Institutional review board (IRB) approvals or equivalent for research with human subjects}
    \item[] Question: Does the paper describe potential risks incurred by study participants, whether such risks were disclosed to the subjects, and whether Institutional Review Board (IRB) approvals (or an equivalent approval/review based on the requirements of your country or institution) were obtained?
    \item[] Answer: \answerNA{} %
    \item[] Justification: The paper does not involve human subject studies or participant experiments, and therefore no participant risks or IRB approval are relevant.
    \item[] Guidelines:
    \begin{itemize}
        \item The answer \answerNA{} means that the paper does not involve crowdsourcing nor research with human subjects.
        \item Depending on the country in which research is conducted, IRB approval (or equivalent) may be required for any human subjects research. If you obtained IRB approval, you should clearly state this in the paper. 
        \item We recognize that the procedures for this may vary significantly between institutions and locations, and we expect authors to adhere to the NeurIPS Code of Ethics and the guidelines for their institution. 
        \item For initial submissions, do not include any information that would break anonymity (if applicable), such as the institution conducting the review.
    \end{itemize}

\item {\bf Declaration of LLM usage}
    \item[] Question: Does the paper describe the usage of LLMs if it is an important, original, or non-standard component of the core methods in this research? Note that if the LLM is used only for writing, editing, or formatting purposes and does \emph{not} impact the core methodology, scientific rigor, or originality of the research, declaration is not required.
    \item[] Answer: \answerNA{} %
    \item[] Justification: LLMs were only used for writing and editing purposes and did not play any role in the core methodology or experimental design of the paper.
    \item[] Guidelines:
    \begin{itemize}
        \item The answer \answerNA{} means that the core method development in this research does not involve LLMs as any important, original, or non-standard components.
        \item Please refer to our LLM policy in the NeurIPS handbook for what should or should not be described.
    \end{itemize}

\end{enumerate}

\end{document}